\documentclass[11pt]{article}

\usepackage[final]{acl}

\usepackage{times}
\usepackage{latexsym}

\usepackage[T1]{fontenc}

\usepackage[utf8]{inputenc}

\usepackage{microtype}

\usepackage{inconsolata}

\usepackage{graphicx}
\usepackage{hyperref}
\usepackage{url}
\usepackage{booktabs}
\usepackage{amsmath}
\usepackage{amssymb}
\usepackage[table]{xcolor}
\usepackage{tcolorbox}
\usepackage{xcolor}
\usepackage{multirow}
\usepackage{wrapfig}
\usepackage{makecell}
\usepackage{arydshln}
\usepackage{pifont}

\title{Mitigating Catastrophic Forgetting in Target Language Adaptation of LLMs via Source-Shielded Updates}

\author{
Atsuki Yamaguchi$^{1}$ \quad
Terufumi Morishita$^{2}$ \quad
Aline Villavicencio$^{1,3,4}$ \quad
 Nikolaos Aletras$^{1}$\\ 
$^{1}$University of Sheffield, United Kingdom
\quad$^{2}$Hitachi, Ltd., Japan\\
$^{3}$University of Exeter, United Kingdom
\quad$^{4}$Federal University of Rio Grande do Norte, Brazil\\
\texttt{\{ayamaguchi1,a.villavicencio,n.aletras\}@sheffield.ac.uk} 
}

\begin{document}

\maketitle

\begin{abstract}
Expanding the linguistic diversity of instruct large language models (LLMs) is crucial for global accessibility but is often hindered by the reliance on costly specialized target language labeled data and catastrophic forgetting during adaptation.
We tackle this challenge under a realistic, low-resource constraint: adapting instruct LLMs using only unlabeled target language data.
We introduce \textbf{S}ource-\textbf{S}hielded \textbf{U}pdates (\textbf{SSU}), a selective parameter update strategy that proactively preserves source knowledge.
Using a small set of source data and a parameter importance scoring method, SSU identifies parameters critical to maintaining source abilities.
It then applies a column-wise freezing strategy to protect these parameters before adaptation.
Experiments across five typologically diverse languages and 7B and 13B models demonstrate that SSU successfully mitigates catastrophic forgetting. It reduces performance degradation on monolingual source tasks to just 3.4\% (7B) and 2.8\% (13B) on average, a stark contrast to the 20.3\% and 22.3\% from full fine-tuning. SSU also achieves target-language performance highly competitive with full fine-tuning, outperforming it on all benchmarks for 7B models and the majority for 13B models.\footnote{Our code and models are available via \url{https://github.com/gucci-j/ssu}.}
\end{abstract}

\section{Introduction} \label{sec:introduction}
Large language models (LLMs) demonstrate remarkable generalization across numerous applications~\citep{gpt5, deepseekai2025deepseekr1incentivizingreasoningcapability,yang2025qwen3technicalreport,gemmateam2025gemma3technicalreport}.
However, they notoriously underperform in languages absent or underrepresented in their training data, creating a barrier to equitable access for speakers worldwide~\citep{huang-etal-2023-languages}.
The standard approach to resolve this issue is continual pre-training (CPT) or fine-tuning on target language data~\citep{cui2024efficienteffectivetextencoding,ji2025emma500enhancingmassivelymultilingual}.

Yet, adapting instruct models to these languages is uniquely challenging. Such models require specialized instruction-tuning data~\citep{wei2022finetuned, NEURIPS2023_a85b405e}, which is often unavailable or prohibitively costly to create for underrepresented languages~\citep{huang-etal-2024-chat}.
Furthermore, machine-translated data as a low-cost alternative is not consistently effective~\citep{tao-etal-2024-unlocking}.

\begin{figure}
    \centering
    \includegraphics[width=\linewidth]{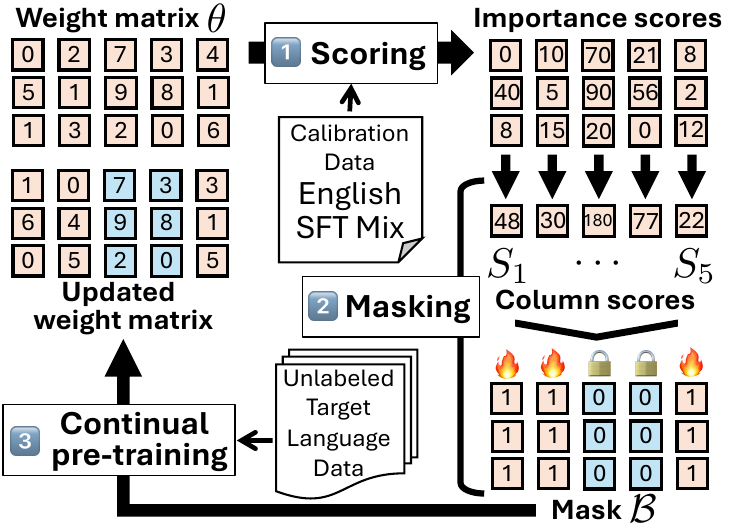}
    \caption{Overview of \textbf{S}ource-\textbf{S}hielded \textbf{U}pdate (\textbf{SSU}). The method comprises three stages: importance scoring, column-wise mask generation, and continual pre-training on unlabeled target data with the masks.}
\label{fig:overview}
\end{figure}

Consequently, using unlabeled target language text is often the only viable option for adaptation.
While this approach can improve target language proficiency, it often triggers catastrophic forgetting~\citep{doi:10.1073/pnas.1611835114,tejaswi-etal-2024-exploring,mundra-etal-2024-empirical,yamaguchi2025elchatadaptingchatlanguage}, where new training erases prior knowledge.
This issue is acute for instruct models, as it cripples the general-purpose functionality of the model, which is primarily derived from core abilities like chat and instruction-following.
In response, previous work has attempted \textbf{post-hoc} mitigation.
For example, \citet{yamaguchi2025elchatadaptingchatlanguage} merge weights of the original and adapted models, while \citet{huang-etal-2024-chat} use a task vector and apply parameter changes from CPT on the base model to the instruct model.
Nonetheless, these methods largely fail to mitigate catastrophic forgetting, substantially degrading these core functionalities.

The shortcomings of post-hoc methods suggest that \textit{mitigation should occur during adaptation.}
We therefore focus on \textbf{the CPT stage}.
Specifically, we leverage selective parameter updates, a method of restricting which weights are modified during training. This approach is proven more effective at mitigating catastrophic forgetting than alternatives like parameter-efficient fine-tuning, regularization, or model merging~\citep{zhang-etal-2024-balancing,hui-etal-2025-hft}.
However, existing selective parameter tuning paradigms for adapting LLMs are ill-suited for adapting instruct models with unlabeled target language text.
They rely either on \textbf{random selection}, offering no principled way to preserve knowledge, or on signals from the new data to guide updates (\textbf{target-focused}) (\S\ref{sec:related_work}).
Target-focused signals are particularly vulnerable because raw text lacks chat templates required to elicit instruction-following behavior.
Optimizing for this incompatible format risks corrupting the very foundational capabilities we aim to preserve due to the structural differences between raw text and chat templates.\looseness=-1

We therefore introduce \textbf{S}ource-\textbf{S}hielded \textbf{U}pdates (\textbf{SSU}), a novel \textbf{source-focused} approach that \textit{proactively shields source knowledge before adaptation begins} (Figure \ref{fig:overview}). First, SSU identifies parameters critical to source abilities using a small set of source data and a parameter importance scoring method, such as those used in model pruning, e.g., Wanda~\citep{sun2024a}.
Second, it uses these element-wise scores to construct a column-wise freezing mask. This structural design is crucial. Unlike naive element-wise freezing that corrupts feature transformations, our column-wise approach preserves them entirely.
Finally, this mask is applied during CPT on unlabeled target language data, keeping the shielded structural units frozen.
This process allows SSU to effectively preserve the general-purpose ability of the model while improving target language performance.

We verify our approach through extensive experiments with five typologically diverse languages and two different model scales (7B and 13B).
We evaluate source language (English) performance across dimensions including chat, instruction-following, safety, and general generation and classification, alongside target language performance.
We summarize our contributions as follows:
\begin{itemize}
    \item A novel method for adapting instruct models to a target language without specialized target instruction-tuning data, addressing a key bottleneck to expand linguistic accessibility.

    \item At two model scales, SSU consistently outperforms all baselines on all core instruction-following and safety tasks. It achieves leading target-language proficiency rivaling full fine-tuning while almost perfectly preserving general source-language performance.
    
    \item Extensive analysis validates the efficacy of SSU, confirming the superiority of column-wise freezing and the importance of source data-driven parameter scoring.
    Qualitatively, we show that SSU avoids the linguistic code-mixing that state-of-the-art methods suffer from, explaining its superior abilities across source chat and instruction-following tasks.
\end{itemize}

\section{Related Work} \label{sec:related_work}
\paragraph{Language Adaptation.}
CPT on target language data is the standard for adapting LLMs~\cite[\textit{inter alia.}]{cui2024efficienteffectivetextencoding,fujii2024continual,yamaguchi-etal-2024-empirical,da-dalt-etal-2024-flor,cahyawijaya-etal-2024-cendol,nguyen-etal-2024-seallms,yamaguchi2024effectivelyexpandvocabularyllms,nag-etal-2025-efficient,ji2025emma500enhancingmassivelymultilingual}. While effective, it often causes catastrophic forgetting, degrading original capabilities~\citep{tejaswi-etal-2024-exploring,mundra-etal-2024-empirical,yamaguchi2025elchatadaptingchatlanguage}.
This trade-off presents a major obstacle for instruct models, where preserving core chat and instruction-following abilities is vital for their general-purpose functionality.\looseness=-1

\paragraph{Catastrophic Forgetting.}
Mitigating catastrophic forgetting is a long-standing challenge in continual learning.
Proposed solutions generally fall into five categories: (1)~\textbf{Regularization} adds a penalty term to the loss function to discourage significant changes to weights deemed important for previous tasks~\citep[\textit{inter alia.}]{doi:10.1073/pnas.1611835114,chen-etal-2020-recall,zhang-etal-2022-clle}.
(2)~\textbf{Replay} interleaves old and new data~\citep[\textit{inter alia.}]{NEURIPS2019_f8d2e80c,NEURIPS2019_fa7cdfad,huang-etal-2024-mitigating,sainz-etal-2025-instructing}.
(3)~\textbf{Model merging} and post-hoc pruning mitigate forgetting by interpolating weights or removing specific task vector updates after fine-tuning~~\citep[\textit{inter alia.}]{pmlr-v162-wortsman22a,NEURIPS2023_1644c9af,pmlr-v235-yu24p,NEURIPS2024_dda5cac5,huang-etal-2025-mitigating}.
(4)~\textbf{Architecture} methods like LoRA~\citep{hu2022lora} train additional new parameters while freezing the original model~\citep[\textit{inter alia.}]{pmlr-v97-houlsby19a,hu2022lora,zhang2023adaptive}.
(5)~\textbf{Selective parameter updates} restrict which existing weights are modified during training~\citep{zhang-etal-2024-balancing,hui-etal-2025-hft}. Our work belongs to this category.\looseness=-1

\paragraph{Selective Parameter Updates.}
While often utilized for training efficiency~\citep{liu2021autofreezeautomaticallyfreezingmodel,lodha-etal-2023-surgical,li2023smartfrz,NEURIPS2024_68716328,NEURIPS2024_6e3b9fb0,pmlr-v235-li24k,ma-etal-2024-sparsity,Li_Zhang_Liu_Gong_Wang_Chen_Cheng_2025,he2025smt}, selective parameter updates have also proven effective for mitigating catastrophic forgetting~\citep{zhang-etal-2024-balancing,hui-etal-2025-hft}. These methods can be broadly categorized as \textbf{dynamic} or \textbf{static}.
Dynamic approaches alter a trainable parameter set during training, based on random selection~\citep{pmlr-v235-li24k,NEURIPS2024_68716328} or target data signals like gradient magnitudes~\citep{liu2021autofreezeautomaticallyfreezingmodel,li2023smartfrz,ma-etal-2024-sparsity,Li_Zhang_Liu_Gong_Wang_Chen_Cheng_2025}.
In contrast, static methods define a fixed set beforehand. This allows for straightforward integration with existing pipelines, enabling the combination of orthogonal mitigation methods like regularization and replay more easily.
For example, a method closest to our work~\citep{hui-etal-2025-hft} randomly freezes components within transformer sub-layers, while others are data-driven based on target data~\citep{lodha-etal-2023-surgical,zhang-etal-2024-balancing,panda2024lotteryticketadaptationmitigating,he2025smt}.

SSU introduces a source-focused static paradigm for language adaptation. Unlike existing methods relying on random choice or target data, SSU uses a small source data sample (e.g., 500 samples) to identify and freeze parameters critical to source knowledge before adaptation.
This proactively shields core abilities, offering a distinct alternative to random or target-data-driven selection criteria.

\section{SSU: Selective Parameter Updates via Importance Freezing}
We address adapting an instruct model using only raw, unlabeled target language data. Unlike prior work that focuses on post-hoc mitigation~\citep{huang-etal-2024-chat,yamaguchi2025elchatadaptingchatlanguage}, Source-Shielded Updates (SSU) targets the CPT process itself.
The goal is to mitigate catastrophic forgetting during CPT, thereby maintaining the general-purpose functionality of an instruct model.
Concurrently, SSU aims to achieve performance gains in the target language tasks comparable to those from full fine-tuning.
Formally, given an instruct model $\mathcal{M}$, calibration data $\mathcal{D}_\text{calib}$, unlabeled target language data $\mathcal{D}_\text{target}$, and a parameter freezing ratio $k$, SSU adapts $\mathcal{M}$ on $\mathcal{D}_\text{target}$ in three stages (Figure \ref{fig:overview}).\looseness=-1

\subsection{Parameter Importance Scoring}
First, SSU scores parameter importance to identify weights critical to source model capabilities.
We posit that a \textbf{source-data-driven score} is suitable, as it directly aligns with the goal of preserving source knowledge. For this purpose, we adopt the importance score from Wanda~\citep{sun2024a}, a popular pruning method.\footnote{While we use Wanda for its simplicity and popularity, \textit{the SSU framework is agnostic to the importance metric.} To demonstrate this, we also evaluate two alternative source-driven scoring methods (\S\ref{sec:analysis}).}
Using a small sample of source data $\mathcal{D}_\text{calib}$, Wanda computes an importance score $s_{ij}$ for each weight $\theta_{ij}$ as the product of its magnitude and the L2-norm of its corresponding input activations $X_j$: $s_{ij} = |\theta_{ij}| \cdot ||X_j||_2$. This identifies weights that are both large and consistently active.
Scores are computed for all parameters in $\mathcal{M}$ except for the embeddings and language modeling head, as all these are updated during training following \citet{hui-etal-2025-hft}.\looseness=-1

\subsection{Column-wise Masking} \label{subsec:freezing}
In the second stage, SSU converts element-wise importance scores into a structured freezing mask. A structured approach is crucial because naive, element-wise freezing disrupts feature transformations and causes catastrophic forgetting (Table \ref{tab:ablation_all}).
To avoid this, SSU operates at the column level.
For instance, in a forward pass $Y=WX$, freezing an entire column of the weight matrix $W$ leaves the corresponding output dimension of $Y$ unchanged, ensuring a complete feature pathway. 
\textit{The approach is analogous to protecting the core structural columns of a building during renovation; the foundational support remains untouched while peripheral elements are modified.}

Mask generation begins by aggregating scores for each column. For a weight matrix $\theta \in \mathbb{R}^{d_{out} \times d_{in}}$, a column corresponds to all parameters associated with a single input feature.
The total importance score $S_j$ for each column $j$ is the sum of its individual scores: $S_j = \sum_{i} s_{ij}$.
$S_j$ robustly measures the contribution of each input feature, identifying the core structural columns to be preserved.
For 1D parameters, such as biases, each element is treated as its own column; thus, its per-weight score $s_i$ serves as its aggregated score $S_i$.\looseness=-1

The binary mask $\mathcal{B}$ for each weight matrix is generated by  ranking columns by their $S_j$ and then selecting the top $k$\% to freeze (50\% by default following \citet{hui-etal-2025-hft}).
The corresponding columns in the mask $\mathcal{B}$ are set to 0 (freeze), while all others are set to 1 (update).

\subsection{Continual Pre-training}
In the third stage, the model $\mathcal{M}$ is continually pre-trained on unlabeled data $\mathcal{D}_\text{target}$ using a standard causal language modeling objective, denoted as the loss $L$.
During the backward pass, the static mask $\mathcal{B}$ is applied to the gradients, zeroing out updates for frozen columns. The gradient update rule for a weight $\theta_{ij}$ is thus $\theta_{ij} \leftarrow \theta_{ij} - \eta \cdot b_{ij} \cdot \nabla_{\theta_{ij}}L$. Here, $\eta$ is the learning rate, and $b_{ij} \in \{0, 1\}$ is the value from the mask $\mathcal{B}$ corresponding to the weight $\theta_{ij}$.
This method preserves knowledge stored in the most critical input-feature pathways, thus mitigating catastrophic forgetting.

\section{Experimental Setup}

\subsection{Source Models}
Following \citet{hui-etal-2025-hft} who used 7B and 13B models from the same family (i.e., Llama 2), we use the 7B and 13B OLMo 2 Instruct models~\citep{olmo20252olmo2furious} for our experiments.
The OLMo 2 models offer strong instruction-following capabilities and fully documented training data, allowing full control and transparency in our language adaptation experiments.\footnote{While our main experiments use OLMo 2, we find that the findings generalize to OLMo 3 (see Appendix \ref{appendix:olmo3}).}

\subsection{Target Languages}
\begin{table}[t]
    \centering
    \renewcommand{\arraystretch}{0.89}
    \small
    \resizebox{\linewidth}{!}{%
    \begin{tabular}{ccccc}
    \toprule
      \textbf{Language} & \textbf{Code} & \textbf{Script} & \textbf{Family} & \textbf{CC Ratio}\\
    \midrule
       English  & en & Latin & Indo-European & 43.7876\\
    \midrule
       Nepali & ne & Devanagari & Indo-European & .0521\\
       Kyrgyz  & ky & Cyrillic & Turkic & .0103\\
       Amharic  & am & Ge'ez & Afro-Asiatic & .0032\\
       Hausa  & ha & Latin & Afro-Asiatic & .0032\\
       Igbo  & ig & Latin & Niger-Congo & .0007\\
    \bottomrule
    \end{tabular}
    }
    \caption{Source (English) and target languages. Code is based on ISO 639-1, and the language-specific ratio in Common Crawl (CC Ratio) as of CC-MAIN-2025-21.} %
    \label{tab:language}
\end{table}

We experiment with five typologically diverse languages (Table \ref{tab:language}) that are significantly underrepresented in the training data of the source models but with wide availability of datasets with consistent task formulations (though data variations preclude direct performance comparisons between languages).
These languages appear at least 840x less frequently than English in Common Crawl (CC),\footnote{CC Ratio is from \href{https://commoncrawl.github.io/cc-crawl-statistics/plots/languages}{the Statistics of CC Monthly Archives}.} which accounts for over 95\% of the OLMo 2 pre-training corpus~\citep{olmo20252olmo2furious}.

\subsection{Calibration and Training Data} \label{subsec:data}
We use tulu-3-sft-olmo-2-mixture~\citep{lambert2025tulu3pushingfrontiers}, the original instruction-tuning data for OLMo 2, for calibration (i.e., choosing which parameters to freeze).
We randomly select 500 samples with a sequence length of 2,048. For CPT, we use a clean subset of MADLAD-400~\citep{kudugunta2023madlad}, sampling 200M tokens per language as recommended by \citet{tejaswi-etal-2024-exploring}.\footnote{During CPT, we remove the chat template to support unlabeled data lacking role annotations (e.g., \texttt{user}).}

\subsection{Baselines}
We compare our approach against baselines from three categories: performance benchmarks, a reference approach from a related paradigm, and state-of-the-art methods.

\paragraph{Source.} Off-the-shelf OLMo 2, reporting performance without any adaptation.
    
\paragraph{FFT.} Full fine-tuning that updates all the parameters via CPT on target language data, quantifying the extent to which a model suffers from catastrophic forgetting without any intervention.

\paragraph{AdaLoRA.}~\citep{zhang2023adaptive}: An architecture-based method to mitigate catastrophic forgetting. This achieves the best overall performance among LoRA-like methods in \citet{hui-etal-2025-hft}.

\paragraph{HFT.} A state-of-the-art \textbf{static} selective parameter update method~\citep{hui-etal-2025-hft}.
It updates 50\% of parameters by randomly freezing two out of the four self-attention matrices ($W_Q, W_K, W_V, W_O$); and two out of three feed-forward matrices ($W_{up}, W_{down}, W_{gate}$) in a random half of the layers and one matrix in the remaining half. Since SSU is also a static method, HFT serves as a key baseline.

\paragraph{GMT.} 
A state-of-the-art \textbf{dynamic} selective parameter update approach~\citep{Li_Zhang_Liu_Gong_Wang_Chen_Cheng_2025} that drops gradients of a pre-defined ratio (50\% in this study for fair comparison with HFT and SSU) with smaller absolute values on the target data.

To validate our use of source calibration data for scoring, we also introduce two calibration data-free ablation variants: (1) \textbf{SSU-Rand} that freezes an equal number of randomly-selected columns. This provides no principled way to preserve functionally important knowledge. (2) \textbf{SSU-Mag} that freezes columns based only on the magnitude score (i.e., $|\theta_{ij}|$; unlike $|\theta_{ij}|\cdot||X_j||_2$ for SSU-Wanda), isolating the effect of the activation term.

We further compare SSU against other recent selective parameter update methods proposed for LLM adaptation, \textbf{LoTA}~\citep{panda2024lotteryticketadaptationmitigating} and \textbf{S2FT}~\citep{NEURIPS2024_6e3b9fb0}, in Appendix \ref{appendix:additional_baselines}. We find that only SSU achieves consistently both strong source preservation and high target gains.

\subsection{Evaluation Benchmarks and Metrics}
\label{subsec:metrics}
We report performance in the source and target languages across standard benchmarks. 

\paragraph{Chat and Instruction-following.}
We report (1) \textbf{IFEval}~\citep{zhou2023instructionfollowingevaluationlargelanguage} zero-shot accuracy (strict prompt);
(2) \textbf{AlpacaEval 2.0} (AE2)~\citep{alpaca_eval} length-controlled win-rate against GPT-4 (1106-preview)~\citep{openai2024gpt4technicalreport}; and
(3) \textbf{MT-Bench} (MTB)~\citep{zheng2023judging} mean Likert-5 score over two turns;
(4) \textbf{GSM8K}~\citep{cobbe2021trainingverifierssolvemath} five-shot exact match for multi-turn mathematical reasoning.

\paragraph{Safety.} We use the Tülu 3 safety evaluation suite~\citep[\textbf{T3}]{lambert2025tulu3pushingfrontiers}. We report the macro average score in a zero-shot setting, following \citet{lambert2025tulu3pushingfrontiers} and \citet{olmo20252olmo2furious}.\footnote{As instruct models typically undergo extensive safety alignment~\citep[\textit{inter alia.}]{gemmateam2025gemma3technicalreport,lambert2025tulu3pushingfrontiers}, verifying that this is not compromised during adaptation is a crucial aspect of our analysis.}

\begin{table*}[t]
\centering
\small
\renewcommand{\arraystretch}{0.89}
\setlength{\aboverulesep}{2.25pt}
\setlength{\belowrulesep}{2.25pt}
\resizebox{\textwidth}{!}{
\begin{tabular}{@{}ll@{\hspace{5pt}}l@{\hspace{6pt}}l@{\hspace{6pt}}l@{\hspace{6pt}}l@{\hspace{6pt}}l@{\hspace{6pt}}l@{\hspace{6pt}}l@{\hspace{6pt}}l@{\hspace{6pt}}l@{\hspace{6pt}}l@{\hspace{6pt}}l@{\hspace{6pt}}l@{\hspace{6pt}}l}
\toprule
& & \multicolumn{4}{c}{\textbf{Chat and Instruction-following} (en)} & \multicolumn{1}{c}{\textbf{Safety}} & \multicolumn{4}{c}{\textbf{Source language} (en)} & \multicolumn{4}{c}{\textbf{Target language}} \\
\cmidrule(lr){3-6} \cmidrule(lr){8-11} \cmidrule(lr){12-15}
\multicolumn{2}{l}{\textbf{Approach}} & \multicolumn{1}{c}{\textbf{IFEval}} & \multicolumn{1}{c}{\textbf{AE2}} & \multicolumn{1}{c}{\textbf{MTB}} & \multicolumn{1}{c}{\textbf{GSM8K}} & \multicolumn{1}{c}{\textbf{T3} (en)} & \multicolumn{1}{c}{\textbf{MT}} & \multicolumn{1}{c}{\textbf{SUM}} & \multicolumn{1}{c}{\textbf{MRC}} & \multicolumn{1}{c}{\textbf{MMLU}} & \multicolumn{1}{c}{\textbf{MT}} & \multicolumn{1}{c}{\textbf{SUM}} & \multicolumn{1}{c}{\textbf{MRC}} & \multicolumn{1}{c}{\textbf{MMLU}} \\
\midrule

\multirow{12}{*}{\rotatebox{90}{7B}} & \cellcolor{gray!20}\raisebox{-0.5ex}{Source} & \cellcolor{gray!20}.675\raisebox{-1.25ex}{\hspace{0.1em}{\scriptsize +0.0}} & \cellcolor{gray!20}32.6\raisebox{-1.25ex}{\hspace{0.1em}{\scriptsize +0.0}} & \cellcolor{gray!20}3.98\raisebox{-1.25ex}{\hspace{0.1em}{\scriptsize +0.0}} & \cellcolor{gray!20}.796\raisebox{-1.25ex}{\hspace{0.1em}{\scriptsize +0.0}} & \cellcolor{gray!20}.851\raisebox{-1.25ex}{\hspace{0.1em}{\scriptsize +0.0}} & \cellcolor{gray!20}30.0\raisebox{-1.25ex}{\hspace{0.1em}{\scriptsize +0.0}} & \cellcolor{gray!20}22.8\raisebox{-1.25ex}{\hspace{0.1em}{\scriptsize +0.0}} & \cellcolor{gray!20}.880\raisebox{-1.25ex}{\hspace{0.1em}{\scriptsize +0.0}} & \cellcolor{gray!20}.618\raisebox{-1.25ex}{\hspace{0.1em}{\scriptsize +0.0}} & \cellcolor{gray!20}20.1\raisebox{-1.25ex}{\hspace{0.1em}{\scriptsize +0.0}} & \cellcolor{gray!20}20.2\raisebox{-1.25ex}{\hspace{0.1em}{\scriptsize +0.0}} & \cellcolor{gray!20}.334\raisebox{-1.25ex}{\hspace{0.1em}{\scriptsize +0.0}} & \cellcolor{gray!20}.304\raisebox{-1.25ex}{\hspace{0.1em}{\scriptsize +0.0}} \\[5pt] \cmidrule{2-15}
 & \raisebox{-0.5ex}{FFT} & .456\raisebox{-1.25ex}{\hspace{0.1em}{\scriptsize -32.4}} & 10.4\raisebox{-1.25ex}{\hspace{0.1em}{\scriptsize -68.1}} & 3.48\raisebox{-1.25ex}{\hspace{0.1em}{\scriptsize -12.5}} & .608\raisebox{-1.25ex}{\hspace{0.1em}{\scriptsize -23.6}} & .797\raisebox{-1.25ex}{\hspace{0.1em}{\scriptsize -6.4}} & \cellcolor{green!20}42.8\raisebox{-1.25ex}{\hspace{0.1em}{\scriptsize +42.6}} & 20.8\raisebox{-1.25ex}{\hspace{0.1em}{\scriptsize -8.7}} & .842\raisebox{-1.25ex}{\hspace{0.1em}{\scriptsize -4.3}} & .580\raisebox{-1.25ex}{\hspace{0.1em}{\scriptsize -6.2}} & \cellcolor{green!20}30.7\raisebox{-1.25ex}{\hspace{0.1em}{\scriptsize +52.8}} & \cellcolor{green!20}22.7\raisebox{-1.25ex}{\hspace{0.1em}{\scriptsize +12.4}} & \cellcolor{green!20}.393\raisebox{-1.25ex}{\hspace{0.1em}{\scriptsize +17.7}} & \underline{\cellcolor{green!20}.325}\raisebox{-1.25ex}{\hspace{0.1em}{\scriptsize +6.8}} \\[5pt]
 & \raisebox{-0.5ex}{AdaLoRA} & \textbf{.669}\raisebox{-1.25ex}{\hspace{0.1em}{\scriptsize -0.8}} & \underline{24.6}\raisebox{-1.25ex}{\hspace{0.1em}{\scriptsize -24.5}} & \underline{3.92}\raisebox{-1.25ex}{\hspace{0.1em}{\scriptsize -1.5}} & \underline{.721}\raisebox{-1.25ex}{\hspace{0.1em}{\scriptsize -9.4}} & .824\raisebox{-1.25ex}{\hspace{0.1em}{\scriptsize -3.2}} & \cellcolor{green!20}34.1\raisebox{-1.25ex}{\hspace{0.1em}{\scriptsize +13.6}} & \underline{22.4}\raisebox{-1.25ex}{\hspace{0.1em}{\scriptsize -1.6}} & \underline{.866}\raisebox{-1.25ex}{\hspace{0.1em}{\scriptsize -1.6}} & \underline{.602}\raisebox{-1.25ex}{\hspace{0.1em}{\scriptsize -2.6}} & 19.9\raisebox{-1.25ex}{\hspace{0.1em}{\scriptsize -1.0}} & \cellcolor{green!20}21.9\raisebox{-1.25ex}{\hspace{0.1em}{\scriptsize +8.4}} & .318\raisebox{-1.25ex}{\hspace{0.1em}{\scriptsize -4.8}} & .299\raisebox{-1.25ex}{\hspace{0.1em}{\scriptsize -1.8}} \\[5pt]
 & \raisebox{-0.5ex}{HFT} & \underline{.621}\raisebox{-1.25ex}{\hspace{0.1em}{\scriptsize -8.0}} & 17.6\raisebox{-1.25ex}{\hspace{0.1em}{\scriptsize -45.9}} & 3.83\raisebox{-1.25ex}{\hspace{0.1em}{\scriptsize -3.7}} & .677\raisebox{-1.25ex}{\hspace{0.1em}{\scriptsize -15.0}} & .826\raisebox{-1.25ex}{\hspace{0.1em}{\scriptsize -3.0}} & \cellcolor{green!20}45.2\raisebox{-1.25ex}{\hspace{0.1em}{\scriptsize +50.6}} & 22.3\raisebox{-1.25ex}{\hspace{0.1em}{\scriptsize -2.1}} & .854\raisebox{-1.25ex}{\hspace{0.1em}{\scriptsize -3.0}} & .595\raisebox{-1.25ex}{\hspace{0.1em}{\scriptsize -3.7}} & \cellcolor{green!20}29.8\raisebox{-1.25ex}{\hspace{0.1em}{\scriptsize +48.3}} & \cellcolor{green!20}22.6\raisebox{-1.25ex}{\hspace{0.1em}{\scriptsize +11.9}} & \cellcolor{green!20}.377\raisebox{-1.25ex}{\hspace{0.1em}{\scriptsize +12.9}} & \cellcolor{green!20}.322\raisebox{-1.25ex}{\hspace{0.1em}{\scriptsize +5.8}} \\[5pt]
 & \raisebox{-0.5ex}{GMT} & .528\raisebox{-1.25ex}{\hspace{0.1em}{\scriptsize -21.7}} & 12.5\raisebox{-1.25ex}{\hspace{0.1em}{\scriptsize -61.6}} & 3.67\raisebox{-1.25ex}{\hspace{0.1em}{\scriptsize -7.7}} & .635\raisebox{-1.25ex}{\hspace{0.1em}{\scriptsize -20.2}} & .795\raisebox{-1.25ex}{\hspace{0.1em}{\scriptsize -6.6}} & \underline{\cellcolor{green!20}45.5}\raisebox{-1.25ex}{\hspace{0.1em}{\scriptsize +51.6}} & 21.6\raisebox{-1.25ex}{\hspace{0.1em}{\scriptsize -5.1}} & .841\raisebox{-1.25ex}{\hspace{0.1em}{\scriptsize -4.4}} & .582\raisebox{-1.25ex}{\hspace{0.1em}{\scriptsize -5.8}} & \underline{\cellcolor{green!20}30.9}\raisebox{-1.25ex}{\hspace{0.1em}{\scriptsize +53.8}} & \textbf{\cellcolor{green!20}22.9}\raisebox{-1.25ex}{\hspace{0.1em}{\scriptsize +13.4}} & \cellcolor{green!20}.385\raisebox{-1.25ex}{\hspace{0.1em}{\scriptsize +15.3}} & \cellcolor{green!20}.319\raisebox{-1.25ex}{\hspace{0.1em}{\scriptsize +4.8}} \\[5pt] \cmidrule{2-15}
 & \raisebox{-0.5ex}{SSU-Rand} & .608\raisebox{-1.25ex}{\hspace{0.1em}{\scriptsize -9.9}} & 18.0\raisebox{-1.25ex}{\hspace{0.1em}{\scriptsize -44.7}} & 3.81\raisebox{-1.25ex}{\hspace{0.1em}{\scriptsize -4.2}} & .683\raisebox{-1.25ex}{\hspace{0.1em}{\scriptsize -14.2}} & \underline{.835}\raisebox{-1.25ex}{\hspace{0.1em}{\scriptsize -1.9}} & \underline{\cellcolor{green!20}45.5}\raisebox{-1.25ex}{\hspace{0.1em}{\scriptsize +51.6}} & \underline{22.4}\raisebox{-1.25ex}{\hspace{0.1em}{\scriptsize -1.6}} & .861\raisebox{-1.25ex}{\hspace{0.1em}{\scriptsize -2.2}} & .597\raisebox{-1.25ex}{\hspace{0.1em}{\scriptsize -3.4}} & \cellcolor{green!20}30.2\raisebox{-1.25ex}{\hspace{0.1em}{\scriptsize +50.3}} & \cellcolor{green!20}22.7\raisebox{-1.25ex}{\hspace{0.1em}{\scriptsize +12.4}} & \underline{\cellcolor{green!20}.394}\raisebox{-1.25ex}{\hspace{0.1em}{\scriptsize +18.0}} & \cellcolor{green!20}.324\raisebox{-1.25ex}{\hspace{0.1em}{\scriptsize +6.4}} \\[5pt]
 & \raisebox{-0.5ex}{SSU-Mag} & .570\raisebox{-1.25ex}{\hspace{0.1em}{\scriptsize -15.5}} & 14.9\raisebox{-1.25ex}{\hspace{0.1em}{\scriptsize -54.2}} & 3.78\raisebox{-1.25ex}{\hspace{0.1em}{\scriptsize -5.0}} & .655\raisebox{-1.25ex}{\hspace{0.1em}{\scriptsize -17.7}} & .822\raisebox{-1.25ex}{\hspace{0.1em}{\scriptsize -3.4}} & \cellcolor{green!20}44.7\raisebox{-1.25ex}{\hspace{0.1em}{\scriptsize +48.9}} & 22.0\raisebox{-1.25ex}{\hspace{0.1em}{\scriptsize -3.4}} & .859\raisebox{-1.25ex}{\hspace{0.1em}{\scriptsize -2.4}} & .593\raisebox{-1.25ex}{\hspace{0.1em}{\scriptsize -4.1}} & \cellcolor{green!20}29.7\raisebox{-1.25ex}{\hspace{0.1em}{\scriptsize +47.8}} & \cellcolor{green!20}22.7\raisebox{-1.25ex}{\hspace{0.1em}{\scriptsize +12.4}} & \cellcolor{green!20}.383\raisebox{-1.25ex}{\hspace{0.1em}{\scriptsize +14.7}} & \cellcolor{green!20}.319\raisebox{-1.25ex}{\hspace{0.1em}{\scriptsize +4.8}} \\[5pt] \noalign{\vskip\aboverulesep}\cdashline{2-15}[2pt/1.2pt]\noalign{\vskip\belowrulesep}
 & \raisebox{-0.5ex}{SSU-Wanda} & \textbf{.669}\raisebox{-1.25ex}{\hspace{0.1em}{\scriptsize -0.8}} & \textbf{27.0}\raisebox{-1.25ex}{\hspace{0.1em}{\scriptsize -17.1}} & \textbf{3.96}\raisebox{-1.25ex}{\hspace{0.1em}{\scriptsize -0.5}} & \textbf{.752}\raisebox{-1.25ex}{\hspace{0.1em}{\scriptsize -5.5}} & \textbf{.850}\raisebox{-1.25ex}{\hspace{0.1em}{\scriptsize -0.1}} & \textbf{\cellcolor{green!20}45.7}\raisebox{-1.25ex}{\hspace{0.1em}{\scriptsize +52.3}} & \textbf{\cellcolor{green!20}22.8}\raisebox{-1.25ex}{\hspace{0.1em}{\scriptsize +0.1}} & \textbf{.869}\raisebox{-1.25ex}{\hspace{0.1em}{\scriptsize -1.3}} & \textbf{.606}\raisebox{-1.25ex}{\hspace{0.1em}{\scriptsize -2.0}} & \textbf{\cellcolor{green!20}31.0}\raisebox{-1.25ex}{\hspace{0.1em}{\scriptsize +54.3}} & \underline{\cellcolor{green!20}22.8}\raisebox{-1.25ex}{\hspace{0.1em}{\scriptsize +12.9}} & \textbf{\cellcolor{green!20}.403}\raisebox{-1.25ex}{\hspace{0.1em}{\scriptsize +20.7}} & \textbf{\cellcolor{green!20}.333}\raisebox{-1.25ex}{\hspace{0.1em}{\scriptsize +9.4}} \\[5pt]
\midrule
\multirow{12}{*}{\rotatebox{90}{13B}} & \cellcolor{gray!20}\raisebox{-0.5ex}{Source} & \cellcolor{gray!20}.763\raisebox{-1.25ex}{\hspace{0.1em}{\scriptsize +0.0}} & \cellcolor{gray!20}37.2\raisebox{-1.25ex}{\hspace{0.1em}{\scriptsize +0.0}} & \cellcolor{gray!20}4.06\raisebox{-1.25ex}{\hspace{0.1em}{\scriptsize +0.0}} & \cellcolor{gray!20}.853\raisebox{-1.25ex}{\hspace{0.1em}{\scriptsize +0.0}} & \cellcolor{gray!20}.821\raisebox{-1.25ex}{\hspace{0.1em}{\scriptsize +0.0}} & \cellcolor{gray!20}33.3\raisebox{-1.25ex}{\hspace{0.1em}{\scriptsize +0.0}} & \cellcolor{gray!20}24.5\raisebox{-1.25ex}{\hspace{0.1em}{\scriptsize +0.0}} & \cellcolor{gray!20}.897\raisebox{-1.25ex}{\hspace{0.1em}{\scriptsize +0.0}} & \cellcolor{gray!20}.665\raisebox{-1.25ex}{\hspace{0.1em}{\scriptsize +0.0}} & \cellcolor{gray!20}22.4\raisebox{-1.25ex}{\hspace{0.1em}{\scriptsize +0.0}} & \cellcolor{gray!20}20.7\raisebox{-1.25ex}{\hspace{0.1em}{\scriptsize +0.0}} & \cellcolor{gray!20}.374\raisebox{-1.25ex}{\hspace{0.1em}{\scriptsize +0.0}} & \cellcolor{gray!20}.329\raisebox{-1.25ex}{\hspace{0.1em}{\scriptsize +0.0}} \\[5pt] \cmidrule{2-15}
 & \raisebox{-0.5ex}{FFT} & .448\raisebox{-1.25ex}{\hspace{0.1em}{\scriptsize -41.3}} & 14.5\raisebox{-1.25ex}{\hspace{0.1em}{\scriptsize -61.1}} & 3.52\raisebox{-1.25ex}{\hspace{0.1em}{\scriptsize -13.3}} & .740\raisebox{-1.25ex}{\hspace{0.1em}{\scriptsize -13.3}} & .737\raisebox{-1.25ex}{\hspace{0.1em}{\scriptsize -10.2}} & \cellcolor{green!20}40.1\raisebox{-1.25ex}{\hspace{0.1em}{\scriptsize +20.3}} & 15.7\raisebox{-1.25ex}{\hspace{0.1em}{\scriptsize -35.8}} & .892\raisebox{-1.25ex}{\hspace{0.1em}{\scriptsize -0.5}} & .647\raisebox{-1.25ex}{\hspace{0.1em}{\scriptsize -2.7}} & \cellcolor{green!20}33.6\raisebox{-1.25ex}{\hspace{0.1em}{\scriptsize +50.1}} & \cellcolor{green!20}22.9\raisebox{-1.25ex}{\hspace{0.1em}{\scriptsize +10.4}} & \textbf{\cellcolor{green!20}.492}\raisebox{-1.25ex}{\hspace{0.1em}{\scriptsize +31.6}} & \textbf{\cellcolor{green!20}.361}\raisebox{-1.25ex}{\hspace{0.1em}{\scriptsize +9.8}} \\[5pt]
 & \raisebox{-0.5ex}{AdaLoRA} & \underline{.719}\raisebox{-1.25ex}{\hspace{0.1em}{\scriptsize -5.8}} & \underline{32.1}\raisebox{-1.25ex}{\hspace{0.1em}{\scriptsize -13.8}} & \textbf{4.05}\raisebox{-1.25ex}{\hspace{0.1em}{\scriptsize -0.2}} & \underline{.815}\raisebox{-1.25ex}{\hspace{0.1em}{\scriptsize -4.5}} & \underline{.799}\raisebox{-1.25ex}{\hspace{0.1em}{\scriptsize -2.7}} & \cellcolor{green!20}36.6\raisebox{-1.25ex}{\hspace{0.1em}{\scriptsize +9.8}} & \textbf{24.4}\raisebox{-1.25ex}{\hspace{0.1em}{\scriptsize -0.2}} & \textbf{\cellcolor{green!20}.898}\raisebox{-1.25ex}{\hspace{0.1em}{\scriptsize +0.1}} & \underline{.660}\raisebox{-1.25ex}{\hspace{0.1em}{\scriptsize -0.8}} & \cellcolor{green!20}23.0\raisebox{-1.25ex}{\hspace{0.1em}{\scriptsize +2.7}} & \cellcolor{green!20}22.3\raisebox{-1.25ex}{\hspace{0.1em}{\scriptsize +7.5}} & .365\raisebox{-1.25ex}{\hspace{0.1em}{\scriptsize -2.4}} & .311\raisebox{-1.25ex}{\hspace{0.1em}{\scriptsize -5.4}} \\[5pt]
 & \raisebox{-0.5ex}{HFT} & .631\raisebox{-1.25ex}{\hspace{0.1em}{\scriptsize -17.3}} & 25.8\raisebox{-1.25ex}{\hspace{0.1em}{\scriptsize -30.7}} & \underline{3.92}\raisebox{-1.25ex}{\hspace{0.1em}{\scriptsize -3.4}} & .776\raisebox{-1.25ex}{\hspace{0.1em}{\scriptsize -9.0}} & .785\raisebox{-1.25ex}{\hspace{0.1em}{\scriptsize -4.4}} & \underline{\cellcolor{green!20}44.1}\raisebox{-1.25ex}{\hspace{0.1em}{\scriptsize +32.2}} & 20.7\raisebox{-1.25ex}{\hspace{0.1em}{\scriptsize -15.3}} & .894\raisebox{-1.25ex}{\hspace{0.1em}{\scriptsize -0.3}} & .658\raisebox{-1.25ex}{\hspace{0.1em}{\scriptsize -1.1}} & \underline{\cellcolor{green!20}33.7}\raisebox{-1.25ex}{\hspace{0.1em}{\scriptsize +50.5}} & \cellcolor{green!20}22.8\raisebox{-1.25ex}{\hspace{0.1em}{\scriptsize +9.9}} & \cellcolor{green!20}.476\raisebox{-1.25ex}{\hspace{0.1em}{\scriptsize +27.3}} & \cellcolor{green!20}.355\raisebox{-1.25ex}{\hspace{0.1em}{\scriptsize +8.0}} \\[5pt]
 & \raisebox{-0.5ex}{GMT} & .497\raisebox{-1.25ex}{\hspace{0.1em}{\scriptsize -34.9}} & 19.3\raisebox{-1.25ex}{\hspace{0.1em}{\scriptsize -48.2}} & 3.64\raisebox{-1.25ex}{\hspace{0.1em}{\scriptsize -10.3}} & .754\raisebox{-1.25ex}{\hspace{0.1em}{\scriptsize -11.6}} & .755\raisebox{-1.25ex}{\hspace{0.1em}{\scriptsize -8.0}} & \cellcolor{green!20}37.5\raisebox{-1.25ex}{\hspace{0.1em}{\scriptsize +12.5}} & 16.5\raisebox{-1.25ex}{\hspace{0.1em}{\scriptsize -32.5}} & .896\raisebox{-1.25ex}{\hspace{0.1em}{\scriptsize -0.1}} & .654\raisebox{-1.25ex}{\hspace{0.1em}{\scriptsize -1.7}} & \cellcolor{green!20}33.5\raisebox{-1.25ex}{\hspace{0.1em}{\scriptsize +49.6}} & \cellcolor{green!20}22.8\raisebox{-1.25ex}{\hspace{0.1em}{\scriptsize +9.9}} & \cellcolor{green!20}.473\raisebox{-1.25ex}{\hspace{0.1em}{\scriptsize +26.5}} & \cellcolor{green!20}.353\raisebox{-1.25ex}{\hspace{0.1em}{\scriptsize +7.4}} \\[5pt] \cmidrule{2-15}
 & \raisebox{-0.5ex}{SSU-Rand} & .630\raisebox{-1.25ex}{\hspace{0.1em}{\scriptsize -17.5}} & 24.7\raisebox{-1.25ex}{\hspace{0.1em}{\scriptsize -33.7}} & 3.89\raisebox{-1.25ex}{\hspace{0.1em}{\scriptsize -4.1}} & .781\raisebox{-1.25ex}{\hspace{0.1em}{\scriptsize -8.5}} & .783\raisebox{-1.25ex}{\hspace{0.1em}{\scriptsize -4.6}} & \cellcolor{green!20}43.9\raisebox{-1.25ex}{\hspace{0.1em}{\scriptsize +31.6}} & 21.7\raisebox{-1.25ex}{\hspace{0.1em}{\scriptsize -11.3}} & \textbf{\cellcolor{green!20}.898}\raisebox{-1.25ex}{\hspace{0.1em}{\scriptsize +0.1}} & .656\raisebox{-1.25ex}{\hspace{0.1em}{\scriptsize -1.4}} & \cellcolor{green!20}33.6\raisebox{-1.25ex}{\hspace{0.1em}{\scriptsize +50.1}} & \underline{\cellcolor{green!20}23.0}\raisebox{-1.25ex}{\hspace{0.1em}{\scriptsize +10.9}} & \cellcolor{green!20}.478\raisebox{-1.25ex}{\hspace{0.1em}{\scriptsize +27.8}} & \cellcolor{green!20}.356\raisebox{-1.25ex}{\hspace{0.1em}{\scriptsize +8.3}} \\[5pt]
 & \raisebox{-0.5ex}{SSU-Mag} & .572\raisebox{-1.25ex}{\hspace{0.1em}{\scriptsize -25.1}} & 20.6\raisebox{-1.25ex}{\hspace{0.1em}{\scriptsize -44.7}} & 3.80\raisebox{-1.25ex}{\hspace{0.1em}{\scriptsize -6.4}} & .763\raisebox{-1.25ex}{\hspace{0.1em}{\scriptsize -10.6}} & .776\raisebox{-1.25ex}{\hspace{0.1em}{\scriptsize -5.5}} & \cellcolor{green!20}40.2\raisebox{-1.25ex}{\hspace{0.1em}{\scriptsize +20.6}} & 20.2\raisebox{-1.25ex}{\hspace{0.1em}{\scriptsize -17.4}} & .892\raisebox{-1.25ex}{\hspace{0.1em}{\scriptsize -0.5}} & .657\raisebox{-1.25ex}{\hspace{0.1em}{\scriptsize -1.2}} & \cellcolor{green!20}32.8\raisebox{-1.25ex}{\hspace{0.1em}{\scriptsize +46.5}} & \cellcolor{green!20}22.6\raisebox{-1.25ex}{\hspace{0.1em}{\scriptsize +8.9}} & \cellcolor{green!20}.467\raisebox{-1.25ex}{\hspace{0.1em}{\scriptsize +24.9}} & \cellcolor{green!20}.350\raisebox{-1.25ex}{\hspace{0.1em}{\scriptsize +6.5}} \\[5pt] \noalign{\vskip\aboverulesep}\cdashline{2-15}[2pt/1.2pt]\noalign{\vskip\belowrulesep}
 & \raisebox{-0.5ex}{SSU-Wanda} & \textbf{.730}\raisebox{-1.25ex}{\hspace{0.1em}{\scriptsize -4.4}} & \textbf{33.4}\raisebox{-1.25ex}{\hspace{0.1em}{\scriptsize -10.3}} & \textbf{4.05}\raisebox{-1.25ex}{\hspace{0.1em}{\scriptsize -0.2}} & \textbf{.822}\raisebox{-1.25ex}{\hspace{0.1em}{\scriptsize -3.7}} & \textbf{.805}\raisebox{-1.25ex}{\hspace{0.1em}{\scriptsize -2.0}} & \textbf{\cellcolor{green!20}48.2}\raisebox{-1.25ex}{\hspace{0.1em}{\scriptsize +44.5}} & \underline{24.2}\raisebox{-1.25ex}{\hspace{0.1em}{\scriptsize -1.0}} & \underline{\cellcolor{green!20}.897}\raisebox{-1.25ex}{\hspace{0.1em}{\scriptsize +0.0}} & \textbf{.661}\raisebox{-1.25ex}{\hspace{0.1em}{\scriptsize -0.6}} & \textbf{\cellcolor{green!20}34.1}\raisebox{-1.25ex}{\hspace{0.1em}{\scriptsize +52.3}} & \textbf{\cellcolor{green!20}23.2}\raisebox{-1.25ex}{\hspace{0.1em}{\scriptsize +11.8}} & \underline{\cellcolor{green!20}.486}\raisebox{-1.25ex}{\hspace{0.1em}{\scriptsize +29.9}} & \underline{\cellcolor{green!20}.359}\raisebox{-1.25ex}{\hspace{0.1em}{\scriptsize +9.2}} \\[5pt]
\bottomrule
\end{tabular}
}
\caption{Aggregated average performance across all languages per task.
\colorbox{green!20}{Green} denotes scores better than Source with subscripts showing relative changes (\%).
\textbf{Bold} and \underline{underlined} indicate best and second-best methods for each model scale.
Tables~\ref{tab:source_instruct}, \ref{tab:safety}, \ref{tab:source}, and \ref{tab:target} include a full suite of results.}
\label{tab:aggregated_with_deltas}
\end{table*}

\paragraph{Source Language {\rm (}English{\rm )}.}
We evaluate target-to-English machine translation (\textbf{MT}) on FLORES-200~\citep{nllb-22}, reporting three-shot chrF++~\citep{popovic-2017-chrf} on 500 samples, following previous work~\citep{ahia-etal-2023-languages,yamaguchi2025elchatadaptingchatlanguage}. 
For summarization (\textbf{SUM}) on XL-SUM~\citep{hasan-etal-2021-xl}, we use zero-shot chrF++ on 500 samples.
For machine reading comprehension (\textbf{MRC}) on Belebele~\citep{bandarkar-etal-2024-belebele} and general reasoning on \textbf{MMLU}~\citep{hendrycks2021measuring}, we report three-shot and five-shot accuracy, respectively, on their test sets.

\paragraph{Target Language.} We evaluate English-to-target \textbf{MT}, \textbf{SUM}, and \textbf{MRC} on the same target-language subsets of respective datasets and settings. For reasoning, we use Global MMLU~\citep{singh-etal-2025-global} and report five-shot accuracy on its test set.

We report average scores over three runs for generative tasks and use a single deterministic run with temperature zero for classification tasks.
Further details (e.g., prompt templates) are in Appendix \ref{appendix:evaluation_setup}.

\section{Results} \label{sec:results}

Table \ref{tab:aggregated_with_deltas} shows performance across the four task groups: chat and instruction-following, safety, source language, and target language.

\paragraph{Chat and Instruction-following.}
Our SSU-Wanda achieves the best performance on all chat and instruction-following benchmarks, exhibiting the smallest average relative performance drops from Source of 5.9\% (7B) and 4.7\% (13B).
This result is particularly important as these tasks directly measure core instruct model capabilities, such as multi-step reasoning and following complex constraints. The performance of SSU-Wanda demonstrates its efficacy in retaining source knowledge and abilities.
The architecture-based method, AdaLoRA, performs second best with average degradations of 9.0\% (7B) and 6.1\% (13B). This corroborates previous findings that LoRA-style adaptation tends to forget less. However, as we discuss later, it also learn less from target data~\citep{biderman2024lora,hui-etal-2025-hft}.

In contrast, other methods exhibit more substantial performance drops. The state-of-the-art selective parameter update baselines lag considerably behind SSU-Wanda.
For instance, the performance of HFT drops by 18.0\% (7B) and 15.1\% (13B), while the target-data-driven GMT degrades by 27.7\% (7B) and 26.3\% (13B).
Notably, the static HFT method preserves source capabilities more effectively than the dynamic GMT method, supporting our main hypothesis that optimizing on signals from unstructured target data risks corrupting the foundational abilities of an instruct model (\S\ref{sec:introduction}).
The risk of standard adaptation is starkly illustrated by the overall performance of full fine-tuning (FFT). FFT suffers a drastic average performance loss of 34.1\% (7B) and 32.3\% (13B).

Finally, the low performance of baseline SSU variants (SSU-Rand and SSU-Mag) highlights the importance of the source-data-driven scoring. While both freezing random columns (SSU-Rand) and columns selected by magnitude alone (SSU-Mag) outperform FFT, they substantially underperform SSU-Wanda.
SSU-Rand performance is 18.2\% (7B) and 16.0\% (13B) lower than Source, while SSU-Mag causes even greater drops of 23.0\% (7B) and 21.7\% (13B). 
The substantial underperformance of these calibration data-free approaches underscores the critical need for a source-data-informed importance scoring method for preserving the core capabilities of an instruct model in the source language. 
As we demonstrate in \S\ref{sec:analysis}, this principle is not limited to Wanda; other source-data-driven scoring methods are also highly effective, confirming the versatility of the SSU framework.

\paragraph{Safety.}
SSU-Wanda also best preserves the safety alignment of the source, with small performance drops of only 0.1\% (7B) and 2.0\% (13B) compared to Source. In contrast, FFT and the target-data-driven GMT cause large drops, with safety scores dropping by up to 10.2\%. While other selective methods partially preserve source performance, they still lag behind SSU-Wanda.

\paragraph{Source Language.}

SSU-Wanda not only preserves source capabilities but also enhances them in the cross-lingual translation task. 
It ranks top for the 7B model across all benchmarks and leads in MT and MMLU for the 13B model with a close second in SUM and MRC.
Notably, its performance on target-to-English MT improves substantially by up to 52.3\% relative to Source.
Monolingual task performance (SUM, MRC, and MMLU) is almost perfectly maintained, with relative drops never exceeding 2.0\% (7B) and 1.0\% (13B). AdaLoRA is the second-best performer overall, also showing strong preservation across monolingual tasks. However, its gains in the MT task are substantially smaller, the worst among all approaches.
This suggests that while LoRA-based methods effectively prevent forgetting, the structural isolation of their updates may be less adept at integrating new linguistic knowledge for complex cross-lingual tasks.
The remaining adaptation methods generally exhibit greater performance degradation than SSU-Wanda, consistent with instruction-following and safety results.\looseness=-1

\begin{figure*}[t]
\centering
\includegraphics[width=\textwidth]{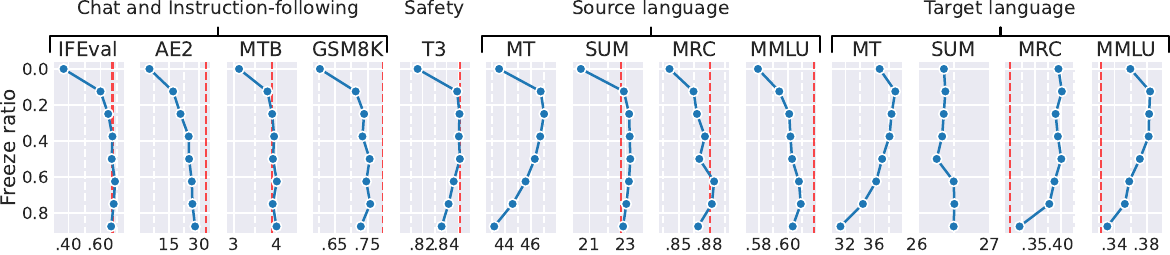}
\caption{
Performance across freezing ratios using SSU on Igbo as target language.
The dashed red line indicates Source performance (omitted for MT and SUM due to very low scores).
See \S\ref{subsec:metrics} for details on evaluation metrics.
}
\label{fig:ratio}
\end{figure*}

\paragraph{Target Language.}
Finally, SSU-Wanda demonstrates exceptional performance on target language tasks, securing the best results across all benchmarks for both model scales in the majority of cases.
Crucially, its performance is highly competitive with FFT, even surpassing it on all benchmarks for 7B models and on half for 13B models.
The performance difference between SSU and FFT is consistently minimal, confirming that SSU-Wanda achieves the target-language gains of a full update with drastically smaller catastrophic forgetting.
This aligns with observations from optimization theory, arguing that freezing parameters acts as a regularization term that stabilizes training and enables a sparse fine-tuned model to match or exceed the performance of its dense counterpart~\citep{Fu_Yang_So_Lam_Bing_Collier_2023,10656381,hui-etal-2025-hft}. All the other selective parameter update methods also yield solid improvements, though typically smaller than those of SSU-Wanda.
In contrast, AdaLoRA shows the smallest improvement and often fails to surpass the source model. This confirms that LoRA-based methods have a smaller inductive bias from the target data~\citep{biderman2024lora,hui-etal-2025-hft}. 
This highlights the unique effectiveness of SSU-Wanda, which successfully masters tasks in the target language while preserving its original knowledge and abilities in the source.

\textit{Overall, SSU-Wanda demonstrates the benefits of full fine-tuning without the associated catastrophic forgetting, consistently outperforming all other evaluated methods.}

\section{Analysis} \label{sec:analysis}

This section evaluates the robustness of the SSU framework by isolating the impact of core design choices and hyperparameters.
Due to resource constraints, we use the 7B model with our primary method, SSU-Wanda.
We select Igbo as the target language, as it is the most underrepresented language among our target languages (Table \ref{tab:language}).

\paragraph{Parameter Freezing Ratio.}
While we use a default 50\% freezing ratio for fair comparison with baselines following \citet{hui-etal-2025-hft}, this hyperparameter impacts performance.
We therefore evaluate freezing ratios from 0\% (defaulting to FFT) to 87.5\% in 12.5\% increments.
Figure \ref{fig:ratio} shows that source language performance, such as chat and safety, generally improves with higher freezing ratios.
In contrast, performance on target language tasks often shows an opposite trend, degrading as more parameters are frozen, with a particularly sharp drop in MMLU after reaching a 37.5\% ratio. Target-to-English MT is a notable exception. Although the models generate English text, performance declines as the freezing ratio increases, particularly after 37.5\%. This trend contradicts other source tasks.
This occurs because MT requires knowledge of both source and target languages.

\begin{table*}[t]
\centering
\small
\renewcommand{\arraystretch}{0.89}
\setlength{\aboverulesep}{2.25pt}
\setlength{\belowrulesep}{2.25pt}
\resizebox{\textwidth}{!}{
\begin{tabular}{@{}l@{\hspace{3pt}}l@{\hspace{4pt}}c@{\hspace{6pt}}c@{\hspace{6pt}}c@{\hspace{6pt}}c@{\hspace{6pt}}c@{\hspace{6pt}}c@{\hspace{6pt}}c@{\hspace{6pt}}c@{\hspace{6pt}}c@{\hspace{6pt}}c@{\hspace{6pt}}c@{\hspace{6pt}}c@{\hspace{6pt}}c}\toprule
& & \multicolumn{4}{c}{\textbf{Chat and Instruction-following}} & \textbf{Safety} & \multicolumn{4}{c}{\textbf{Source language}} & \multicolumn{4}{c}{\textbf{Target language} (Igbo)} \\
\cmidrule(lr){3-6} \cmidrule(lr){8-11} \cmidrule(lr){12-15}
& \textbf{Approach} & \textbf{IFEval} & \textbf{AE2} & \textbf{MTB} & \textbf{GSM8K} & \textbf{T3} & \textbf{MT} & \textbf{SUM} & \textbf{MRC} & \textbf{MMLU} & \textbf{MT} & \textbf{SUM} & \textbf{MRC} & \textbf{MMLU} \\
\midrule

& \cellcolor{gray!20}Source & \cellcolor{gray!20}.675 & \cellcolor{gray!20}32.6 & \cellcolor{gray!20}3.98 & \cellcolor{gray!20}.796 & \cellcolor{gray!20}.851 & \cellcolor{gray!20}28.5 & \cellcolor{gray!20}22.8 & \cellcolor{gray!20}.880 & \cellcolor{gray!20}.618 & \cellcolor{gray!20}23.0 & \cellcolor{gray!20}23.3 & \cellcolor{gray!20}.301 & \cellcolor{gray!20}.323 \\ 
& SSU (Default) & .670 & 25.0 & 3.92 & .756 & .851 & 46.3 & 23.3 & .870 & .603 & 37.1 & 26.3 & .401 & .371 \\
\midrule

\multirow{2}{*}{\large \ding{172}} & Row-wise & .548 & 11.3 & 3.74 & .675 & .846 & 46.0 & 21.8 & .862 & .598 & 36.9 & 26.5 & .407 & .358 \\
& Element-wise & .457 & 7.7 & 3.35 & .657 & .829 & 46.4 & 21.1 & .851 & .587 & 38.3 & 26.5 & .399 & .370 \\
\midrule

\multirow{4}{*}{\large \ding{173}} & SSU-Rand & .564 & 12.5 & 3.75 & .680 & .838 & 45.9 & 22.4 & .856 & .597 & 37.3 & 26.4 & .401 & .355 \\
& SSU-Mag & .497 & 8.9 & 3.59 & .638 & .828 & 45.1 & 21.7 & .852 & .592 & 36.6 & 26.2 & .379 & .348 \\ \noalign{\vskip\aboverulesep}\cdashline{2-15}[2pt/1.2pt]\noalign{\vskip\belowrulesep}
& SSU-SparseGPT & .678 & 24.5 & 3.89 & .751 & .843 & 46.2 & 23.1 & .876 & .604 & 37.2 & 26.5 & .400 & .372 \\
& SSU-FIM & .669 & 26.3 & 3.94 & .747 & .847 & 46.4 & 23.2 & .874 & .609 & 37.1 & 26.5 & .399 & .371 \\

\midrule
{\large \ding{174}} & Alpaca & .673 & 24.0 & 3.97 & .750 & .849 & 46.7 & 23.1 & .874 & .604 & 37.1 & 26.2 & .394 & .379 \\

\bottomrule
\end{tabular}%
}
\caption{Ablation studies on {\large \ding{172}} freezing structures, {\large \ding{173}} importance scoring methods, and {\large \ding{174}} calibration data sources. All models are evaluated on 7B and Igbo as the target language. ``Default'' denotes SSU-Wanda with column-wise freezing using the original tulu-3-sft-olmo-2-mixture data for calibration.
}
\label{tab:ablation_all}
\end{table*}

Our results show a trade-off between source knowledge retention and target language acquisition.
Therefore, we recommend practitioners tailor the freezing ratio to specific goals:
\textbf{General purpose}: A default 50\% ratio offers balanced performance.
\textbf{Source-capability priority}: A higher ratio ($\ge$ 60\%) is optimal, as performance on tasks like IFEval, MRC, and MMLU plateaus around this point.
\textbf{Target-language priority}: A lower ratio ($\le$ 40\%) is preferable, given the performance drops observed in MT and MMLU beyond this threshold.
We extend this analysis to baselines in Appendix \ref{appendix:ratio_baseline}, finding that HFT consistently underperforms SSU despite following a similar performance-scaling pattern, while GMT fails to preserve source capabilities regardless of the ratio.

\paragraph{Alternative Freezing Methods.}

SSU employs column-wise freezing to preserve the entire processing pathway of critical source features (\S\ref{subsec:freezing}).
To validate this design choice, we compare its effectiveness against row-wise and element-wise freezing.
As shown in Table \ref{tab:ablation_all}~\ding{172}, the results demonstrate a clear advantage for our column-wise approach.
Column-wise freezing consistently achieves the best performance on chat, safety, and source language tasks.\footnote{While row-wise freezing preserves all connections from a single input neuron, it fails to protect any single, complete output feature. This explains its weaker performance across chat, safety, and source language tasks.} On target language tasks, it remains highly competitive, with only a 1.2 point drop on MT compared to element-wise freezing.
These results validate the guiding hypothesis for the design of SSU: \textit{preserving entire feature pathways is a critical strategy to safeguard source knowledge while enabling effective target-language adaptation.}
We provide a theoretical grounding for these structural constraints and their relation to the stability-plasticity dilemma in Appendix \ref{appendix:theory}.

\paragraph{Alternative Importance Scoring Methods.}

SSU is compatible with importance scoring methods beyond Wanda.
To demonstrate this, we evaluate two source-data-driven methods: SparseGPT~\citep{pmlr-v202-frantar23a} and the diagonal of the Fisher Information Matrix~\citep[FIM]{doi:10.1073/pnas.1611835114}; see Appendix \ref{appendix:impl_details} for details.
In monolingual source tasks, SSU-SparseGPT and SSU-FIM show comparable average performance drops (4.3\% and 3.5\%, respectively) to SSU-Wanda (4.0\%), as shown in Table \ref{tab:ablation_all}~\ding{173}.
This contrasts sharply with the larger drops of data-free variants like SSU-Rand (13.5\%) and SSU-Mag (17.9\%).
These findings demonstrate the versatility of SSU, offering strong performance across various source-data-driven scoring methods.\looseness=-1

\begin{table}[th]
\centering
\renewcommand{\arraystretch}{0.89}
\small
\resizebox{\linewidth}{!}{
\begin{tabular}{llc@{\hspace{6pt}}c@{\hspace{6pt}}c@{\hspace{6pt}}c@{\hspace{6pt}}c@{\hspace{6pt}}}
\toprule
& & \multicolumn{5}{c}{\textbf{HumanEval} ($\uparrow$)}\\
\multicolumn{2}{l}{\textbf{Approach}} & ne & ky & am & ha & ig\\
\midrule

\multirow{9}{*}{\rotatebox{90}{7B}} & \cellcolor{gray!20}\raisebox{-0.5ex}{Source} & \cellcolor{gray!20}.445\raisebox{-1.25ex}{\hspace{0.1em}{\scriptsize +0.0}} & \cellcolor{gray!20}.445\raisebox{-1.25ex}{\hspace{0.1em}{\scriptsize +0.0}} & \cellcolor{gray!20}.445\raisebox{-1.25ex}{\hspace{0.1em}{\scriptsize +0.0}} & \cellcolor{gray!20}.445\raisebox{-1.25ex}{\hspace{0.1em}{\scriptsize +0.0}} & \cellcolor{gray!20}.445\raisebox{-1.25ex}{\hspace{0.1em}{\scriptsize +0.0}} \\ \cmidrule{2-7}
 & \raisebox{-0.5ex}{FFT} & .287\raisebox{-1.25ex}{\hspace{0.1em}{\scriptsize -35.5}} & .226\raisebox{-1.25ex}{\hspace{0.1em}{\scriptsize -49.2}} & .268\raisebox{-1.25ex}{\hspace{0.1em}{\scriptsize -39.8}} & .128\raisebox{-1.25ex}{\hspace{0.1em}{\scriptsize -71.2}} & .201\raisebox{-1.25ex}{\hspace{0.1em}{\scriptsize -54.8}} \\
 & \raisebox{-0.5ex}{AdaLoRA} & \textbf{.451}\raisebox{-1.25ex}{\hspace{0.1em}{\scriptsize +1.3}} & \textbf{.384}\raisebox{-1.25ex}{\hspace{0.1em}{\scriptsize -13.7}} & .354\raisebox{-1.25ex}{\hspace{0.1em}{\scriptsize -20.5}} & \underline{.323}\raisebox{-1.25ex}{\hspace{0.1em}{\scriptsize -27.4}} & .348\raisebox{-1.25ex}{\hspace{0.1em}{\scriptsize -21.8}} \\
 & \raisebox{-0.5ex}{HFT} & .384\raisebox{-1.25ex}{\hspace{0.1em}{\scriptsize -13.7}} & \underline{.335}\raisebox{-1.25ex}{\hspace{0.1em}{\scriptsize -24.7}} & \underline{.366}\raisebox{-1.25ex}{\hspace{0.1em}{\scriptsize -17.8}} & \underline{.323}\raisebox{-1.25ex}{\hspace{0.1em}{\scriptsize -27.4}} & \underline{.354}\raisebox{-1.25ex}{\hspace{0.1em}{\scriptsize -20.5}} \\
 & \raisebox{-0.5ex}{GMT} & .274\raisebox{-1.25ex}{\hspace{0.1em}{\scriptsize -38.4}} & .287\raisebox{-1.25ex}{\hspace{0.1em}{\scriptsize -35.5}} & .323\raisebox{-1.25ex}{\hspace{0.1em}{\scriptsize -27.4}} & .177\raisebox{-1.25ex}{\hspace{0.1em}{\scriptsize -60.2}} & .256\raisebox{-1.25ex}{\hspace{0.1em}{\scriptsize -42.5}} \\ \cmidrule{2-7}
 & \raisebox{-0.5ex}{SSU-Rand} & .396\raisebox{-1.25ex}{\hspace{0.1em}{\scriptsize -11.0}} & \underline{.335}\raisebox{-1.25ex}{\hspace{0.1em}{\scriptsize -24.7}} & .323\raisebox{-1.25ex}{\hspace{0.1em}{\scriptsize -27.4}} & .262\raisebox{-1.25ex}{\hspace{0.1em}{\scriptsize -41.1}} & .323\raisebox{-1.25ex}{\hspace{0.1em}{\scriptsize -27.4}} \\
 & \raisebox{-0.5ex}{SSU-Mag} & .317\raisebox{-1.25ex}{\hspace{0.1em}{\scriptsize -28.8}} & .293\raisebox{-1.25ex}{\hspace{0.1em}{\scriptsize -34.2}} & .311\raisebox{-1.25ex}{\hspace{0.1em}{\scriptsize -30.1}} & .268\raisebox{-1.25ex}{\hspace{0.1em}{\scriptsize -39.8}} & .293\raisebox{-1.25ex}{\hspace{0.1em}{\scriptsize -34.2}} \\ \noalign{\vskip\aboverulesep}\cdashline{2-7}[2pt/1.2pt]\noalign{\vskip\belowrulesep}
 & \raisebox{-0.5ex}{SSU-Wanda} & \underline{.402}\raisebox{-1.25ex}{\hspace{0.1em}{\scriptsize -9.7}} & \textbf{.384}\raisebox{-1.25ex}{\hspace{0.1em}{\scriptsize -13.7}} & \textbf{.396}\raisebox{-1.25ex}{\hspace{0.1em}{\scriptsize -11.0}} & \textbf{.390}\raisebox{-1.25ex}{\hspace{0.1em}{\scriptsize -12.4}} & \textbf{.421}\raisebox{-1.25ex}{\hspace{0.1em}{\scriptsize -5.4}} \\
\midrule
\multirow{9}{*}{\rotatebox{90}{13B}} & \cellcolor{gray!20}\raisebox{-0.5ex}{Source} & \cellcolor{gray!20}.524\raisebox{-1.25ex}{\hspace{0.1em}{\scriptsize +0.0}} & \cellcolor{gray!20}.524\raisebox{-1.25ex}{\hspace{0.1em}{\scriptsize +0.0}} & \cellcolor{gray!20}.524\raisebox{-1.25ex}{\hspace{0.1em}{\scriptsize +0.0}} & \cellcolor{gray!20}.524\raisebox{-1.25ex}{\hspace{0.1em}{\scriptsize +0.0}} & \cellcolor{gray!20}.524\raisebox{-1.25ex}{\hspace{0.1em}{\scriptsize +0.0}} \\ \cmidrule{2-7}
 & \raisebox{-0.5ex}{FFT} & .445\raisebox{-1.25ex}{\hspace{0.1em}{\scriptsize -15.1}} & .317\raisebox{-1.25ex}{\hspace{0.1em}{\scriptsize -39.5}} & .451\raisebox{-1.25ex}{\hspace{0.1em}{\scriptsize -14.0}} & .152\raisebox{-1.25ex}{\hspace{0.1em}{\scriptsize -71.0}} & .152\raisebox{-1.25ex}{\hspace{0.1em}{\scriptsize -71.0}} \\
 & \raisebox{-0.5ex}{AdaLoRA} & .476\raisebox{-1.25ex}{\hspace{0.1em}{\scriptsize -9.2}} & \underline{.457}\raisebox{-1.25ex}{\hspace{0.1em}{\scriptsize -12.9}} & \underline{.500}\raisebox{-1.25ex}{\hspace{0.1em}{\scriptsize -4.7}} & .433\raisebox{-1.25ex}{\hspace{0.1em}{\scriptsize -17.4}} & \underline{.476}\raisebox{-1.25ex}{\hspace{0.1em}{\scriptsize -9.2}} \\
 & \raisebox{-0.5ex}{HFT} & .451\raisebox{-1.25ex}{\hspace{0.1em}{\scriptsize -14.0}} & .439\raisebox{-1.25ex}{\hspace{0.1em}{\scriptsize -16.3}} & .451\raisebox{-1.25ex}{\hspace{0.1em}{\scriptsize -14.0}} & .378\raisebox{-1.25ex}{\hspace{0.1em}{\scriptsize -27.9}} & .433\raisebox{-1.25ex}{\hspace{0.1em}{\scriptsize -17.4}} \\
 & \raisebox{-0.5ex}{GMT} & .451\raisebox{-1.25ex}{\hspace{0.1em}{\scriptsize -14.0}} & .439\raisebox{-1.25ex}{\hspace{0.1em}{\scriptsize -16.3}} & .463\raisebox{-1.25ex}{\hspace{0.1em}{\scriptsize -11.7}} & .360\raisebox{-1.25ex}{\hspace{0.1em}{\scriptsize -31.3}} & .378\raisebox{-1.25ex}{\hspace{0.1em}{\scriptsize -27.9}} \\ \cmidrule{2-7}
 & \raisebox{-0.5ex}{SSU-Rand} & \textbf{.524}\raisebox{-1.25ex}{\hspace{0.1em}{\scriptsize -0.1}} & .451\raisebox{-1.25ex}{\hspace{0.1em}{\scriptsize -14.0}} & .463\raisebox{-1.25ex}{\hspace{0.1em}{\scriptsize -11.7}} & \underline{.451}\raisebox{-1.25ex}{\hspace{0.1em}{\scriptsize -14.0}} & .415\raisebox{-1.25ex}{\hspace{0.1em}{\scriptsize -20.9}} \\
 & \raisebox{-0.5ex}{SSU-Mag} & .415\raisebox{-1.25ex}{\hspace{0.1em}{\scriptsize -20.9}} & .427\raisebox{-1.25ex}{\hspace{0.1em}{\scriptsize -18.6}} & .427\raisebox{-1.25ex}{\hspace{0.1em}{\scriptsize -18.6}} & .250\raisebox{-1.25ex}{\hspace{0.1em}{\scriptsize -52.3}} & .329\raisebox{-1.25ex}{\hspace{0.1em}{\scriptsize -37.3}} \\ \noalign{\vskip\aboverulesep}\cdashline{2-7}[2pt/1.2pt]\noalign{\vskip\belowrulesep}
 & \raisebox{-0.5ex}{SSU-Wanda} & \underline{.482}\raisebox{-1.25ex}{\hspace{0.1em}{\scriptsize -8.1}} & \textbf{.512}\raisebox{-1.25ex}{\hspace{0.1em}{\scriptsize -2.4}} & \textbf{.537}\raisebox{-1.25ex}{\hspace{0.1em}{\scriptsize +2.4}} & \textbf{.482}\raisebox{-1.25ex}{\hspace{0.1em}{\scriptsize -8.1}} & \textbf{.500}\raisebox{-1.25ex}{\hspace{0.1em}{\scriptsize -4.7}} \\

\bottomrule
\end{tabular}
}
\caption{Coding performance on HumanEval (pass@1).
\textbf{Bold} and \underline{underlined} indicate best and second-best methods for each model scale.}
\label{tab:coding}
\end{table}

\paragraph{Calibration Data for Parameter Importance Scoring.}

SSU-Wanda requires source calibration data to identify critical model weights since it relies on Wanda for parameter importance scoring.
While we use the original instruction-tuning data for OLMo 2 in our main experiments, this is often unavailable for other frontier models.
We therefore investigate the efficacy of using an alternative, publicly available dataset.
Specifically, we use Alpaca~\citep{alpaca} as the calibration dataset and follow the same preprocessing and training procedures as the original data.
Table \ref{tab:ablation_all}~\ding{174} shows that performance with Alpaca is highly comparable to that with the original data, with a maximum difference of 1.0, demonstrating the robustness of SSU-Wanda to the choice of calibration data.
We observe similar robustness regarding calibration data size; reducing samples from 500 to 128 yields negligible performance differences (see Appendix \ref{appendix:calibration_size}).\looseness=-1

\begin{table*}[th]
\centering
\small
\renewcommand{\arraystretch}{0.89}
\setlength{\aboverulesep}{2.25pt}
\setlength{\belowrulesep}{2.25pt}
\resizebox{\textwidth}{!}{
\begin{tabular}{lc@{\hspace{6pt}}c@{\hspace{6pt}}c@{\hspace{6pt}}c@{\hspace{6pt}}c@{\hspace{6pt}}c@{\hspace{6pt}}c@{\hspace{6pt}}c@{\hspace{6pt}}c@{\hspace{6pt}}c@{\hspace{6pt}}c@{\hspace{6pt}}c@{\hspace{6pt}}c}
\toprule
& \multicolumn{4}{c}{\textbf{Chat and Instruction-following}} & \textbf{Safety} & \multicolumn{4}{c}{\textbf{Source language}} & \multicolumn{4}{c}{\textbf{Target language} (Igbo)} \\
\cmidrule(lr){2-5} \cmidrule(lr){7-10} \cmidrule(lr){11-14}
\textbf{Approach} & \textbf{IFEval} & \textbf{AE2} & \textbf{MTB} & \textbf{GSM8K} & \textbf{T3} & \textbf{MT} & \textbf{SUM} & \textbf{MRC} & \textbf{MMLU} & \textbf{MT} & \textbf{SUM} & \textbf{MRC} & \textbf{MMLU} \\
\midrule

\cellcolor{gray!20}Source & \cellcolor{gray!20}.675 & \cellcolor{gray!20}32.6 & \cellcolor{gray!20}3.98 & \cellcolor{gray!20}.796 & \cellcolor{gray!20}.851 & \cellcolor{gray!20}28.5 & \cellcolor{gray!20}22.8 & \cellcolor{gray!20}.880 & \cellcolor{gray!20}.618 & \cellcolor{gray!20}23.0 & \cellcolor{gray!20}23.3 & \cellcolor{gray!20}.301 & \cellcolor{gray!20}.323 \\ \midrule
FFT & .645 & 17.1 & 3.95 & .685 & .835 & \underline{42.6} & 21.9 & .857 & .604 & \textbf{30.9} & 26.2 & .341 & \underline{.349} \\
AdaLoRA & .678 & \textbf{30.6} & \textbf{4.05} & \underline{.750} & .837 & 28.4 & \underline{22.5} & .874 & \textbf{.614} & 16.7 & 24.8 & .270 & .318 \\
HFT & \textbf{.693} & 25.1 & 3.89 & .732 & \underline{.841} & 42.3 & 22.4 & .870 & .607 & 29.3 & \textbf{26.6} & .328 & .346 \\
GMT & .665 & 23.2 & 3.93 & .726 & .838 & \textbf{43.0} & \underline{22.5} & \textbf{.879} & .611 & \underline{30.7} & \underline{26.3} & \underline{.349} & .347 \\
\midrule
SSU-Rand & \underline{.682} & 24.4 & 3.95 & .729 & .831 & 42.0 & \underline{22.5} & .871 & .610 & 28.9 & \underline{26.3} & .337 & .343 \\
SSU-Mag & .664 & 21.3 & 3.97 & .704 & .831 & \underline{42.6} & 22.2 & .874 & .607 & 28.9 & 26.2 & .340 & .336\\\noalign{\vskip\aboverulesep}\cdashline{1-14}[2pt/1.2pt]\noalign{\vskip\belowrulesep}
SSU-Wanda & .671 & \underline{27.9} & \underline{3.98} & \textbf{.783} & \textbf{.848} & 42.3 & \textbf{22.6} & \underline{.878} & \underline{.613} & 28.9 & \textbf{26.6} & \textbf{.357} & \textbf{.352} \\

\bottomrule
\end{tabular}
}
\caption{Performance of models adapted with 20M tokens of target language data.
All models are evaluated on 7B and Igbo as the target language.
\textbf{Bold} and \underline{underlined} indicate best and second-best methods.}
\label{tab:20m_tokens_performance}
\end{table*}

\paragraph{Universality of Shielded Parameters.}
We investigate whether shielded parameters are specific to the English language.
We hypothesize that SSU preserves universal functional units, such as logic and reasoning, rather than surface-level linguistic features.
To evaluate this, we measure performance on HumanEval~\citep{chen2021evaluatinglargelanguagemodels}, where logic transcends natural language.
Table \ref{tab:coding} demonstrates that SSU-Wanda maintains coding proficiency near the levels of Source.
In contrast, FFT and GMT suffer substantial degradation. For the 7B models, SSU-Wanda shows a 10.4\% average relative performance drop, whereas FFT suffers a severe loss of 49.7\%.
The 13B models exhibit a comparable trend, with SSU-Wanda declining by only 4.2\%.
These results confirm that SSU safeguards fundamental capabilities, such as reasoning and logic, which are shared across languages.
A proxy analysis regarding target-language instruction-following in Appendix \ref{appendix:proxy} further supports these findings.

\paragraph{Ultra-low-resource Settings.}
To evaluate the efficacy of SSU under extreme data constraints, we adapt models using only 20M tokens, representing 10\% of our default adaptation set.
As shown in Table \ref{tab:20m_tokens_performance}, SSU-Wanda achieves the best or second best performances in 10 out of 13 tasks in this ultra-low-resource regime.
While the reduced training data naturally limits overall weight drift, SSU-Wanda exhibits substantially better retention of core capabilities (AE2, GSM8K, Safety) than baselines, which show immediate degradation even with minimal updates.
AdaLoRA remains a notable exception, as it ``learns less and forgets less''~\citep{biderman2024lora,hui-etal-2025-hft}, resulting in strong source retention but substantially weaker target-language acquisition.
Furthermore, SSU-Wanda achieves target-language improvements in SUM (26.6), MRC (.357), and MMLU (.352) that exceed those of FFT.
This confirms that shielding critical source parameters acts as a beneficial regularizer for acquiring target linguistic features even when training data is scarce.

\paragraph{Qualitative Analysis.}
SSU-Wanda surpasses other state-of-the-art selective parameter update baselines across all chat and instruction-following benchmarks (Table \ref{tab:aggregated_with_deltas}).
This performance gap stems partly from the susceptibility of baseline methods to code-mixing (i.e., the unintentional blending of multiple languages in responses) or generating responses entirely in the target language, despite English instructions.
Specifically, analyzing the language ratio in generated responses on AE2 shows that SSU restricts code-mixing to merely 1.0\% of its responses on average for the 7B models.
In contrast, HFT and GMT generate code-mixed text in 6.4\% and 16.9\%, respectively.\footnote{We use GlotLID~\citep[Commit 28d4264]{kargaran-etal-2023-glotlid} to identify code-mixed responses where normalized English confidence falls below 0.9.}
This substantial reduction in the occurrence of code-mixing reflects the more robust retention of source language abilities and superior chat performance.
A typical example of this behavior for models trained on Igbo is provided below.
\begin{tcolorbox}[
    colback=lightgray!10, 
    colframe=black, 
    boxrule=0.2mm, 
    width=\linewidth,
    arc=2mm, 
    auto outer arc,
    top=1mm,
    bottom=1mm,
]
\scriptsize
\setlength{\leftskip}{-1.25em}
\setlength{\rightskip}{-1.25em}
\texttt{\bf Instruction in EN:} How do I take care of a wooden table?\vspace{0.5em}\\
\texttt{\bf HFT Response:} \textit{To take care nke a wood table, clean ya na a soft duster ma \d{o} b\d{u} microfiber towel iji wep\d{u} dust na grime. N'ihe banyere stains, mix otu teaspoon nke mild detergent na mmiri na-a teaspoon... }
(\textit{continued.}) \vspace{0.5em}\\
\texttt{\bf SSU Response:} \textit{To take care, clean your wooden table regularly with mild soap and water. Use a soft cloth for polishing, applying wood polish or beeswax to maintain its natural finish. Avoid placing hot items directly on the surface to prevent scratches. Keep it away from direct sunlight and excessive moisture.}
\end{tcolorbox}

\section{Conclusion}
We introduced \textbf{S}ource-\textbf{S}hielded \textbf{U}pdates (\textbf{SSU}) for language adaptation of instruct models using only unlabeled target language data.
Our SSU framework proactively identifies critical source knowledge using an importance scoring method and a small set of source calibration data.
It then shields this knowledge via a column-wise freezing strategy before adaptation, effectively preventing catastrophic forgetting in the source language.
Extensive experiments across five languages and two model scales show that SSU best preserves crucial source capabilities, such as instruction-following and safety, over strong baselines while achieving target language proficiency matching or surpassing full fine-tuning.
This work provides an effective and scalable pathway to expand the linguistic reach of instruct models without costly, specialized data, opening avenues for robust model adaptation.

\section*{Limitations}

\paragraph{Baselines Scope.}
This paper primarily compares SSU against state-of-the-art selective parameter update methods for LLM adaptation, specifically HFT and GMT. Additional evaluations against LoTA and S2FT are provided in Appendix \ref{appendix:additional_baselines} to ensure an extensive evaluation.
Strategies such as source data mixing~\citep{zheng-etal-2024-breaking,sainz-etal-2025-instructing} and model merging and post-hoc pruning~\citep{blevins-etal-2024-breaking,huang-etal-2025-mitigating} are orthogonal to this work (as discussed in \S\ref{sec:related_work}).
Furthermore, foundational continual learning methods for task-incremental learning, such as HAT~\citep{pmlr-v80-serra18a}, remain computationally prohibitive for billion-parameter models (see Appendix \ref{appendix:related_work} for discussion).
Consequently, this work prioritizes scalable, LLM-specific methods for comparison to maintain practical relevance. Exploring the synergy between SSU and orthogonal strategies such as model merging or replay remains a promising direction for future research.

\paragraph{Hyperparameter Selection.}
Due to the substantial computational cost of fine-tuning and evaluating 100+ adapted models (e.g., Table \ref{tab:aggregated_with_deltas} encompasses 70 adapted models), this study does not perform exhaustive hyperparameter searches for all approaches including both baselines and the proposed method.
Instead, the experimental protocol follows established language adaptation literature for instruct models~\citep{yamaguchi2025elchatadaptingchatlanguage}.
For freezing ratios, this work adopts the 50\% sparsity level used in HFT~\citep{hui-etal-2025-hft} to facilitate fair comparison, with sensitivity analysis provided in \S\ref{sec:analysis} and Appendix \ref{appendix:ratio_baseline}.
While reported performance might not represent the global optimum for each method across languages, avoiding exhaustive tuning prevents introducing bias toward methods with larger search spaces. Utilizing a standard configuration ensures a rigorous and equitable evaluation of the underlying methods.

\section*{Ethical Considerations}
While the current study on SSU should not present immediate ethical conflicts given its scope on catastrophic forgetting mitigation, the deployment of adapted instruct models in underrepresented languages (e.g., Nepali, Kyrgyz, or Amharic) requires further scrutiny.
These adapted models may unintentionally reinforce harmful biases or introduce safety vulnerabilities that standard benchmarks fail to detect.
Consequently, responsible deployment and continued research into cross-lingual safety alignment remain essential.

\section*{Acknowledgements}
We would like to thank Mingzi Cao, Xingwei Tan, and Huiyin Xue for their valuable feedback.
We acknowledge (1) IT Services at the University of Sheffield for the provision of services for high-performance computing; (2) the use of the University of Oxford Advanced Research Computing (ARC) facility; and (3) EuroHPC Joint Undertaking for awarding us access to MeluXina at LuxProvide, Luxembourg.
AY is supported by the Engineering and Physical Sciences Research Council (EPSRC)  [grant number EP/W524360/1] and the Japan Student Services Organization (JASSO) Student Exchange Support Program (Graduate Scholarship for Degree Seeking Students). AV research is partly supported by UKRI (grants MR/U506734/1 and EP/T02450X/1), CNPq (406926/2025-5) and EQUATE. NA is partly supported by AstraZeneca and the EPSRC (grant EP/Y009800/1).

\bibliography{custom,anthology1,anthology2}

\clearpage
\begin{tcolorbox}[
    title=Appendix Directory,
    colback=lightgray!10,
    colframe=black,
    fonttitle=\bfseries, 
    rounded corners
]
\begin{itemize}
    \item \textbf{Appendix A:} \hyperref[appendix:evaluation_setup]{Evaluation Details}
    
    \item \textbf{Appendix B:} \hyperref[appendix:impl_details]{Implementation Details}
        \begin{itemize}
            \item \hyperref[appendix:general_setup]{General Setup}
            \item \hyperref[appendix:scoring_methods]{Alternative Scoring Method Implementations}
        \end{itemize}
    
    \item \textbf{Appendix C:} \hyperref[appendix:results]{Supplementary Results}
    
    \item \textbf{Appendix D:} \hyperref[appendix:analysis]{Supplementary Analysis}
        \begin{itemize}
            \item \hyperref[appendix:ratio_baseline]{Impact of Freezing Ratio on Baselines}

            \item \hyperref[appendix:calibration_size]{Calibration Data Size for Parameter Importance Scoring}

            \item \hyperref[appendix:additional_baselines]{Comparison to Additional Baselines}

            \item \hyperref[appendix:olmo3]{Generalization to OLMo 3 Architecture}
            
            \item \hyperref[appendix:theory]{Theoretical Analysis}

            \item \hyperref[appendix:proxy]{Proxy Evaluation on Target-Language Instruction-following}
        \end{itemize}
    
    \item \textbf{Appendix E:} \hyperref[appendix:related_work]{Extended Related Work}

    \item \textbf{Appendix F:} \hyperref[appendix:license]{License}

    \item \textbf{Appendix G:} \hyperref[appendix:genai]{Use of Generative AI Tools}
\end{itemize}
\end{tcolorbox}

\appendix
\section{Evaluation Details} \label{appendix:evaluation_setup}

\paragraph{LLM-as-a-Judge.}
Following \citet{yamaguchi2025elchatadaptingchatlanguage}, we use judgments from \href{https://openai.com/index/gpt-4-1/}{GPT-4.1 nano (2025-04-14)} for AE2 and \href{https://huggingface.co/flowaicom/Flow-Judge-v0.1}{\texttt{Flow-Judge-v0.1}} for MTB.

\paragraph{Prompt Templates.}
Table \ref{tab:prompt} shows language-specific prompt templates for each task.

\section{Implementation Details} \label{appendix:impl_details}

\subsection{General Setup} \label{appendix:general_setup}
\paragraph{Hyperparameters.} Tables \ref{tab:params_cpt} and \ref{tab:params_generation} list the hyperparameters in CPT and evaluation, respectively.

\paragraph{Software.} We use HF datasets~\cite[v3.6.0]{lhoest-etal-2021-datasets} for preprocessing, HF transformers~\cite[v4.52.4]{wolf-etal-2020-transformers}, HF peft~\cite[v0.15.2]{peft}, FlashAttention-2~\cite[v2.7.4]{dao2024flashattention} and PyTorch~\cite[v2.6.0]{10.1145/3620665.3640366} for training.
We use lm-evaluation-harness~\cite[v0.4.8]{eval-harness} for IFEval and GSM8K evaluation, alpaca-eval~\cite[v0.6.6]{alpaca_eval} for AE2 evaluation, Ai2 Safety Tool for T3 evaluation,\footnote{Following \citet{lambert2025tulu3pushingfrontiers}, we use their forked version: \url{https://github.com/nouhadziri/safety-eval-fork} (Commit 2920bb8).} and HF LightEval~\cite[Commit 327071f]{lighteval} for the rest.

\paragraph{Hardware.} We mainly use a single AMD MI300X GPU with ROCm 6.4.1 for experiments. Additionally, we use either a single NVIDIA H100 80GB, A100 80GB, or A100 40GB GPU with CUDA 12.9 for evaluation.

\paragraph{Training Cost and Computational Efficiency.}
A primary advantage of the SSU framework is its efficiency, owing to the one-shot nature of the static importance scoring. We break down the computational overhead into two components:

\begin{itemize}
    \item \textbf{Scoring (Stage 1):} Generating the importance mask is highly efficient. For a 7B model with 500 calibration samples (sequence length 2,048), the scoring process takes approximately 95 seconds on a single AMD MI300X GPU. As this stage primarily involves forward passes to collect activations, it is less compute-intensive than training and can even be offloaded to a CPU if GPU memory is limited.
    
    \item \textbf{Adaptation (Stage 3):} Unlike dynamic gradient-masking methods (e.g., GMT), SSU utilizes a static mask. This introduces zero additional overhead during the backward pass. In our OLMo 2 7B experiments, the total training time for SSU (34,156s) was essentially equivalent to full fine-tuning (34,979s), with the minor difference attributable to standard hardware variance.
\end{itemize}

Overall, the pre-computation overhead for SSU represents less than 0.3\% of the total training time, making it a nearly ``cost-free'' intervention relative to standard adaptation.

\subsection{Alternative Scoring Method Implementations} \label{appendix:scoring_methods}

\paragraph{SSU-SparseGPT.} 
This method employs a metric from \citet{pmlr-v202-frantar23a} that approximates second-order information.
The score for any weight $\theta_{ij}$ in an input column $j$ is the average squared activation of the corresponding input neuron: $s_{ij} = \mathbb{E}_{x \in \mathcal{D}_{\text{calib}}} x_j^2$.

\paragraph{SSU-FIM.}
This method uses the diagonal of the Fisher Information Matrix, which measures output sensitivity to parameter changes~\citep{doi:10.1073/pnas.1611835114}. 
We approximate the Fisher score for a parameter $\theta_{ij}$ as the average squared gradient of the negative log-likelihood loss $L$ over $\mathcal{D}_\text{calib}$: $s_{ij} = \mathbb{E}_{(x, y) \in \mathcal{D}_{\text{calib}}}(\frac{\partial L}{\partial \theta_{ij}})^2$.

\begin{table}[!htbp]
\centering
\small
\resizebox{\linewidth}{!}{%
\begin{tabular}{lc}
\toprule
\textbf{Hyperparameters} & \textbf{Values} \\
\midrule
Batch size & 32\\
Number of training steps & 12,208\\
Optimizer & adamw\_apex\_fused \\
Adam $\epsilon$ & 1e-8\\
Adam $\beta_1$ & 0.9\\
Adam $\beta_2$ & 0.999\\
Sequence length & 512\\
Learning rate & 5e-5\\
Learning rate scheduler & cosine\\
Warmup steps & First 5\% of steps \\
Weight decay & 0.01\\
Attention dropout & 0.0\\
Training precision & BF16\\
\midrule
\multicolumn{2}{c}{\textbf{HFT}, \textbf{GMT}, \textbf{SSU}}\\
Target freezing ratio & 0.5\\
\midrule
\multicolumn{2}{c}{\textbf{GMT}}\\
Accumulation interval & 4\\
\midrule
\multicolumn{2}{c}{\textbf{AdaLoRA}}\\
Target $r$ & 8\\
LoRA $\alpha$ & 32\\
LoRA dropout & 0.05\\
$T_\text{init}$ & 1,000\\
$T_\text{final}$ & 8,546\\
$\delta_t$ & 20\\
LoRA $\beta_1$ & 0.85\\
LoRA $\beta_2$ & 0.85\\
Coefficient of orthogonal regularization & 0.5\\
\midrule
\multicolumn{2}{c}{\textbf{LoTA}}\\
Mask calibration steps & 100\\
\midrule
\multicolumn{2}{c}{\textbf{S2FT}}\\
$d_\text{ratio}$ (Down) & 0.015 (equivalent to LoRA $r=8$)\\
$o_\text{ratio}$ (Output) & 0.015 (equivalent to LoRA $r=8$)\\
\bottomrule
\end{tabular}%
}
\caption{Hyperparameters for continual pre-training. Values for GMT and AdaLoRA were selected based on our setup, as they were not provided in their respective original papers \citep{Li_Zhang_Liu_Gong_Wang_Chen_Cheng_2025, hui-etal-2025-hft}.}
\label{tab:params_cpt}
\end{table}

\begin{table}[!htbp]
\centering
\small
\begin{tabular}{lc}
\toprule
\textbf{Parameters} & \textbf{Values} \\
\midrule
Temperature & 0.8\\
Repetition penalty & 1.1\\
Top $k$ & 40\\
Top $p$ & 0.9 (MT, SUM, MTBench)\\
& 0.8 (AE2, IFEval, GSM8K)\\
Sampling & True\\
Max. generated tokens & 128 (MT, SUM)\\
& 512 (AE2)\\
& 1,024 (MTBench)\\
& 1,280 (IFEval)\\
& N/A (GSM8K)\\
\bottomrule
\end{tabular}
\caption{Parameters for generation tasks. N/A for GSM8K indicates that a model generates text until it detects default stop symbols or reaches its maximum sequence length.}
\label{tab:params_generation}
\end{table}

\begin{table*}[!htbp]
    \centering
    \includegraphics[width=\textwidth]{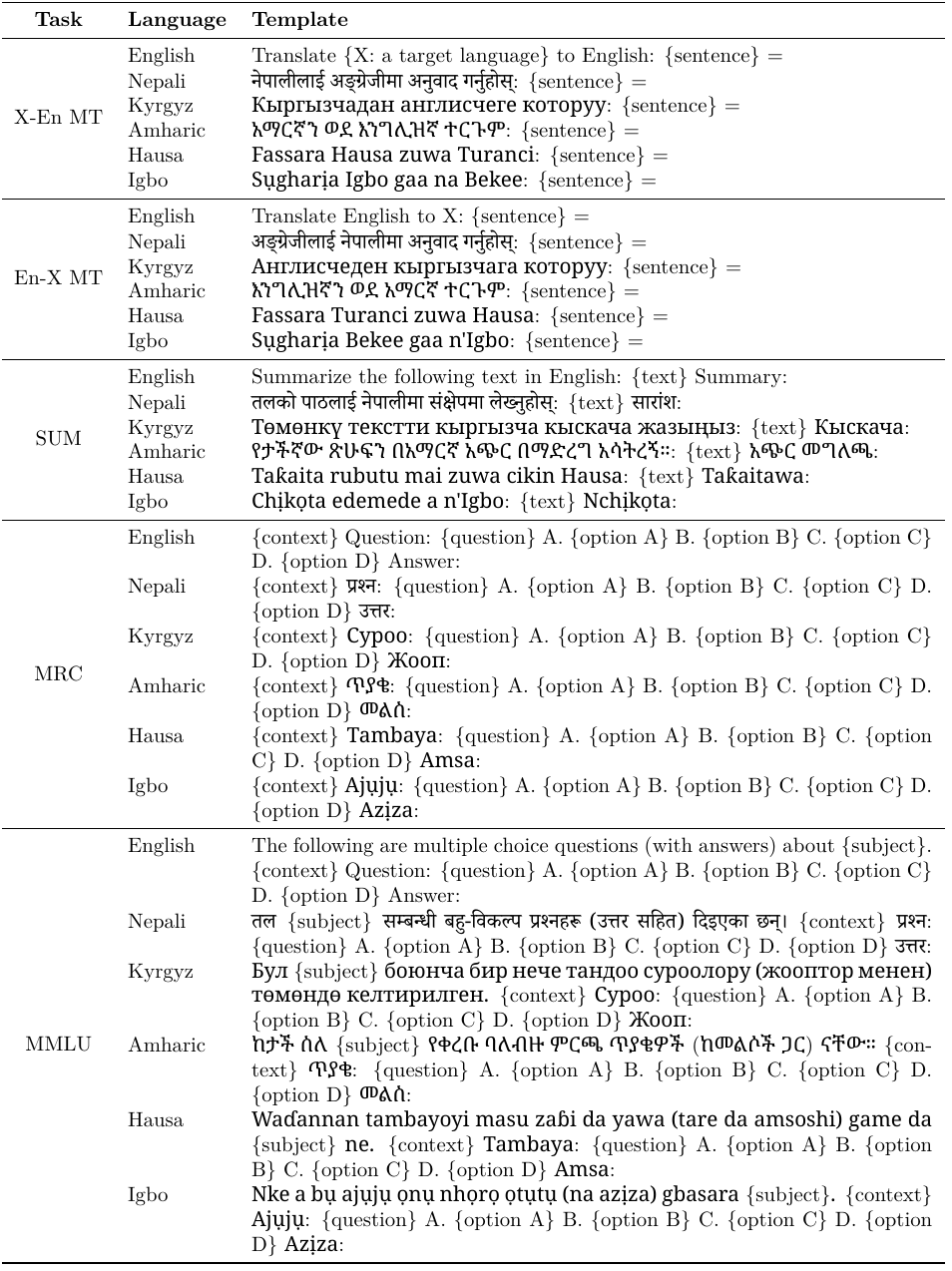}
    \caption{Language-specific prompt templates. We generate the templates for each target language using a machine translation API, following \citet{yong-etal-2023-bloom}.}
    \label{tab:prompt}
\end{table*}

\section{Supplementary Results} \label{appendix:results}
Tables \ref{tab:source_instruct}, \ref{tab:safety}, \ref{tab:source}, and \ref{tab:target} show performances on English chat and instruction-following benchmarks, English safety alignment benchmark, general English benchmarks, and general target language benchmarks, respectively.
Results for IFEval, AE2, MTB, GSM8K, MT, and SUM are averaged across three different runs. The rest are single-run results as they are evaluated in a deterministic-manner.

\begin{table*}[t]
\centering
\small
\resizebox{\textwidth}{!}{
\begin{tabular}{llc@{\hspace{6pt}}c@{\hspace{6pt}}c@{\hspace{6pt}}c@{\hspace{6pt}}c@{\hspace{6pt}}c@{\hspace{6pt}}c@{\hspace{6pt}}c@{\hspace{6pt}}c@{\hspace{6pt}}c@{\hspace{6pt}}c@{\hspace{6pt}}c@{\hspace{6pt}}c@{\hspace{6pt}}c@{\hspace{6pt}}c@{\hspace{6pt}}c@{\hspace{6pt}}c@{\hspace{6pt}}c@{\hspace{6pt}}c@{\hspace{6pt}}c}
\toprule
& & \multicolumn{5}{c}{\textbf{IFEval}} & \multicolumn{5}{c}{\textbf{AE2}} & \multicolumn{5}{c}{\textbf{MTB}} & \multicolumn{5}{c}{\textbf{GSM8K}} \\
\cmidrule(lr){3-7} \cmidrule(lr){8-12} \cmidrule(lr){13-17} \cmidrule(lr){18-22}
\multicolumn{2}{l}{\textbf{Approach}} & ne & ky & am & ha & ig & ne & ky & am & ha & ig & ne & ky & am & ha & ig & ne & ky & am & ha & ig \\
\midrule

\multirow{9}{*}{\rotatebox{90}{7B}} & \cellcolor{gray!20}Source & \cellcolor{gray!20}.675 & \cellcolor{gray!20}.675 & \cellcolor{gray!20}.675 & \cellcolor{gray!20}.675 & \cellcolor{gray!20}.675 & \cellcolor{gray!20}32.6 & \cellcolor{gray!20}32.6 & \cellcolor{gray!20}32.6 & \cellcolor{gray!20}32.6 & \cellcolor{gray!20}32.6 & \cellcolor{gray!20}3.98 & \cellcolor{gray!20}3.98 & \cellcolor{gray!20}3.98 & \cellcolor{gray!20}3.98 & \cellcolor{gray!20}3.98 & \cellcolor{gray!20}.796 & \cellcolor{gray!20}.796 & \cellcolor{gray!20}.796 & \cellcolor{gray!20}.796 & \cellcolor{gray!20}.796 \\ \cmidrule{2-22}
 & FFT & .520 & .480 & .495 & .417 & .369 & 14.3 & 12.6 & 12.1 & 7.8 & 5.2 & 3.80 & 3.50 & 3.60 & 3.40 & 3.12 & .623 & .619 & .593 & .602 & .604 \\
 & AdaLoRA & \textbf{.668} & \textbf{.679} & \textbf{.681} & \underline{.646} & \underline{.669} & \underline{27.2} & \underline{25.7} & \underline{25.7} & \textbf{24.6} & \underline{20.0} & \underline{3.98} & \underline{3.96} & \underline{3.89} & \textbf{3.92} & \underline{3.87} & \underline{.736} & \underline{.742} & \underline{.737} & \underline{.704} & \underline{.685} \\
 & HFT & .636 & .652 & .636 & .604 & .578 & 22.6 & 18.3 & 21.0 & \underline{15.1} & 11.1 & 3.95 & 3.82 & 3.85 & 3.77 & 3.73 & .699 & .689 & .692 & .646 & .659 \\
 & GMT & .596 & .571 & .577 & .405 & .492 & 17.7 & 14.2 & 16.1 & 7.3 & 7.3 & 3.92 & 3.74 & 3.79 & 3.44 & 3.49 & .671 & .607 & .645 & .606 & .648 \\ \cmidrule{2-22}
 & SSU-Rand & .619 & .624 & .634 & .599 & .564 & 24.0 & 19.1 & 19.8 & 14.8 & 12.5 & 3.86 & 3.81 & 3.87 & 3.79 & 3.75 & .701 & .678 & .693 & .660 & .680 \\
 & SSU-Mag & .595 & .617 & .591 & .548 & .497 & 19.2 & 16.8 & 18.3 & 11.5 & 8.9 & 3.87 & 3.86 & 3.81 & 3.79 & 3.59 & .682 & .665 & .660 & .629 & .638 \\ \noalign{\vskip\aboverulesep}\cdashline{2-22}[2pt/1.2pt]\noalign{\vskip\belowrulesep}
 & SSU-Wanda & \underline{.655} & \underline{.664} & \underline{.661} & \textbf{.688} & \textbf{.670} & \textbf{28.1} & \textbf{28.7} & \textbf{28.5} & \textbf{24.6} & \textbf{25.0} & \textbf{4.02} & \textbf{4.02} & \textbf{3.96} & \underline{3.91} & \textbf{3.92} & \textbf{.746} & \textbf{.759} & \textbf{.749} & \textbf{.741} & \textbf{.756} \\
\midrule
\multirow{9}{*}{\rotatebox{90}{13B}} & \cellcolor{gray!20}Source & \cellcolor{gray!20}.763 & \cellcolor{gray!20}.763 & \cellcolor{gray!20}.763 & \cellcolor{gray!20}.763 & \cellcolor{gray!20}.763 & \cellcolor{gray!20}37.2 & \cellcolor{gray!20}37.2 & \cellcolor{gray!20}37.2 & \cellcolor{gray!20}37.2 & \cellcolor{gray!20}37.2 & \cellcolor{gray!20}4.06 & \cellcolor{gray!20}4.06 & \cellcolor{gray!20}4.06 & \cellcolor{gray!20}4.06 & \cellcolor{gray!20}4.06 & \cellcolor{gray!20}.853 & \cellcolor{gray!20}.853 & \cellcolor{gray!20}.853 & \cellcolor{gray!20}.853 & \cellcolor{gray!20}.853 \\ \cmidrule{2-22}
 & FFT & .549 & .468 & .506 & .405 & .314 & 23.6 & 14.7 & 18.6 & 11.9 & 3.7 & 3.91 & 3.66 & 3.69 & 3.43 & 2.93 & .768 & .730 & .732 & .733 & .737 \\
 & AdaLoRA & \textbf{.720} & \textbf{.733} & \textbf{.737} & \underline{.728} & \underline{.675} & \underline{34.6} & \textbf{34.1} & \textbf{33.2} & \underline{30.0} & \underline{28.7} & \textbf{4.10} & \underline{4.08} & \textbf{4.09} & \underline{4.03} & \underline{3.94} & \underline{.812} & \underline{.814} & \underline{.812} & \textbf{.821} & \underline{.815} \\
 & HFT & .693 & .680 & .676 & .578 & .528 & 31.2 & 29.1 & 27.4 & 23.4 & 17.9 & \underline{4.08} & 4.04 & 3.99 & 3.84 & 3.69 & .802 & .793 & .762 & .760 & .765 \\
 & GMT & .628 & .527 & .543 & .404 & .381 & 28.1 & 20.1 & 19.8 & 16.2 & 12.3 & 3.91 & 3.89 & 3.54 & 3.55 & 3.34 & .787 & .759 & .688 & .763 & .771 \\ \cmidrule{2-22}
 & SSU-Rand & .672 & .703 & .677 & .558 & .539 & 30.2 & 28.2 & 26.8 & 21.9 & 16.2 & 3.97 & 3.97 & 3.98 & 3.85 & 3.66 & .787 & .795 & .777 & .766 & .780 \\
 & SSU-Mag & .651 & .648 & .636 & .489 & .434 & 28.3 & 24.8 & 23.5 & 16.8 & 9.7 & 4.00 & 3.93 & 3.98 & 3.76 & 3.35 & .782 & .768 & .755 & .756 & .751 \\ \noalign{\vskip\aboverulesep}\cdashline{2-22}[2pt/1.2pt]\noalign{\vskip\belowrulesep}
 & SSU-Wanda & \underline{.718} & \underline{.723} & \underline{.733} & \textbf{.739} & \textbf{.739} & \textbf{34.7} & \underline{33.7} & \underline{32.2} & \textbf{33.8} & \textbf{32.8} & 4.04 & \textbf{4.11} & \underline{4.01} & \textbf{4.10} & \textbf{4.01} & \textbf{.831} & \textbf{.827} & \textbf{.814} & \underline{.808} & \textbf{.830} \\

\bottomrule
\end{tabular}
}
\caption{Performance on chat and instruction-following tasks in English.
The best and second-best adaptation approaches for each model scale are indicated in \textbf{bold} and \underline{underlined}, respectively.}
\label{tab:source_instruct}
\end{table*}

\begin{table}[!htbp]
\centering
\small
\begin{tabular}{llc@{\hspace{6pt}}c@{\hspace{6pt}}c@{\hspace{6pt}}c@{\hspace{6pt}}c@{\hspace{6pt}}}
\toprule
& & \multicolumn{5}{c}{\textbf{T3} ($\uparrow$)}\\
\multicolumn{2}{l}{\textbf{Approach}} & ne & ky & am & ha & ig\\
\midrule

\multirow{9}{*}{\rotatebox{90}{7B}} & \cellcolor{gray!20}Source & \cellcolor{gray!20}.851 & \cellcolor{gray!20}.851 & \cellcolor{gray!20}.851 & \cellcolor{gray!20}.851 & \cellcolor{gray!20}.851 \\ \cmidrule{2-7}
 & FFT & .770 & .791 & .800 & .807 & .816 \\
 & AdaLoRA & \textbf{.842} & .829 & .836 & .806 & .805 \\
 & HFT & .812 & .816 & .839 & \underline{.833} & .828 \\
 & GMT & .777 & .791 & .811 & .782 & .812 \\ \cmidrule{2-7}
 & SSU-Rand & \underline{.824} & \underline{.838} & \underline{.841} & .832 & \underline{.838} \\
 & SSU-Mag & .811 & .813 & .831 & .829 & .828 \\ \noalign{\vskip\aboverulesep}\cdashline{2-7}[2pt/1.2pt]\noalign{\vskip\belowrulesep}
 & SSU-Wanda & \textbf{.842} & \textbf{.846} & \textbf{.855} & \textbf{.856} & \textbf{.851} \\
\midrule
\multirow{9}{*}{\rotatebox{90}{13B}} & \cellcolor{gray!20}Source & \cellcolor{gray!20}.821 & \cellcolor{gray!20}.821 & \cellcolor{gray!20}.821 & \cellcolor{gray!20}.821 & \cellcolor{gray!20}.821 \\ \cmidrule{2-7}
 & FFT & .745 & .710 & .792 & .657 & .782 \\
 & AdaLoRA & \textbf{.816} & \textbf{.805} & .815 & .759 & .799 \\
 & HFT & .790 & .743 & \underline{.817} & .764 & \underline{.812} \\
 & GMT & .756 & .735 & .751 & .736 & .798 \\ \cmidrule{2-7}
 & SSU-Rand & .798 & .756 & .792 & \underline{.768} & .799 \\
 & SSU-Mag & .774 & .742 & .804 & .747 & .811 \\ \noalign{\vskip\aboverulesep}\cdashline{2-7}[2pt/1.2pt]\noalign{\vskip\belowrulesep}
 & SSU-Wanda & \underline{.809} & \underline{.789} & \textbf{.819} & \textbf{.797} & \textbf{.813} \\

\bottomrule
\end{tabular}
\caption{Performance on Tülu 3 safety evaluation suite (T3).
The best and second-best adaptation approaches for each model scale are indicated in \textbf{bold} and \underline{underlined}, respectively.}
\label{tab:safety}
\end{table}

\begin{table*}[t]
\centering
\small
\resizebox{\textwidth}{!}{
\begin{tabular}{llc@{\hspace{6pt}}c@{\hspace{6pt}}c@{\hspace{6pt}}c@{\hspace{6pt}}c@{\hspace{6pt}}c@{\hspace{6pt}}c@{\hspace{6pt}}c@{\hspace{6pt}}c@{\hspace{6pt}}c@{\hspace{6pt}}c@{\hspace{6pt}}c@{\hspace{6pt}}c@{\hspace{6pt}}c@{\hspace{6pt}}c@{\hspace{6pt}}c@{\hspace{6pt}}c@{\hspace{6pt}}c@{\hspace{6pt}}c@{\hspace{6pt}}c@{\hspace{6pt}}c}
\toprule
& & \multicolumn{5}{c}{\textbf{MT}} & \multicolumn{5}{c}{\textbf{SUM}} & \multicolumn{5}{c}{\textbf{MRC}} & \multicolumn{5}{c}{\textbf{MMLU}} \\
\cmidrule(lr){3-7} \cmidrule(lr){8-12} \cmidrule(lr){13-17} \cmidrule(lr){18-22}
\multicolumn{2}{l}{\textbf{Approach}} & ne & ky & am & ha & ig & ne & ky & am & ha & ig & ne & ky & am & ha & ig & ne & ky & am & ha & ig \\
\midrule

\multirow{9}{*}{\rotatebox{90}{7B}} & \cellcolor{gray!20}Source & \cellcolor{gray!20}45.4 & \cellcolor{gray!20}28.8 & \cellcolor{gray!20}19.5 & \cellcolor{gray!20}27.9 & \cellcolor{gray!20}28.5 & \cellcolor{gray!20}22.8 & \cellcolor{gray!20}22.8 & \cellcolor{gray!20}22.8 & \cellcolor{gray!20}22.8 & \cellcolor{gray!20}22.8 & \cellcolor{gray!20}.880 & \cellcolor{gray!20}.880 & \cellcolor{gray!20}.880 & \cellcolor{gray!20}.880 & \cellcolor{gray!20}.880 & \cellcolor{gray!20}.618 & \cellcolor{gray!20}.618 & \cellcolor{gray!20}.618 & \cellcolor{gray!20}.618 & \cellcolor{gray!20}.618 \\ \cmidrule{2-22}
 & FFT & \cellcolor{green!20}49.5 & \textbf{\cellcolor{green!20}44.2} & \cellcolor{green!20}28.0 & \cellcolor{green!20}48.6 & \cellcolor{green!20}43.6 & 21.8 & 20.6 & 20.1 & 21.1 & 20.5 & .842 & .829 & .852 & .843 & .841 & .574 & .582 & .586 & .578 & .579 \\
 & AdaLoRA & \cellcolor{green!20}47.6 & \cellcolor{green!20}33.1 & 14.1 & \cellcolor{green!20}39.8 & \cellcolor{green!20}36.2 & 22.4 & \underline{\cellcolor{green!20}22.9} & \textbf{22.6} & 22.1 & 22.1 & \textbf{.874} & \textbf{.878} & .871 & \underline{.860} & .847 & \textbf{.608} & \textbf{.614} & \textbf{.611} & .585 & .593 \\
 & HFT & \textbf{\cellcolor{green!20}52.5} & \cellcolor{green!20}43.7 & \cellcolor{green!20}35.8 & \cellcolor{green!20}48.4 & \cellcolor{green!20}45.4 & \underline{22.6} & 22.7 & 22.0 & 22.1 & 22.3 & .858 & .863 & .857 & .846 & .847 & .596 & .597 & .604 & \underline{.586} & .594 \\
 & GMT & \cellcolor{green!20}50.3 & \cellcolor{green!20}43.7 & \textbf{\cellcolor{green!20}37.8} & \cellcolor{green!20}49.1 & \textbf{\cellcolor{green!20}46.7} & 22.4 & 22.2 & 21.6 & 20.5 & 21.5 & .850 & .818 & .856 & .829 & .853 & .579 & .578 & .599 & .565 & .591 \\ \cmidrule{2-22}
 & SSU-Rand & \cellcolor{green!20}51.6 & \underline{\cellcolor{green!20}44.1} & \underline{\cellcolor{green!20}36.4} & \underline{\cellcolor{green!20}49.4} & \cellcolor{green!20}45.9 & \textbf{22.7} & \cellcolor{green!20}22.8 & 22.1 & \underline{22.2} & \underline{22.4} & .858 & .864 & \underline{.872} & .856 & \underline{.856} & .600 & .599 & .605 & .584 & \underline{.597} \\
 & SSU-Mag & \cellcolor{green!20}51.4 & \cellcolor{green!20}43.4 & \cellcolor{green!20}35.8 & \cellcolor{green!20}47.9 & \cellcolor{green!20}45.1 & 22.5 & 22.0 & 21.9 & 22.1 & 21.7 & .863 & .864 & .867 & .849 & .852 & .592 & .595 & .607 & .581 & .592 \\ \noalign{\vskip\aboverulesep}\cdashline{2-22}[2pt/1.2pt]\noalign{\vskip\belowrulesep}
 & SSU-Wanda & \underline{\cellcolor{green!20}52.3} & \cellcolor{green!20}43.9 & \underline{\cellcolor{green!20}36.4} & \textbf{\cellcolor{green!20}49.7} & \underline{\cellcolor{green!20}46.3} & \textbf{22.7} & \textbf{\cellcolor{green!20}23.1} & \underline{22.2} & \textbf{\cellcolor{green!20}22.9} & \textbf{\cellcolor{green!20}23.3} & \underline{.871} & \underline{.868} & \textbf{.874} & \textbf{.863} & \textbf{.870} & \underline{.606} & \underline{.608} & \underline{.609} & \textbf{.605} & \textbf{.603} \\
\midrule
\multirow{9}{*}{\rotatebox{90}{13B}} & \cellcolor{gray!20}Source & \cellcolor{gray!20}50.7 & \cellcolor{gray!20}30.5 & \cellcolor{gray!20}22.7 & \cellcolor{gray!20}31.0 & \cellcolor{gray!20}31.9 & \cellcolor{gray!20}24.5 & \cellcolor{gray!20}24.5 & \cellcolor{gray!20}24.5 & \cellcolor{gray!20}24.5 & \cellcolor{gray!20}24.5 & \cellcolor{gray!20}.897 & \cellcolor{gray!20}.897 & \cellcolor{gray!20}.897 & \cellcolor{gray!20}.897 & \cellcolor{gray!20}.897 & \cellcolor{gray!20}.665 & \cellcolor{gray!20}.665 & \cellcolor{gray!20}.665 & \cellcolor{gray!20}.665 & \cellcolor{gray!20}.665 \\ \cmidrule{2-22}
 & FFT & 49.7 & \cellcolor{green!20}39.2 & \cellcolor{green!20}39.2 & \cellcolor{green!20}43.5 & 28.8 & 21.5 & 8.6 & 19.0 & 14.4 & 14.8 & .890 & .891 & \textbf{\cellcolor{green!20}.901} & .891 & .889 & .650 & .643 & .657 & .650 & .637 \\
 & AdaLoRA & \cellcolor{green!20}52.1 & \cellcolor{green!20}33.1 & 19.8 & \cellcolor{green!20}40.6 & \cellcolor{green!20}37.2 & 24.1 & \textbf{\cellcolor{green!20}25.6} & \textbf{24.4} & \textbf{\cellcolor{green!20}24.7} & \underline{23.4} & \textbf{\cellcolor{green!20}.906} & \underline{\cellcolor{green!20}.901} & \cellcolor{green!20}.898 & .894 & \underline{.892} & \textbf{.662} & \textbf{.663} & .662 & \textbf{.660} & .651 \\
 & HFT & \underline{\cellcolor{green!20}55.1} & \cellcolor{green!20}38.6 & \underline{\cellcolor{green!20}41.6} & \underline{\cellcolor{green!20}50.1} & \cellcolor{green!20}35.1 & \underline{\cellcolor{green!20}24.5} & 20.5 & 22.7 & 16.8 & 18.8 & .897 & .896 & .893 & \textbf{\cellcolor{green!20}.899} & .888 & \underline{.659} & .652 & \textbf{.665} & .657 & \underline{.655} \\
 & GMT & 48.7 & \cellcolor{green!20}37.1 & \cellcolor{green!20}23.2 & \cellcolor{green!20}45.2 & \cellcolor{green!20}33.4 & 23.4 & 12.9 & 15.9 & 14.1 & 16.4 & .892 & .893 & \underline{\cellcolor{green!20}.900} & .896 & \textbf{.897} & .653 & .658 & .660 & .654 & .643 \\ \cmidrule{2-22}
 & SSU-Rand & \cellcolor{green!20}54.4 & \underline{\cellcolor{green!20}39.7} & \cellcolor{green!20}36.3 & \cellcolor{green!20}49.7 & \underline{\cellcolor{green!20}39.6} & \textbf{\cellcolor{green!20}24.9} & 23.6 & 22.9 & 16.6 & 20.4 & .897 & \textbf{\cellcolor{green!20}.903} & \underline{\cellcolor{green!20}.900} & .897 & .891 & .658 & .654 & .663 & .653 & .653 \\
 & SSU-Mag & \cellcolor{green!20}53.4 & \cellcolor{green!20}37.4 & \cellcolor{green!20}32.5 & \cellcolor{green!20}45.9 & 31.5 & 24.4 & 20.6 & 20.7 & 16.8 & 18.6 & .893 & .896 & .896 & .894 & .883 & \underline{.659} & .656 & .662 & \underline{.659} & .647 \\ \noalign{\vskip\aboverulesep}\cdashline{2-22}[2pt/1.2pt]\noalign{\vskip\belowrulesep}
 & SSU-Wanda & \textbf{\cellcolor{green!20}55.7} & \textbf{\cellcolor{green!20}45.1} & \textbf{\cellcolor{green!20}43.8} & \textbf{\cellcolor{green!20}51.4} & \textbf{\cellcolor{green!20}45.1} & 24.4 & \underline{\cellcolor{green!20}25.3} & \underline{24.0} & \underline{23.8} & \textbf{23.8} & \underline{\cellcolor{green!20}.898} & \underline{\cellcolor{green!20}.901} & .893 & \underline{\cellcolor{green!20}.898} & \textbf{.897} & \textbf{.662} & \underline{.660} & \underline{.664} & \underline{.659} & \textbf{.659} \\

\bottomrule
\end{tabular}
}
\caption{Performance on source language (English) tasks.
Scores that are better than Source are highlighted in \colorbox{green!20}{green}.
The best and second-best adaptation approaches for each model scale are indicated in \textbf{bold} and \underline{underlined}, respectively.}
\label{tab:source}
\end{table*}

\begin{table*}[t]
\centering
\small
\resizebox{\textwidth}{!}{
\begin{tabular}{llc@{\hspace{6pt}}c@{\hspace{6pt}}c@{\hspace{6pt}}c@{\hspace{6pt}}c@{\hspace{6pt}}c@{\hspace{6pt}}c@{\hspace{6pt}}c@{\hspace{6pt}}c@{\hspace{6pt}}c@{\hspace{6pt}}c@{\hspace{6pt}}c@{\hspace{6pt}}c@{\hspace{6pt}}c@{\hspace{6pt}}c@{\hspace{6pt}}c@{\hspace{6pt}}c@{\hspace{6pt}}c@{\hspace{6pt}}c@{\hspace{6pt}}c@{\hspace{6pt}}c}
\toprule
& & \multicolumn{5}{c}{\textbf{MT}} & \multicolumn{5}{c}{\textbf{SUM}} & \multicolumn{5}{c}{\textbf{MRC}} & \multicolumn{5}{c}{\textbf{MMLU}} \\
\cmidrule(lr){3-7} \cmidrule(lr){8-12} \cmidrule(lr){13-17} \cmidrule(lr){18-22}
\multicolumn{2}{l}{\textbf{Approach}} & ne & ky & am & ha & ig & ne & ky & am & ha & ig & ne & ky & am & ha & ig & ne & ky & am & ha & ig \\
\midrule

\multirow{9}{*}{\rotatebox{90}{7B}} & \cellcolor{gray!20}Source & \cellcolor{gray!20}27.0 & \cellcolor{gray!20}21.1 & \cellcolor{gray!20}5.1 & \cellcolor{gray!20}24.4 & \cellcolor{gray!20}23.0 & \cellcolor{gray!20}22.4 & \cellcolor{gray!20}22.9 & \cellcolor{gray!20}8.6 & \cellcolor{gray!20}23.7 & \cellcolor{gray!20}23.3 & \cellcolor{gray!20}.382 & \cellcolor{gray!20}.379 & \cellcolor{gray!20}.276 & \cellcolor{gray!20}.332 & \cellcolor{gray!20}.301 & \cellcolor{gray!20}.301 & \cellcolor{gray!20}.301 & \cellcolor{gray!20}.276 & \cellcolor{gray!20}.321 & \cellcolor{gray!20}.323 \\ \cmidrule{2-22}
 & FFT & \cellcolor{green!20}32.5 & \textbf{\cellcolor{green!20}33.8} & \textbf{\cellcolor{green!20}12.1} & \cellcolor{green!20}38.6 & \cellcolor{green!20}36.7 & 22.1 & \cellcolor{green!20}23.7 & \underline{\cellcolor{green!20}9.3} & \cellcolor{green!20}32.2 & \underline{\cellcolor{green!20}26.4} & .360 & \underline{\cellcolor{green!20}.441} & \cellcolor{green!20}.309 & \textbf{\cellcolor{green!20}.460} & \cellcolor{green!20}.396 & .293 & \cellcolor{green!20}.312 & \cellcolor{green!20}.288 & \textbf{\cellcolor{green!20}.372} & \cellcolor{green!20}.360 \\
 & AdaLoRA & \cellcolor{green!20}28.1 & \cellcolor{green!20}22.3 & 4.0 & 22.9 & 22.3 & 21.7 & \cellcolor{green!20}23.1 & 6.5 & \cellcolor{green!20}31.6 & \textbf{\cellcolor{green!20}26.6} & .351 & .343 & .276 & .328 & .291 & \underline{\cellcolor{green!20}.309} & \cellcolor{green!20}.311 & .272 & .278 & \cellcolor{green!20}.324 \\
 & HFT & \cellcolor{green!20}32.7 & \cellcolor{green!20}32.4 & \cellcolor{green!20}9.6 & \cellcolor{green!20}37.5 & \cellcolor{green!20}36.9 & \textbf{22.4} & \underline{\cellcolor{green!20}23.8} & 8.6 & \cellcolor{green!20}32.1 & \cellcolor{green!20}26.3 & .368 & \cellcolor{green!20}.411 & \cellcolor{green!20}.282 & \cellcolor{green!20}.438 & \cellcolor{green!20}.388 & .293 & \underline{\cellcolor{green!20}.314} & \cellcolor{green!20}.287 & \cellcolor{green!20}.346 & \textbf{\cellcolor{green!20}.373} \\
 & GMT & \cellcolor{green!20}32.3 & \underline{\cellcolor{green!20}33.5} & \underline{\cellcolor{green!20}11.6} & \underline{\cellcolor{green!20}39.0} & \textbf{\cellcolor{green!20}38.3} & \underline{22.3} & \underline{\cellcolor{green!20}23.8} & \textbf{\cellcolor{green!20}9.9} & \textbf{\cellcolor{green!20}32.4} & \cellcolor{green!20}26.2 & .346 & \cellcolor{green!20}.419 & \underline{\cellcolor{green!20}.312} & \cellcolor{green!20}.451 & \underline{\cellcolor{green!20}.398} & .279 & \cellcolor{green!20}.308 & \textbf{\cellcolor{green!20}.296} & \cellcolor{green!20}.353 & \cellcolor{green!20}.361 \\ \cmidrule{2-22}
 & SSU-Rand & \underline{\cellcolor{green!20}33.2} & \cellcolor{green!20}32.6 & \cellcolor{green!20}9.5 & \cellcolor{green!20}38.4 & \underline{\cellcolor{green!20}37.3} & \textbf{22.4} & \underline{\cellcolor{green!20}23.8} & \cellcolor{green!20}8.8 & \cellcolor{green!20}32.2 & \underline{\cellcolor{green!20}26.4} & \underline{\cellcolor{green!20}.388} & \cellcolor{green!20}.428 & \cellcolor{green!20}.299 & \underline{\cellcolor{green!20}.457} & \textbf{\cellcolor{green!20}.401} & \cellcolor{green!20}.305 & \cellcolor{green!20}.311 & \cellcolor{green!20}.288 & \underline{\cellcolor{green!20}.362} & \cellcolor{green!20}.355 \\
 & SSU-Mag & \cellcolor{green!20}33.1 & \cellcolor{green!20}32.2 & \cellcolor{green!20}9.7 & \cellcolor{green!20}37.1 & \cellcolor{green!20}36.6 & 22.2 & \cellcolor{green!20}23.7 & \cellcolor{green!20}9.2 & \underline{\cellcolor{green!20}32.3} & \cellcolor{green!20}26.2 & .372 & \cellcolor{green!20}.418 & \cellcolor{green!20}.297 & \cellcolor{green!20}.451 & \cellcolor{green!20}.379 & \cellcolor{green!20}.303 & \cellcolor{green!20}.307 & \underline{\cellcolor{green!20}.291} & \cellcolor{green!20}.346 & \cellcolor{green!20}.348 \\ \noalign{\vskip\aboverulesep}\cdashline{2-22}[2pt/1.2pt]\noalign{\vskip\belowrulesep}
 & SSU-Wanda & \textbf{\cellcolor{green!20}34.0} & \cellcolor{green!20}32.2 & \cellcolor{green!20}9.0 & \textbf{\cellcolor{green!20}42.6} & \cellcolor{green!20}37.1 & \textbf{22.4} & \textbf{\cellcolor{green!20}24.2} & \cellcolor{green!20}8.9 & \cellcolor{green!20}32.2 & \cellcolor{green!20}26.3 & \textbf{\cellcolor{green!20}.401} & \textbf{\cellcolor{green!20}.458} & \textbf{\cellcolor{green!20}.316} & \cellcolor{green!20}.439 & \textbf{\cellcolor{green!20}.401} & \textbf{\cellcolor{green!20}.313} & \textbf{\cellcolor{green!20}.329} & \textbf{\cellcolor{green!20}.296} & \cellcolor{green!20}.355 & \underline{\cellcolor{green!20}.371} \\
\midrule
\multirow{9}{*}{\rotatebox{90}{13B}} & \cellcolor{gray!20}Source & \cellcolor{gray!20}32.4 & \cellcolor{gray!20}22.5 & \cellcolor{gray!20}6.0 & \cellcolor{gray!20}25.3 & \cellcolor{gray!20}25.7 & \cellcolor{gray!20}22.9 & \cellcolor{gray!20}23.2 & \cellcolor{gray!20}10.0 & \cellcolor{gray!20}25.3 & \cellcolor{gray!20}22.4 & \cellcolor{gray!20}.501 & \cellcolor{gray!20}.393 & \cellcolor{gray!20}.318 & \cellcolor{gray!20}.348 & \cellcolor{gray!20}.310 & \cellcolor{gray!20}.345 & \cellcolor{gray!20}.322 & \cellcolor{gray!20}.293 & \cellcolor{gray!20}.333 & \cellcolor{gray!20}.351 \\ \cmidrule{2-22}
 & FFT & \cellcolor{green!20}37.5 & \textbf{\cellcolor{green!20}36.9} & \textbf{\cellcolor{green!20}16.5} & \cellcolor{green!20}40.2 & \cellcolor{green!20}37.1 & 21.8 & \underline{\cellcolor{green!20}23.7} & \underline{\cellcolor{green!20}10.6} & \underline{\cellcolor{green!20}32.7} & \cellcolor{green!20}25.4 & .500 & \textbf{\cellcolor{green!20}.564} & \textbf{\cellcolor{green!20}.381} & \textbf{\cellcolor{green!20}.579} & \cellcolor{green!20}.438 & .342 & \cellcolor{green!20}.335 & \underline{\cellcolor{green!20}.315} & \textbf{\cellcolor{green!20}.417} & \textbf{\cellcolor{green!20}.397} \\
 & AdaLoRA & \cellcolor{green!20}33.7 & \cellcolor{green!20}24.0 & 5.7 & \cellcolor{green!20}26.3 & 25.4 & 22.2 & 22.9 & 9.4 & \cellcolor{green!20}31.6 & \cellcolor{green!20}25.4 & .448 & .391 & .293 & \cellcolor{green!20}.371 & \cellcolor{green!20}.322 & .340 & .307 & .277 & .324 & .307 \\
 & HFT & \underline{\cellcolor{green!20}37.6} & \cellcolor{green!20}36.3 & \cellcolor{green!20}14.4 & \cellcolor{green!20}41.6 & \underline{\cellcolor{green!20}38.4} & 21.9 & \cellcolor{green!20}23.4 & \cellcolor{green!20}10.4 & \cellcolor{green!20}32.4 & \textbf{\cellcolor{green!20}26.1} & .498 & \cellcolor{green!20}.538 & \cellcolor{green!20}.376 & \cellcolor{green!20}.538 & \cellcolor{green!20}.429 & \cellcolor{green!20}.348 & \cellcolor{green!20}.356 & \cellcolor{green!20}.312 & \cellcolor{green!20}.384 & \cellcolor{green!20}.375 \\
 & GMT & \cellcolor{green!20}37.3 & \underline{\cellcolor{green!20}36.6} & \textbf{\cellcolor{green!20}16.5} & \cellcolor{green!20}40.2 & \cellcolor{green!20}36.8 & 22.0 & \cellcolor{green!20}23.4 & 9.8 & \underline{\cellcolor{green!20}32.7} & \underline{\cellcolor{green!20}26.0} & \underline{.501} & \underline{\cellcolor{green!20}.559} & \cellcolor{green!20}.355 & \cellcolor{green!20}.530 & \cellcolor{green!20}.420 & \cellcolor{green!20}.348 & \cellcolor{green!20}.356 & \textbf{\cellcolor{green!20}.318} & \underline{\cellcolor{green!20}.404} & .338 \\ \cmidrule{2-22}
 & SSU-Rand & \cellcolor{green!20}37.5 & \cellcolor{green!20}36.1 & \underline{\cellcolor{green!20}14.5} & \underline{\cellcolor{green!20}41.8} & \cellcolor{green!20}37.9 & \underline{22.3} & \cellcolor{green!20}23.4 & \cellcolor{green!20}10.4 & \textbf{\cellcolor{green!20}32.9} & \textbf{\cellcolor{green!20}26.1} & .492 & \cellcolor{green!20}.556 & \cellcolor{green!20}.364 & \cellcolor{green!20}.540 & \underline{\cellcolor{green!20}.440} & \underline{\cellcolor{green!20}.352} & \textbf{\cellcolor{green!20}.361} & \cellcolor{green!20}.313 & \cellcolor{green!20}.383 & \cellcolor{green!20}.369 \\
 & SSU-Mag & \cellcolor{green!20}37.2 & \cellcolor{green!20}36.1 & \underline{\cellcolor{green!20}14.5} & \cellcolor{green!20}39.7 & \cellcolor{green!20}36.5 & 22.0 & 23.0 & 9.7 & \cellcolor{green!20}32.1 & \underline{\cellcolor{green!20}26.0} & .474 & \cellcolor{green!20}.533 & \cellcolor{green!20}.361 & \underline{\cellcolor{green!20}.546} & \cellcolor{green!20}.419 & \cellcolor{green!20}.345 & \underline{\cellcolor{green!20}.357} & \cellcolor{green!20}.311 & \cellcolor{green!20}.394 & .342 \\ \noalign{\vskip\aboverulesep}\cdashline{2-22}[2pt/1.2pt]\noalign{\vskip\belowrulesep}
 & SSU-Wanda & \textbf{\cellcolor{green!20}37.9} & \cellcolor{green!20}35.7 & \cellcolor{green!20}13.7 & \textbf{\cellcolor{green!20}44.0} & \textbf{\cellcolor{green!20}39.1} & \textbf{22.8} & \textbf{\cellcolor{green!20}23.8} & \textbf{\cellcolor{green!20}11.0} & \cellcolor{green!20}32.3 & \cellcolor{green!20}25.9 & \textbf{\cellcolor{green!20}.520} & \cellcolor{green!20}.549 & \underline{\cellcolor{green!20}.377} & \cellcolor{green!20}.542 & \textbf{\cellcolor{green!20}.441} & \textbf{\cellcolor{green!20}.354} & \cellcolor{green!20}.355 & \cellcolor{green!20}.302 & \cellcolor{green!20}.390 & \underline{\cellcolor{green!20}.395} \\

\bottomrule
\end{tabular}
}
\caption{Performance on target language tasks.
Scores that are better than Source are highlighted in \colorbox{green!20}{green}.
The best and second-best adaptation approaches for each model scale are indicated in \textbf{bold} and \underline{underlined}, respectively.}
\label{tab:target}
\end{table*}

\clearpage

\begin{figure*}[th]
\centering
\includegraphics[width=\textwidth]{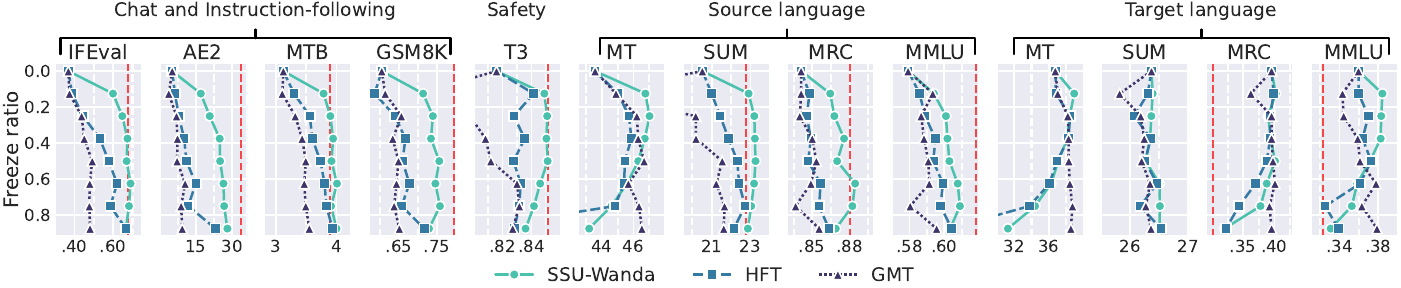}
\caption{
Model performance (SSU-Wanda, HFT, GMT) on Igbo as target language across freezing ratios.
The dashed red line indicates Source performance (omitted for MT and SUM due to very low scores).
Some data points for HFT and GMT are also omitted due to extremely low performance.
}
\label{fig:ratio_appendix}
\end{figure*}

\begin{table*}[th]
\centering
\small
\resizebox{\textwidth}{!}{
\begin{tabular}{lc@{\hspace{6pt}}c@{\hspace{6pt}}c@{\hspace{6pt}}c@{\hspace{6pt}}c@{\hspace{6pt}}c@{\hspace{6pt}}c@{\hspace{6pt}}c@{\hspace{6pt}}c@{\hspace{6pt}}c@{\hspace{6pt}}c@{\hspace{6pt}}c@{\hspace{6pt}}c}
\toprule
& \multicolumn{4}{c}{\textbf{Chat and Instruction-following}} & \textbf{Safety} & \multicolumn{4}{c}{\textbf{Source language}} & \multicolumn{4}{c}{\textbf{Target language} (Igbo)} \\
\cmidrule(lr){2-5} \cmidrule(lr){7-10} \cmidrule(lr){11-14}
\textbf{Approach} & \textbf{IFEval} & \textbf{AE2} & \textbf{MTB} & \textbf{GSM8K} & \textbf{T3} & \textbf{MT} & \textbf{SUM} & \textbf{MRC} & \textbf{MMLU} & \textbf{MT} & \textbf{SUM} & \textbf{MRC} & \textbf{MMLU} \\
\midrule

\cellcolor{gray!20}Source & \cellcolor{gray!20}.675 & \cellcolor{gray!20}32.6 & \cellcolor{gray!20}3.98 & \cellcolor{gray!20}.796 & \cellcolor{gray!20}.851 & \cellcolor{gray!20}28.5 & \cellcolor{gray!20}22.8 & \cellcolor{gray!20}.880 & \cellcolor{gray!20}.618 & \cellcolor{gray!20}23.0 & \cellcolor{gray!20}23.3 & \cellcolor{gray!20}.301 & \cellcolor{gray!20}.323 \\ \midrule
500 examples (Default) & .670 & \textbf{25.0} & \textbf{3.92} & \textbf{.756} & .851 & 46.3 & \textbf{23.3} & .870 & \textbf{.603} & 37.1 & \textbf{26.3} & .401 & \textbf{.371} \\
128 examples & \textbf{.682} & 24.3 & 3.89 & .754 & \textbf{.852} & \textbf{46.4} & 23.2 & \textbf{.873} & .600 & \textbf{37.2} & \textbf{26.3} & \textbf{.410} & \textbf{.371} \\

\bottomrule
\end{tabular}
}
\caption{Performance of SSU-Wanda with different number of calibration samples.
We use Igbo as the target language and tulu-3-sft-olmo-2-mixture as the calibration dataset.
\textbf{Bold} indicates best adaptation approach.}
\label{tab:calibration_size}
\end{table*}

\begin{table*}[th]
\centering
\small
\resizebox{\textwidth}{!}{
\begin{tabular}{lc@{\hspace{6pt}}c@{\hspace{6pt}}c@{\hspace{6pt}}c@{\hspace{6pt}}c@{\hspace{6pt}}c@{\hspace{6pt}}c@{\hspace{6pt}}c@{\hspace{6pt}}c@{\hspace{6pt}}c@{\hspace{6pt}}c@{\hspace{6pt}}c@{\hspace{6pt}}c}
\toprule
& \multicolumn{4}{c}{\textbf{Chat and Instruction-following}} & \textbf{Safety} & \multicolumn{4}{c}{\textbf{Source language}} & \multicolumn{4}{c}{\textbf{Target language} (Igbo)} \\
\cmidrule(lr){2-5} \cmidrule(lr){7-10} \cmidrule(lr){11-14}
\textbf{Approach} & \textbf{IFEval} & \textbf{AE2} & \textbf{MTB} & \textbf{GSM8K} & \textbf{T3} & \textbf{MT} & \textbf{SUM} & \textbf{MRC} & \textbf{MMLU} & \textbf{MT} & \textbf{SUM} & \textbf{MRC} & \textbf{MMLU} \\
\midrule

\cellcolor{gray!20}\raisebox{-0.5ex}{Source} & \cellcolor{gray!20}.675\raisebox{-1.25ex}{\hspace{0.1em}{\scriptsize +0.0}} & \cellcolor{gray!20}32.6\raisebox{-1.25ex}{\hspace{0.1em}{\scriptsize +0.0}} & \cellcolor{gray!20}3.98\raisebox{-1.25ex}{\hspace{0.1em}{\scriptsize +0.0}} & \cellcolor{gray!20}.796\raisebox{-1.25ex}{\hspace{0.1em}{\scriptsize +0.0}} & \cellcolor{gray!20}.851\raisebox{-1.25ex}{\hspace{0.1em}{\scriptsize +0.0}} & \cellcolor{gray!20}28.5\raisebox{-1.25ex}{\hspace{0.1em}{\scriptsize +0.0}} & \cellcolor{gray!20}22.8\raisebox{-1.25ex}{\hspace{0.1em}{\scriptsize +0.0}} & \cellcolor{gray!20}.880\raisebox{-1.25ex}{\hspace{0.1em}{\scriptsize +0.0}} & \cellcolor{gray!20}.618\raisebox{-1.25ex}{\hspace{0.1em}{\scriptsize +0.0}} & \cellcolor{gray!20}23.0\raisebox{-1.25ex}{\hspace{0.1em}{\scriptsize +0.0}} & \cellcolor{gray!20}23.3\raisebox{-1.25ex}{\hspace{0.1em}{\scriptsize +0.0}} & \cellcolor{gray!20}.301\raisebox{-1.25ex}{\hspace{0.1em}{\scriptsize +0.0}} & \cellcolor{gray!20}.323\raisebox{-1.25ex}{\hspace{0.1em}{\scriptsize +0.0}} \\ \midrule
\raisebox{-0.5ex}{SSU-Wanda} & \underline{.670}\raisebox{-1.25ex}{\hspace{0.1em}{\scriptsize -0.7}} & \underline{25.0}\raisebox{-1.25ex}{\hspace{0.1em}{\scriptsize -23.2}} & 3.92\raisebox{-1.25ex}{\hspace{0.1em}{\scriptsize -1.5}} & \underline{.756}\raisebox{-1.25ex}{\hspace{0.1em}{\scriptsize -5.0}} & \textbf{.851}\raisebox{-1.25ex}{\hspace{0.1em}{\scriptsize -0.0}} & \textbf{46.3}\raisebox{-1.25ex}{\hspace{0.1em}{\scriptsize +62.7}} & \textbf{23.3}\raisebox{-1.25ex}{\hspace{0.1em}{\scriptsize +2.3}} & \underline{.870}\raisebox{-1.25ex}{\hspace{0.1em}{\scriptsize -1.1}} & .603\raisebox{-1.25ex}{\hspace{0.1em}{\scriptsize -2.4}} & \underline{37.1}\raisebox{-1.25ex}{\hspace{0.1em}{\scriptsize +61.7}} & \underline{26.3}\raisebox{-1.25ex}{\hspace{0.1em}{\scriptsize +12.9}} & \underline{.401}\raisebox{-1.25ex}{\hspace{0.1em}{\scriptsize +33.2}} & \underline{.371}\raisebox{-1.25ex}{\hspace{0.1em}{\scriptsize +14.9}} \\ \midrule
\raisebox{-0.5ex}{LoTA (90\% Sparsity)} & .638\raisebox{-1.25ex}{\hspace{0.1em}{\scriptsize -5.4}} & 20.4\raisebox{-1.25ex}{\hspace{0.1em}{\scriptsize -37.4}} & \underline{3.98}\raisebox{-1.25ex}{\hspace{0.1em}{\scriptsize +0.0}} & .706\raisebox{-1.25ex}{\hspace{0.1em}{\scriptsize -11.3}} & .827\raisebox{-1.25ex}{\hspace{0.1em}{\scriptsize -2.8}} & 45.2\raisebox{-1.25ex}{\hspace{0.1em}{\scriptsize +58.8}} & \underline{22.7}\raisebox{-1.25ex}{\hspace{0.1em}{\scriptsize -0.3}} & \underline{.864}\raisebox{-1.25ex}{\hspace{0.1em}{\scriptsize -1.8}} & \textbf{.606}\raisebox{-1.25ex}{\hspace{0.1em}{\scriptsize -2.0}} & 34.4\raisebox{-1.25ex}{\hspace{0.1em}{\scriptsize +49.9}} & 26.2\raisebox{-1.25ex}{\hspace{0.1em}{\scriptsize +12.5}} & .366\raisebox{-1.25ex}{\hspace{0.1em}{\scriptsize +21.5}} & .360\raisebox{-1.25ex}{\hspace{0.1em}{\scriptsize +11.5}} \\
\raisebox{-0.5ex}{LoTA (50\% Sparsity)} & .449\raisebox{-1.25ex}{\hspace{0.1em}{\scriptsize -33.4}} & 8.3\raisebox{-1.25ex}{\hspace{0.1em}{\scriptsize -74.5}} & 3.45\raisebox{-1.25ex}{\hspace{0.1em}{\scriptsize -13.3}} & .636\raisebox{-1.25ex}{\hspace{0.1em}{\scriptsize -20.1}} & .824\raisebox{-1.25ex}{\hspace{0.1em}{\scriptsize -3.2}} & \underline{45.8}\raisebox{-1.25ex}{\hspace{0.1em}{\scriptsize +60.9}} & 21.5\raisebox{-1.25ex}{\hspace{0.1em}{\scriptsize -5.6}} & .844\raisebox{-1.25ex}{\hspace{0.1em}{\scriptsize -4.1}} & .590\raisebox{-1.25ex}{\hspace{0.1em}{\scriptsize -4.6}} & \textbf{37.8}\raisebox{-1.25ex}{\hspace{0.1em}{\scriptsize +64.7}} & \textbf{26.4}\raisebox{-1.25ex}{\hspace{0.1em}{\scriptsize +13.4}} & \textbf{.402}\raisebox{-1.25ex}{\hspace{0.1em}{\scriptsize +33.5}} & \textbf{.372}\raisebox{-1.25ex}{\hspace{0.1em}{\scriptsize +15.2}} \\
\raisebox{-0.5ex}{S2FT (Down)} & \textbf{.695}\raisebox{-1.25ex}{\hspace{0.1em}{\scriptsize +3.0}} & \textbf{27.9}\raisebox{-1.25ex}{\hspace{0.1em}{\scriptsize -14.3}} & \textbf{3.99}\raisebox{-1.25ex}{\hspace{0.1em}{\scriptsize +0.3}} & \underline{.732}\raisebox{-1.25ex}{\hspace{0.1em}{\scriptsize -8.0}} & \underline{.834}\raisebox{-1.25ex}{\hspace{0.1em}{\scriptsize -2.0}} & 36.7\raisebox{-1.25ex}{\hspace{0.1em}{\scriptsize +29.0}} & 22.6\raisebox{-1.25ex}{\hspace{0.1em}{\scriptsize -0.7}} & .857\raisebox{-1.25ex}{\hspace{0.1em}{\scriptsize -2.6}} & \underline{.603}\raisebox{-1.25ex}{\hspace{0.1em}{\scriptsize -2.4}} & 21.7\raisebox{-1.25ex}{\hspace{0.1em}{\scriptsize -5.4}} & 26.0\raisebox{-1.25ex}{\hspace{0.1em}{\scriptsize +11.6}} & .303\raisebox{-1.25ex}{\hspace{0.1em}{\scriptsize +0.6}} & .331\raisebox{-1.25ex}{\hspace{0.1em}{\scriptsize +2.5}} \\

\bottomrule
\end{tabular}
}
\caption{Performance of additional adaptation baselines: LoTA and S2FT using Igbo as the target. 
\textbf{Bold} and \underline{underlined} denote best and second-best adaptation approaches with relative changes (\%) in subscripts.}
\label{tab:additional_baselines}
\end{table*}

\begin{table*}[th]
\centering
\scriptsize
\resizebox{\textwidth}{!}{
\begin{tabular}{l@{\hspace{4pt}}c@{\hspace{6pt}}c@{\hspace{6pt}}c@{\hspace{6pt}}c@{\hspace{6pt}}c@{\hspace{6pt}}c@{\hspace{6pt}}c@{\hspace{6pt}}c@{\hspace{6pt}}c@{\hspace{6pt}}c@{\hspace{6pt}}c@{\hspace{6pt}}c@{\hspace{6pt}}c}
\toprule
& \multicolumn{4}{c}{\textbf{Chat and Instruction-following}} & \textbf{Safety} & \multicolumn{4}{c}{\textbf{Source language}} & \multicolumn{4}{c}{\textbf{Target language} (Igbo)} \\
\cmidrule(lr){2-5} \cmidrule(lr){7-10} \cmidrule(lr){11-14}
\textbf{Approach} & \textbf{IFEval} & \textbf{AE2} & \textbf{MTB} & \textbf{GSM8K} & \textbf{T3} & \textbf{MT} & \textbf{SUM} & \textbf{MRC} & \textbf{MMLU} & \textbf{MT} & \textbf{SUM} & \textbf{MRC} & \textbf{MMLU} \\
\midrule

\cellcolor{gray!20}\raisebox{-0.5ex}{Source} & \cellcolor{gray!20}.675\raisebox{-1.25ex}{\hspace{0.1em}{\scriptsize +0.0}} & \cellcolor{gray!20}32.6\raisebox{-1.25ex}{\hspace{0.1em}{\scriptsize +0.0}} & \cellcolor{gray!20}3.98\raisebox{-1.25ex}{\hspace{0.1em}{\scriptsize +0.0}} & \cellcolor{gray!20}.796\raisebox{-1.25ex}{\hspace{0.1em}{\scriptsize +0.0}} & \cellcolor{gray!20}.851\raisebox{-1.25ex}{\hspace{0.1em}{\scriptsize +0.0}} & \cellcolor{gray!20}28.5\raisebox{-1.25ex}{\hspace{0.1em}{\scriptsize +0.0}} & \cellcolor{gray!20}22.8\raisebox{-1.25ex}{\hspace{0.1em}{\scriptsize +0.0}} & \cellcolor{gray!20}.880\raisebox{-1.25ex}{\hspace{0.1em}{\scriptsize +0.0}} & \cellcolor{gray!20}.618\raisebox{-1.25ex}{\hspace{0.1em}{\scriptsize +0.0}} & \cellcolor{gray!20}23.0\raisebox{-1.25ex}{\hspace{0.1em}{\scriptsize +0.0}} & \cellcolor{gray!20}23.3\raisebox{-1.25ex}{\hspace{0.1em}{\scriptsize +0.0}} & \cellcolor{gray!20}.301\raisebox{-1.25ex}{\hspace{0.1em}{\scriptsize +0.0}} & \cellcolor{gray!20}.323\raisebox{-1.25ex}{\hspace{0.1em}{\scriptsize +0.0}} \\ \midrule
\raisebox{-0.5ex}{SSU-Wanda} & .670\raisebox{-1.25ex}{\hspace{0.1em}{\scriptsize -0.7}} & 25.0\raisebox{-1.25ex}{\hspace{0.1em}{\scriptsize -23.2}} & 3.92\raisebox{-1.25ex}{\hspace{0.1em}{\scriptsize -1.5}} & \textbf{.756}\raisebox{-1.25ex}{\hspace{0.1em}{\scriptsize -5.0}} & \textbf{.851}\raisebox{-1.25ex}{\hspace{0.1em}{\scriptsize -0.0}} & \underline{46.3}\raisebox{-1.25ex}{\hspace{0.1em}{\scriptsize +62.7}} & \textbf{23.3}\raisebox{-1.25ex}{\hspace{0.1em}{\scriptsize +2.3}} & \textbf{.870}\raisebox{-1.25ex}{\hspace{0.1em}{\scriptsize -1.1}} & \underline{.603}\raisebox{-1.25ex}{\hspace{0.1em}{\scriptsize -2.4}} & 37.1\raisebox{-1.25ex}{\hspace{0.1em}{\scriptsize +61.7}} & 26.3\raisebox{-1.25ex}{\hspace{0.1em}{\scriptsize +12.9}} & \underline{.401}\raisebox{-1.25ex}{\hspace{0.1em}{\scriptsize +33.2}} & .371\raisebox{-1.25ex}{\hspace{0.1em}{\scriptsize +14.9}} \\ \midrule
\raisebox{-0.5ex}{LoTA (12.5\%)} & .367\raisebox{-1.25ex}{\hspace{0.1em}{\scriptsize -45.6}} & 5.4\raisebox{-1.25ex}{\hspace{0.1em}{\scriptsize -83.4}} & 3.10\raisebox{-1.25ex}{\hspace{0.1em}{\scriptsize -22.1}} & .590\raisebox{-1.25ex}{\hspace{0.1em}{\scriptsize -25.9}} & .811\raisebox{-1.25ex}{\hspace{0.1em}{\scriptsize -4.7}} & 42.1\raisebox{-1.25ex}{\hspace{0.1em}{\scriptsize +47.9}} & 20.4\raisebox{-1.25ex}{\hspace{0.1em}{\scriptsize -10.4}} & .857\raisebox{-1.25ex}{\hspace{0.1em}{\scriptsize -2.6}} & .587\raisebox{-1.25ex}{\hspace{0.1em}{\scriptsize -5.0}} & 37.1\raisebox{-1.25ex}{\hspace{0.1em}{\scriptsize +61.7}} & 26.3\raisebox{-1.25ex}{\hspace{0.1em}{\scriptsize +12.9}} & \textbf{.402}\raisebox{-1.25ex}{\hspace{0.1em}{\scriptsize +33.5}} & \textbf{.374}\raisebox{-1.25ex}{\hspace{0.1em}{\scriptsize +15.8}} \\
\raisebox{-0.5ex}{LoTA (25.0\%)} & .366\raisebox{-1.25ex}{\hspace{0.1em}{\scriptsize -45.8}} & 5.0\raisebox{-1.25ex}{\hspace{0.1em}{\scriptsize -84.6}} & 3.09\raisebox{-1.25ex}{\hspace{0.1em}{\scriptsize -22.3}} & .590\raisebox{-1.25ex}{\hspace{0.1em}{\scriptsize -25.9}} & .812\raisebox{-1.25ex}{\hspace{0.1em}{\scriptsize -4.6}} & 42.2\raisebox{-1.25ex}{\hspace{0.1em}{\scriptsize +48.3}} & 20.4\raisebox{-1.25ex}{\hspace{0.1em}{\scriptsize -10.4}} & .857\raisebox{-1.25ex}{\hspace{0.1em}{\scriptsize -2.6}} & .587\raisebox{-1.25ex}{\hspace{0.1em}{\scriptsize -5.0}} & 37.1\raisebox{-1.25ex}{\hspace{0.1em}{\scriptsize +61.7}} & \underline{26.4}\raisebox{-1.25ex}{\hspace{0.1em}{\scriptsize +13.4}} & \textbf{.402}\raisebox{-1.25ex}{\hspace{0.1em}{\scriptsize +33.5}} & \textbf{.374}\raisebox{-1.25ex}{\hspace{0.1em}{\scriptsize +15.8}} \\
\raisebox{-0.5ex}{LoTA (37.5\%)} & .367\raisebox{-1.25ex}{\hspace{0.1em}{\scriptsize -45.6}} & 4.9\raisebox{-1.25ex}{\hspace{0.1em}{\scriptsize -85.0}} & 3.02\raisebox{-1.25ex}{\hspace{0.1em}{\scriptsize -24.1}} & .590\raisebox{-1.25ex}{\hspace{0.1em}{\scriptsize -25.9}} & .811\raisebox{-1.25ex}{\hspace{0.1em}{\scriptsize -4.7}} & 42.5\raisebox{-1.25ex}{\hspace{0.1em}{\scriptsize +49.3}} & 20.4\raisebox{-1.25ex}{\hspace{0.1em}{\scriptsize -10.4}} & .857\raisebox{-1.25ex}{\hspace{0.1em}{\scriptsize -2.6}} & .587\raisebox{-1.25ex}{\hspace{0.1em}{\scriptsize -5.0}} & 37.2\raisebox{-1.25ex}{\hspace{0.1em}{\scriptsize +62.1}} & \textbf{26.5}\raisebox{-1.25ex}{\hspace{0.1em}{\scriptsize +13.8}} & \textbf{.402}\raisebox{-1.25ex}{\hspace{0.1em}{\scriptsize +33.5}} & \textbf{.374}\raisebox{-1.25ex}{\hspace{0.1em}{\scriptsize +15.8}} \\
\raisebox{-0.5ex}{LoTA (50.0\%)} & .449\raisebox{-1.25ex}{\hspace{0.1em}{\scriptsize -33.4}} & 8.3\raisebox{-1.25ex}{\hspace{0.1em}{\scriptsize -74.5}} & 3.45\raisebox{-1.25ex}{\hspace{0.1em}{\scriptsize -13.3}} & .636\raisebox{-1.25ex}{\hspace{0.1em}{\scriptsize -20.1}} & .824\raisebox{-1.25ex}{\hspace{0.1em}{\scriptsize -3.2}} & 45.8\raisebox{-1.25ex}{\hspace{0.1em}{\scriptsize +60.9}} & 21.5\raisebox{-1.25ex}{\hspace{0.1em}{\scriptsize -5.6}} & .844\raisebox{-1.25ex}{\hspace{0.1em}{\scriptsize -4.1}} & .590\raisebox{-1.25ex}{\hspace{0.1em}{\scriptsize -4.6}} & \underline{37.8}\raisebox{-1.25ex}{\hspace{0.1em}{\scriptsize +64.7}} & \underline{26.4}\raisebox{-1.25ex}{\hspace{0.1em}{\scriptsize +13.4}} & \textbf{.402}\raisebox{-1.25ex}{\hspace{0.1em}{\scriptsize +33.5}} & \underline{.372}\raisebox{-1.25ex}{\hspace{0.1em}{\scriptsize +15.2}} \\
\raisebox{-0.5ex}{LoTA (62.5\%)} & .508\raisebox{-1.25ex}{\hspace{0.1em}{\scriptsize -24.7}} & 8.8\raisebox{-1.25ex}{\hspace{0.1em}{\scriptsize -73.0}} & 3.49\raisebox{-1.25ex}{\hspace{0.1em}{\scriptsize -12.3}} & .660\raisebox{-1.25ex}{\hspace{0.1em}{\scriptsize -17.1}} & .832\raisebox{-1.25ex}{\hspace{0.1em}{\scriptsize -2.3}} & \textbf{46.7}\raisebox{-1.25ex}{\hspace{0.1em}{\scriptsize +64.1}} & 21.6\raisebox{-1.25ex}{\hspace{0.1em}{\scriptsize -5.1}} & .853\raisebox{-1.25ex}{\hspace{0.1em}{\scriptsize -3.1}} & .596\raisebox{-1.25ex}{\hspace{0.1em}{\scriptsize -3.6}} & \textbf{37.9}\raisebox{-1.25ex}{\hspace{0.1em}{\scriptsize +65.1}} & \underline{26.4}\raisebox{-1.25ex}{\hspace{0.1em}{\scriptsize +13.4}} & \textbf{.402}\raisebox{-1.25ex}{\hspace{0.1em}{\scriptsize +33.5}} & .370\raisebox{-1.25ex}{\hspace{0.1em}{\scriptsize +14.6}} \\
\raisebox{-0.5ex}{LoTA (75.0\%)} & .573\raisebox{-1.25ex}{\hspace{0.1em}{\scriptsize -15.1}} & 10.2\raisebox{-1.25ex}{\hspace{0.1em}{\scriptsize -68.7}} & 3.76\raisebox{-1.25ex}{\hspace{0.1em}{\scriptsize -5.5}} & .672\raisebox{-1.25ex}{\hspace{0.1em}{\scriptsize -15.6}} & .838\raisebox{-1.25ex}{\hspace{0.1em}{\scriptsize -1.6}} & \underline{46.3}\raisebox{-1.25ex}{\hspace{0.1em}{\scriptsize +62.7}} & 22.2\raisebox{-1.25ex}{\hspace{0.1em}{\scriptsize -2.5}} & .853\raisebox{-1.25ex}{\hspace{0.1em}{\scriptsize -3.1}} & .593\raisebox{-1.25ex}{\hspace{0.1em}{\scriptsize -4.1}} & 37.6\raisebox{-1.25ex}{\hspace{0.1em}{\scriptsize +63.8}} & 26.3\raisebox{-1.25ex}{\hspace{0.1em}{\scriptsize +12.9}} & .389\raisebox{-1.25ex}{\hspace{0.1em}{\scriptsize +29.2}} & .369\raisebox{-1.25ex}{\hspace{0.1em}{\scriptsize +14.3}} \\
\raisebox{-0.5ex}{LoTA (87.5\%)} & .648\raisebox{-1.25ex}{\hspace{0.1em}{\scriptsize -4.0}} & 18.0\raisebox{-1.25ex}{\hspace{0.1em}{\scriptsize -44.7}} & 3.84\raisebox{-1.25ex}{\hspace{0.1em}{\scriptsize -3.5}} & .681\raisebox{-1.25ex}{\hspace{0.1em}{\scriptsize -14.5}} & .844\raisebox{-1.25ex}{\hspace{0.1em}{\scriptsize -0.8}} & 45.8\raisebox{-1.25ex}{\hspace{0.1em}{\scriptsize +60.9}} & \underline{22.9}\raisebox{-1.25ex}{\hspace{0.1em}{\scriptsize +0.6}} & .863\raisebox{-1.25ex}{\hspace{0.1em}{\scriptsize -1.9}} & \underline{.603}\raisebox{-1.25ex}{\hspace{0.1em}{\scriptsize -2.4}} & 35.1\raisebox{-1.25ex}{\hspace{0.1em}{\scriptsize +52.9}} & 26.2\raisebox{-1.25ex}{\hspace{0.1em}{\scriptsize +12.5}} & .376\raisebox{-1.25ex}{\hspace{0.1em}{\scriptsize +24.9}} & .348\raisebox{-1.25ex}{\hspace{0.1em}{\scriptsize +7.8}} \\
\raisebox{-0.5ex}{$\mathbf{\star}$~LoTA (90\%)} & .638\raisebox{-1.25ex}{\hspace{0.1em}{\scriptsize -5.4}} & 20.4\raisebox{-1.25ex}{\hspace{0.1em}{\scriptsize -37.4}} & \underline{3.98}\raisebox{-1.25ex}{\hspace{0.1em}{\scriptsize +0.0}} & .706\raisebox{-1.25ex}{\hspace{0.1em}{\scriptsize -11.3}} & .827\raisebox{-1.25ex}{\hspace{0.1em}{\scriptsize -2.8}} & 45.2\raisebox{-1.25ex}{\hspace{0.1em}{\scriptsize +58.8}} & 22.7\raisebox{-1.25ex}{\hspace{0.1em}{\scriptsize -0.3}} & \underline{.864}\raisebox{-1.25ex}{\hspace{0.1em}{\scriptsize -1.8}} & \textbf{.606}\raisebox{-1.25ex}{\hspace{0.1em}{\scriptsize -2.0}} & 34.4\raisebox{-1.25ex}{\hspace{0.1em}{\scriptsize +49.9}} & 26.2\raisebox{-1.25ex}{\hspace{0.1em}{\scriptsize +12.5}} & .366\raisebox{-1.25ex}{\hspace{0.1em}{\scriptsize +21.5}} & .360\raisebox{-1.25ex}{\hspace{0.1em}{\scriptsize +11.5}} \\ \midrule
\raisebox{-0.5ex}{$\mathbf{\star}$~S2FT (Down)} & \textbf{.695}\raisebox{-1.25ex}{\hspace{0.1em}{\scriptsize +3.0}} & \textbf{27.9}\raisebox{-1.25ex}{\hspace{0.1em}{\scriptsize -14.3}} & \textbf{3.99}\raisebox{-1.25ex}{\hspace{0.1em}{\scriptsize +0.3}} & .732\raisebox{-1.25ex}{\hspace{0.1em}{\scriptsize -8.0}} & .834\raisebox{-1.25ex}{\hspace{0.1em}{\scriptsize -2.0}} & 36.7\raisebox{-1.25ex}{\hspace{0.1em}{\scriptsize +29.0}} & 22.6\raisebox{-1.25ex}{\hspace{0.1em}{\scriptsize -0.7}} & .857\raisebox{-1.25ex}{\hspace{0.1em}{\scriptsize -2.6}} & \underline{.603}\raisebox{-1.25ex}{\hspace{0.1em}{\scriptsize -2.4}} & 21.7\raisebox{-1.25ex}{\hspace{0.1em}{\scriptsize -5.4}} & 26.0\raisebox{-1.25ex}{\hspace{0.1em}{\scriptsize +11.6}} & .303\raisebox{-1.25ex}{\hspace{0.1em}{\scriptsize +0.6}} & .331\raisebox{-1.25ex}{\hspace{0.1em}{\scriptsize +2.5}} \\
\raisebox{-0.5ex}{S2FT (Down + Output)} & .635\raisebox{-1.25ex}{\hspace{0.1em}{\scriptsize -5.9}} & 19.5\raisebox{-1.25ex}{\hspace{0.1em}{\scriptsize -40.1}} & 3.75\raisebox{-1.25ex}{\hspace{0.1em}{\scriptsize -5.7}} & .306\raisebox{-1.25ex}{\hspace{0.1em}{\scriptsize -61.6}} & .822\raisebox{-1.25ex}{\hspace{0.1em}{\scriptsize -3.4}} & 30.0\raisebox{-1.25ex}{\hspace{0.1em}{\scriptsize +5.4}} & 21.9\raisebox{-1.25ex}{\hspace{0.1em}{\scriptsize -3.8}} & .632\raisebox{-1.25ex}{\hspace{0.1em}{\scriptsize -28.2}} & .393\raisebox{-1.25ex}{\hspace{0.1em}{\scriptsize -36.4}} & 19.7\raisebox{-1.25ex}{\hspace{0.1em}{\scriptsize -14.2}} & 25.3\raisebox{-1.25ex}{\hspace{0.1em}{\scriptsize +8.6}} & .279\raisebox{-1.25ex}{\hspace{0.1em}{\scriptsize -7.3}} & .245\raisebox{-1.25ex}{\hspace{0.1em}{\scriptsize -24.1}} \\
\raisebox{-0.5ex}{S2FT (Down; $r=16$)} & \underline{.678}\raisebox{-1.25ex}{\hspace{0.1em}{\scriptsize +0.5}} & \underline{25.7}\raisebox{-1.25ex}{\hspace{0.1em}{\scriptsize -21.1}} & 3.96\raisebox{-1.25ex}{\hspace{0.1em}{\scriptsize -0.5}} & \underline{.735}\raisebox{-1.25ex}{\hspace{0.1em}{\scriptsize -7.7}} & .841\raisebox{-1.25ex}{\hspace{0.1em}{\scriptsize -1.2}} & 38.7\raisebox{-1.25ex}{\hspace{0.1em}{\scriptsize +36.0}} & 22.8\raisebox{-1.25ex}{\hspace{0.1em}{\scriptsize +0.1}} & .852\raisebox{-1.25ex}{\hspace{0.1em}{\scriptsize -3.2}} & \textbf{.606}\raisebox{-1.25ex}{\hspace{0.1em}{\scriptsize -2.0}} & 24.7\raisebox{-1.25ex}{\hspace{0.1em}{\scriptsize +7.6}} & 25.9\raisebox{-1.25ex}{\hspace{0.1em}{\scriptsize +11.2}} & .314\raisebox{-1.25ex}{\hspace{0.1em}{\scriptsize +4.3}} & .328\raisebox{-1.25ex}{\hspace{0.1em}{\scriptsize +1.6}} \\
\raisebox{-0.5ex}{S2FT (Down; $r=32$)} & .661\raisebox{-1.25ex}{\hspace{0.1em}{\scriptsize -2.0}} & 21.6\raisebox{-1.25ex}{\hspace{0.1em}{\scriptsize -33.7}} & 3.92\raisebox{-1.25ex}{\hspace{0.1em}{\scriptsize -1.5}} & .706\raisebox{-1.25ex}{\hspace{0.1em}{\scriptsize -11.3}} & .837\raisebox{-1.25ex}{\hspace{0.1em}{\scriptsize -1.7}} & 41.7\raisebox{-1.25ex}{\hspace{0.1em}{\scriptsize +46.5}} & 22.7\raisebox{-1.25ex}{\hspace{0.1em}{\scriptsize -0.3}} & .860\raisebox{-1.25ex}{\hspace{0.1em}{\scriptsize -2.3}} & \underline{.603}\raisebox{-1.25ex}{\hspace{0.1em}{\scriptsize -2.4}} & 27.4\raisebox{-1.25ex}{\hspace{0.1em}{\scriptsize +19.4}} & 26.1\raisebox{-1.25ex}{\hspace{0.1em}{\scriptsize +12.1}} & .316\raisebox{-1.25ex}{\hspace{0.1em}{\scriptsize +4.9}} & .333\raisebox{-1.25ex}{\hspace{0.1em}{\scriptsize +3.1}} \\
\raisebox{-0.5ex}{S2FT (Down; $r=64$)} & .652\raisebox{-1.25ex}{\hspace{0.1em}{\scriptsize -3.4}} & 19.7\raisebox{-1.25ex}{\hspace{0.1em}{\scriptsize -39.5}} & 3.82\raisebox{-1.25ex}{\hspace{0.1em}{\scriptsize -4.0}} & .683\raisebox{-1.25ex}{\hspace{0.1em}{\scriptsize -14.2}} & \underline{.846}\raisebox{-1.25ex}{\hspace{0.1em}{\scriptsize -0.6}} & 43.2\raisebox{-1.25ex}{\hspace{0.1em}{\scriptsize +51.8}} & \underline{22.9}\raisebox{-1.25ex}{\hspace{0.1em}{\scriptsize +0.6}} & .859\raisebox{-1.25ex}{\hspace{0.1em}{\scriptsize -2.4}} & \underline{.603}\raisebox{-1.25ex}{\hspace{0.1em}{\scriptsize -2.4}} & 31.0\raisebox{-1.25ex}{\hspace{0.1em}{\scriptsize +35.1}} & 26.3\raisebox{-1.25ex}{\hspace{0.1em}{\scriptsize +12.9}} & .317\raisebox{-1.25ex}{\hspace{0.1em}{\scriptsize +5.3}} & .344\raisebox{-1.25ex}{\hspace{0.1em}{\scriptsize +6.5}} \\

\bottomrule
\end{tabular}
}
\caption{Ablation study of LoTA and S2FT across varying sparsity levels with Igbo as the target language.
\textbf{Bold} and \underline{underlined} denote best and second-best adaptation approaches with relative changes in subscripts.
$\mathbf{\star}$ indicates the default configuration tested in Table \ref{tab:additional_baselines}.
}
\label{tab:lota_s2ft_ablation}
\end{table*}

\begin{table*}[th!]
\centering
\small
\resizebox{\textwidth}{!}{
\begin{tabular}{lc@{\hspace{6pt}}c@{\hspace{6pt}}c@{\hspace{6pt}}c@{\hspace{6pt}}c@{\hspace{6pt}}c@{\hspace{6pt}}c@{\hspace{6pt}}c@{\hspace{6pt}}c@{\hspace{6pt}}c@{\hspace{6pt}}c@{\hspace{6pt}}c@{\hspace{6pt}}c}
\toprule
& \multicolumn{4}{c}{\textbf{Chat and Instruction-following}} & \textbf{Safety} & \multicolumn{4}{c}{\textbf{Source language}} & \multicolumn{4}{c}{\textbf{Target language} (Igbo)} \\
\cmidrule(lr){2-5} \cmidrule(lr){7-10} \cmidrule(lr){11-14}
\textbf{Approach} & \textbf{IFEval} & \textbf{AE2} & \textbf{MTB} & \textbf{GSM8K} & \textbf{T3} & \textbf{MT} & \textbf{SUM} & \textbf{MRC} & \textbf{MMLU} & \textbf{MT} & \textbf{SUM} & \textbf{MRC} & \textbf{MMLU} \\
\midrule

\cellcolor{gray!20}\raisebox{-0.5ex}{Source} & \cellcolor{gray!20}.797\raisebox{-1.25ex}{\hspace{0.1em}{\scriptsize +0.0}} & \cellcolor{gray!20}29.7\raisebox{-1.25ex}{\hspace{0.1em}{\scriptsize +0.0}} & \cellcolor{gray!20}4.16\raisebox{-1.25ex}{\hspace{0.1em}{\scriptsize +0.0}} & \cellcolor{gray!20}.853\raisebox{-1.25ex}{\hspace{0.1em}{\scriptsize +0.0}} & \cellcolor{gray!20}.786\raisebox{-1.25ex}{\hspace{0.1em}{\scriptsize +0.0}} & \cellcolor{gray!20}29.4\raisebox{-1.25ex}{\hspace{0.1em}{\scriptsize +0.0}} & \cellcolor{gray!20}23.6\raisebox{-1.25ex}{\hspace{0.1em}{\scriptsize +0.0}} & \cellcolor{gray!20}.871\raisebox{-1.25ex}{\hspace{0.1em}{\scriptsize +0.0}} & \cellcolor{gray!20}.625\raisebox{-1.25ex}{\hspace{0.1em}{\scriptsize +0.0}} & \cellcolor{gray!20}24.6\raisebox{-1.25ex}{\hspace{0.1em}{\scriptsize +0.0}} & \cellcolor{gray!20}23.7\raisebox{-1.25ex}{\hspace{0.1em}{\scriptsize +0.0}} & \cellcolor{gray!20}.329\raisebox{-1.25ex}{\hspace{0.1em}{\scriptsize +0.0}} & \cellcolor{gray!20}.330\raisebox{-1.25ex}{\hspace{0.1em}{\scriptsize +0.0}} \\ \midrule
\raisebox{-0.5ex}{FFT} & .717\raisebox{-1.25ex}{\hspace{0.1em}{\scriptsize -10.1}} & 28.0\raisebox{-1.25ex}{\hspace{0.1em}{\scriptsize -5.7}} & 4.11\raisebox{-1.25ex}{\hspace{0.1em}{\scriptsize -1.3}} & .666\raisebox{-1.25ex}{\hspace{0.1em}{\scriptsize -21.9}} & .778\raisebox{-1.25ex}{\hspace{0.1em}{\scriptsize -1.0}} & \cellcolor{green!20}35.0\raisebox{-1.25ex}{\hspace{0.1em}{\scriptsize +19.1}} & \underline{\cellcolor{green!20}23.7}\raisebox{-1.25ex}{\hspace{0.1em}{\scriptsize +0.6}} & .829\raisebox{-1.25ex}{\hspace{0.1em}{\scriptsize -4.8}} & .608\raisebox{-1.25ex}{\hspace{0.1em}{\scriptsize -2.7}} & \textbf{\cellcolor{green!20}34.4}\raisebox{-1.25ex}{\hspace{0.1em}{\scriptsize +40.0}} & \underline{\cellcolor{green!20}26.2}\raisebox{-1.25ex}{\hspace{0.1em}{\scriptsize +10.5}} & \textbf{\cellcolor{green!20}.388}\raisebox{-1.25ex}{\hspace{0.1em}{\scriptsize +18.0}} & \underline{\cellcolor{green!20}.383}\raisebox{-1.25ex}{\hspace{0.1em}{\scriptsize +16.0}} \\
\raisebox{-0.5ex}{GMT} & .741\raisebox{-1.25ex}{\hspace{0.1em}{\scriptsize -7.1}} & 27.7\raisebox{-1.25ex}{\hspace{0.1em}{\scriptsize -6.7}} & 4.16\raisebox{-1.25ex}{\hspace{0.1em}{\scriptsize -0.1}} & .719\raisebox{-1.25ex}{\hspace{0.1em}{\scriptsize -15.7}} & .773\raisebox{-1.25ex}{\hspace{0.1em}{\scriptsize -1.7}} & \underline{\cellcolor{green!20}35.4}\raisebox{-1.25ex}{\hspace{0.1em}{\scriptsize +20.5}} & \textbf{\cellcolor{green!20}23.8}\raisebox{-1.25ex}{\hspace{0.1em}{\scriptsize +1.0}} & .840\raisebox{-1.25ex}{\hspace{0.1em}{\scriptsize -3.6}} & .610\raisebox{-1.25ex}{\hspace{0.1em}{\scriptsize -2.4}} & \underline{\cellcolor{green!20}34.3}\raisebox{-1.25ex}{\hspace{0.1em}{\scriptsize +39.6}} & \underline{\cellcolor{green!20}26.2}\raisebox{-1.25ex}{\hspace{0.1em}{\scriptsize +10.5}} & \underline{\cellcolor{green!20}.379}\raisebox{-1.25ex}{\hspace{0.1em}{\scriptsize +15.2}} & \textbf{\cellcolor{green!20}.385}\raisebox{-1.25ex}{\hspace{0.1em}{\scriptsize +16.6}} \\
\raisebox{-0.5ex}{HFT} & .780\raisebox{-1.25ex}{\hspace{0.1em}{\scriptsize -2.2}} & 28.9\raisebox{-1.25ex}{\hspace{0.1em}{\scriptsize -2.7}} & 4.11\raisebox{-1.25ex}{\hspace{0.1em}{\scriptsize -1.3}} & .695\raisebox{-1.25ex}{\hspace{0.1em}{\scriptsize -18.5}} & \underline{\cellcolor{green!20}.788}\raisebox{-1.25ex}{\hspace{0.1em}{\scriptsize +0.3}} & \cellcolor{green!20}33.5\raisebox{-1.25ex}{\hspace{0.1em}{\scriptsize +14.0}} & 23.5\raisebox{-1.25ex}{\hspace{0.1em}{\scriptsize -0.2}} & .847\raisebox{-1.25ex}{\hspace{0.1em}{\scriptsize -2.8}} & .609\raisebox{-1.25ex}{\hspace{0.1em}{\scriptsize -2.5}} & \cellcolor{green!20}32.8\raisebox{-1.25ex}{\hspace{0.1em}{\scriptsize +33.5}} & \underline{\cellcolor{green!20}26.2}\raisebox{-1.25ex}{\hspace{0.1em}{\scriptsize +10.5}} & \cellcolor{green!20}.364\raisebox{-1.25ex}{\hspace{0.1em}{\scriptsize +10.7}} & \cellcolor{green!20}.368\raisebox{-1.25ex}{\hspace{0.1em}{\scriptsize +11.4}} \\
\raisebox{-0.5ex}{LoTA (90\% Sparsity)} & .781\raisebox{-1.25ex}{\hspace{0.1em}{\scriptsize -2.0}} & \underline{29.3}\raisebox{-1.25ex}{\hspace{0.1em}{\scriptsize -1.3}} & 4.13\raisebox{-1.25ex}{\hspace{0.1em}{\scriptsize -0.8}} & .723\raisebox{-1.25ex}{\hspace{0.1em}{\scriptsize -15.2}} & .784\raisebox{-1.25ex}{\hspace{0.1em}{\scriptsize -0.3}} & \cellcolor{green!20}33.2\raisebox{-1.25ex}{\hspace{0.1em}{\scriptsize +13.0}} & \underline{\cellcolor{green!20}23.7}\raisebox{-1.25ex}{\hspace{0.1em}{\scriptsize +0.6}} & \underline{.853}\raisebox{-1.25ex}{\hspace{0.1em}{\scriptsize -2.1}} & .608\raisebox{-1.25ex}{\hspace{0.1em}{\scriptsize -2.7}} & \cellcolor{green!20}31.8\raisebox{-1.25ex}{\hspace{0.1em}{\scriptsize +29.4}} & \cellcolor{green!20}26.0\raisebox{-1.25ex}{\hspace{0.1em}{\scriptsize +9.7}} & \cellcolor{green!20}.343\raisebox{-1.25ex}{\hspace{0.1em}{\scriptsize +4.3}} & \cellcolor{green!20}.358\raisebox{-1.25ex}{\hspace{0.1em}{\scriptsize +8.4}} \\
\raisebox{-0.5ex}{S2FT (Down)} & \textbf{\cellcolor{green!20}.807}\raisebox{-1.25ex}{\hspace{0.1em}{\scriptsize +1.2}} & 28.6\raisebox{-1.25ex}{\hspace{0.1em}{\scriptsize -3.7}} & \textbf{\cellcolor{green!20}4.18}\raisebox{-1.25ex}{\hspace{0.1em}{\scriptsize +0.4}} & \textbf{.851}\raisebox{-1.25ex}{\hspace{0.1em}{\scriptsize -0.2}} & \textbf{\cellcolor{green!20}.803}\raisebox{-1.25ex}{\hspace{0.1em}{\scriptsize +2.2}} & 28.9\raisebox{-1.25ex}{\hspace{0.1em}{\scriptsize -1.7}} & 23.5\raisebox{-1.25ex}{\hspace{0.1em}{\scriptsize -0.2}} & \textbf{.864}\raisebox{-1.25ex}{\hspace{0.1em}{\scriptsize -0.8}} & \textbf{\cellcolor{green!20}.627}\raisebox{-1.25ex}{\hspace{0.1em}{\scriptsize +0.3}} & 24.1\raisebox{-1.25ex}{\hspace{0.1em}{\scriptsize -1.9}} & 22.8\raisebox{-1.25ex}{\hspace{0.1em}{\scriptsize -3.8}} & .329\raisebox{-1.25ex}{\hspace{0.1em}{\scriptsize +0.0}} & .330\raisebox{-1.25ex}{\hspace{0.1em}{\scriptsize -0.1}} \\
\raisebox{-0.5ex}{SSU-Wanda} & \underline{\cellcolor{green!20}.799}\raisebox{-1.25ex}{\hspace{0.1em}{\scriptsize +0.2}} & \textbf{\cellcolor{green!20}31.0}\raisebox{-1.25ex}{\hspace{0.1em}{\scriptsize +4.4}} & \underline{\cellcolor{green!20}4.17}\raisebox{-1.25ex}{\hspace{0.1em}{\scriptsize +0.1}} & \underline{.777}\raisebox{-1.25ex}{\hspace{0.1em}{\scriptsize -8.9}} & .781\raisebox{-1.25ex}{\hspace{0.1em}{\scriptsize -0.6}} & \textbf{\cellcolor{green!20}37.9}\raisebox{-1.25ex}{\hspace{0.1em}{\scriptsize +29.0}} & 23.5\raisebox{-1.25ex}{\hspace{0.1em}{\scriptsize -0.2}} & .851\raisebox{-1.25ex}{\hspace{0.1em}{\scriptsize -2.3}} & \underline{.618}\raisebox{-1.25ex}{\hspace{0.1em}{\scriptsize -1.1}} & \cellcolor{green!20}34.0\raisebox{-1.25ex}{\hspace{0.1em}{\scriptsize +38.4}} & \textbf{\cellcolor{green!20}26.4}\raisebox{-1.25ex}{\hspace{0.1em}{\scriptsize +11.4}} & \cellcolor{green!20}.357\raisebox{-1.25ex}{\hspace{0.1em}{\scriptsize +8.5}} & \cellcolor{green!20}.366\raisebox{-1.25ex}{\hspace{0.1em}{\scriptsize +10.8}} \\

\bottomrule
\end{tabular}
}
\caption{Performance on Olmo-3-7B-Instruct. We use compare all selective parameter tuning baselines: GMT, HFT, LoTA, and S2FT with SSU-Wanda.
We use Igbo as the target language.
Relative changes (\%) compared to Source are indicated as subscripts.
Scores that are better than Source are highlighted in \colorbox{green!20}{green}.
The best and second-best adaptation approaches are indicated in \textbf{bold} and \underline{underlined}, respectively.}
\label{tab:olmo3_performance}
\end{table*}

\section{Supplementary Analysis} \label{appendix:analysis}
\subsection{Impact of Freezing Ratio on Baselines} \label{appendix:ratio_baseline}

We extend this analysis to state-of-the-art selective parameter update baselines (Figure \ref{fig:ratio_appendix}).
The closest baseline, the static method HFT, follows a trend similar to SSU but fails to surpass the performance of SSU across tasks and freezing ratios.
In contrast, the dynamic method GMT exhibits a different trend.
While it often achieves strong target language and MT performance at ratios above 60\%, it consistently yields low performance on monolingual source tasks regardless of the freezing ratio.
We attribute this to the dynamic nature of GMT, which allows updates to any parameter over time, leading to cumulative corruption from unstructured target data optimization (\S\ref{sec:results}).
Ultimately, this confirms SSU as the optimal method for simultaneously achieving strong source preservation and high target language gains.

\subsection{Calibration Data Size for Parameter Importance Scoring} \label{appendix:calibration_size}

SSU uses 500 source calibration examples by default to compute parameter importance scores (\S\ref{subsec:data}). To assess sensitivity to this hyperparameter, we compare the default (500 examples, $\sim$1M tokens) with a smaller 128-example set ($\sim$0.26M tokens), a size common in model pruning literature~\citep{williams-aletras-2024-impact}.
As shown in Table \ref{tab:calibration_size}, the results demonstrate minimal changes across tasks; the maximum performance difference observed is only 1.2 points on IFEval.
This confirms the robustness of SSU to calibration data size, demonstrating that a small sample set suffices for effective importance scoring.

\subsection{Comparison to Additional Baselines} \label{appendix:additional_baselines}

We compare SSU against two other recent selective parameter update methods: LoTA~\citep{panda2024lotteryticketadaptationmitigating} and S2FT~\citep{NEURIPS2024_6e3b9fb0}. For LoTA, we evaluate both its default 90\% sparsity and a 50\% sparsity setting that matches the freezing ratio of SSU. For S2FT, we evaluate the default sparsity configuration that sparsely tunes only down-projection layers.\looseness=-1

As detailed in Table \ref{tab:additional_baselines}, neither baseline achieves the balanced performance of SSU-Wanda. LoTA at 90\% sparsity exhibits inferior source preservation compared to SSU (7.6\% vs. 4.0\% average drop) and lower target gains (23.9\% vs. 30.7\%). While reducing LoTA sparsity to 50\% improves target gains to 31.7\%, it triggers severe catastrophic forgetting, with monolingual source performance dropping by 19.9\%. S2FT effectively preserves source capabilities (3.3\% drop) but yields negligible target gains (2.3\%). These results underscore that only SSU-Wanda simultaneously achieves strong source preservation and high target language gains comparable to FFT.

\paragraph{Sensitivity Analysis.} To ensure these findings are not artifacts of specific sparsity choices, we extend our evaluation with a fine-grained ablation study across varying sparsity levels (Table \ref{tab:lota_s2ft_ablation}).

\textbf{LoTA}:
We examine LoTA across sparsity ratios in 12.5\% increments. High sparsity configurations (e.g., 90\% and 87.5\%) preserve source performance reasonably well but consistently underperform SSU-Wanda on both source preservation and target acquisition. Conversely, lowering sparsity allows for more adaptation but disproportionately harms source capabilities. For instance, while LoTA at 50\% achieves a 31.7\% average target gain, surpassing the 30.7\% gain of SSU-Wanda.
However, it suffers a drastic 19.9\% drop in monolingual source tasks. This degradation worsens at 37.5\% sparsity, reaching a 25.4\% drop. This confirms that LoTA fails to find an optimal balance between the stability-plasticity trade-off required for effective adaptation.

\textbf{S2FT}:
Following the original paper~\citep{NEURIPS2024_6e3b9fb0}, we sparsely tune the down projection layers using a parameter count equivalent to LoRA with a rank of 8 (Table \ref{tab:params_cpt}). 
We expand the S2FT evaluation by increasing the trainable parameter budget to match LoRA ranks of 16, 32, and 64.
We also test the ``Down and Output'' projection tuning strategy to determine if the poor performance reported for Mistral and Llama3 (attributed to inflexible selection in multi-query attention) applies to OLMo 2.
First, increasing the parameter budget improves target performance slightly but erodes source capabilities without ever matching SSU. At the equivalent of rank 64, S2FT suffers a larger source drop (8.2\%) than SSU-Wanda (4.0\%) while achieving only half the target gains (15.0\% vs. 30.7\%).
Second, we confirm that tuning ``Down and Output'' projections yields suboptimal results for OLMo 2, causing severe drops of up to 23.1\% in source tasks.
In summary, regardless of sparsity-level adjustments, only SSU provides robust source preservation while improving target language abilities to levels comparable to FFT.

\subsection{Generalization to OLMo 3 Architecture} \label{appendix:olmo3}

To evaluate the generalizability of the SSU framework, we measure the performance of the method using the recent Olmo-3-7B-Instruct~\cite{olmo2025olmo3}, which was released on November 20, 2025.
Due to constraints on computational resources, this evaluation focuses on adapting the model to Igbo as the target language. We compare SSU against full fine-tuning (FFT) and all the selective parameter update baselines used in this study.\looseness=-1

Results in Table \ref{tab:olmo3_performance} demonstrate that SSU effectively preserves knowledge from the source, yielding an average relative performance degradation of only 1.1\% on monolingual source tasks. In comparison, FFT and GMT exhibit substantially higher degradation at 5.9\% and 4.5\%, respectively. Although S2FT avoids degradation almost entirely (-0.1\%), it fails to facilitate adaptation and results in performance in the target language that is 1.4\% lower than the original model.

In the target language tasks, SSU achieves average relative gains of 17.3\%. While target-driven signals in GMT lead to higher target improvements (20.5\%), this approach causes substantially more forgetting than SSU (4.5\% versus 1.1\%). Furthermore, SSU outperforms static selective parameter update methods such as HFT and LoTA. Both achieve lower target gains (16.5\% for HFT and 13.0\% for LoTA) and higher source degradation (3.7\% for HFT and 3.0\% for LoTA).

In summary, SSU achieves the most effective balance by maintaining the general-purpose capability while providing consistent performance gains in the target language. This confirms that SSU remains effective for the recent fully-open instruct model.\looseness=-1

\subsection{Theoretical Analysis} \label{appendix:theory}
SSU addresses the stability-plasticity dilemma in neural systems~\citep{grossberg2012studies}, balancing plasticity for new knowledge with stability for prior knowledge. By identifying and freezing a source-critical subnetwork, SSU extends the Lottery Ticket Hypothesis~\citep{frankle2018the} to the domain of transfer learning. The use of an importance score to shield crucial parameters enforces a hard constraint, confining updates to a subspace that avoids interfering with source language knowledge. This aligns with recent findings on spurious forgetting~\citep{zheng2025spurious}, which suggest that performance drops often stem from task misalignment caused by nearly orthogonal weight updates.

Furthermore, SSU employs structured, column-wise masking specifically to preserve entire learned features. Unlike unstructured pruning, which can degrade learned representations arbitrarily, pruning entire columns of a weight matrix corresponds to removing specific neurons or feature detectors~\citep{voita-etal-2019-analyzing}. This structural preservation ensures that the core feature space of the source model remains intact, enabling effective adaptation to the target language.

\subsection{Proxy Evaluation on Target-Language Instruction-following} \label{appendix:proxy}

\begin{table*}[th]
\centering
\small
\begin{tabular}{llcccccccccc}
\toprule
& & \multicolumn{5}{c}{\textbf{Target-to-English MT}} & \multicolumn{5}{c}{\textbf{English-to-target MT}} \\
\cmidrule(lr){3-7} \cmidrule(lr){8-12}
\multicolumn{2}{l}{\textbf{Approach}} & ne & ky & am & ha & ig & ne & ky & am & ha & ig \\
\midrule

\multirow{8}{*}{\rotatebox{90}{7B}} & \cellcolor{gray!20}Source & \cellcolor{gray!20}.906 & \cellcolor{gray!20}.843 & \cellcolor{gray!20}.640 & \cellcolor{gray!20}.905 & \cellcolor{gray!20}.857 & \cellcolor{gray!20}.450 & \cellcolor{gray!20}.513 & \cellcolor{gray!20}.019 & \cellcolor{gray!20}.769 & \cellcolor{gray!20}.798 \\ \cmidrule{2-12}
 & FFT & .785 & .876 & .525 & .870 & .670 & .014 & .399 & .006 & .022 & .398 \\
 & AdaLoRA & .902 & .809 & .321 & .871 & .835 & .083 & .225 & .000 & .061 & .035 \\
 & HFT & \underline{.906} & \underline{.879} & .726 & .898 & .898 & .021 & .510 & .002 & .031 & .667 \\
 & GMT & \textbf{.909} & .873 & .706 & .919 & \underline{.909} & .015 & .463 & .003 & .101 & \underline{.859} \\ \cmidrule{2-12}
 & SSU-Rand & .904 & .877 & \underline{.744} & \underline{.929} & .907 & \underline{.084} & \underline{.545} & \underline{.009} & \underline{.108} & .760 \\
 & SSU-Mag & .897 & .873 & .735 & .831 & .857 & .015 & .471 & .005 & .006 & .581 \\ \noalign{\vskip\aboverulesep}\cdashline{2-12}[2pt/1.2pt]\noalign{\vskip\belowrulesep}
 & SSU-Wanda & .901 & \textbf{.880} & \textbf{.749} & \textbf{.956} & \textbf{.922} & \textbf{.437} & \textbf{.557} & \textbf{.015} & \textbf{.634} & \textbf{.906} \\
\midrule
\multirow{8}{*}{\rotatebox{90}{13B}} & \cellcolor{gray!20}Source & \cellcolor{gray!20}.915 & \cellcolor{gray!20}.871 & \cellcolor{gray!20}.881 & \cellcolor{gray!20}.937 & \cellcolor{gray!20}.942 & \cellcolor{gray!20}.455 & \cellcolor{gray!20}.565 & \cellcolor{gray!20}.041 & \cellcolor{gray!20}.749 & \cellcolor{gray!20}.866 \\ \cmidrule{2-12}
 & FFT & .821 & .749 & .681 & .729 & .318 & .012 & .171 & .004 & .038 & .240 \\
 & AdaLoRA & \textbf{.933} & \underline{.779} & .621 & .888 & \underline{.787} & \underline{.021} & .120 & .001 & .032 & .019 \\
 & HFT & .923 & .752 & \underline{.799} & \underline{.897} & .559 & .019 & \underline{.560} & .005 & .324 & \underline{.661} \\
 & GMT & .826 & .713 & .411 & .758 & .521 & .011 & .252 & \underline{.011} & .087 & .255 \\ \cmidrule{2-12}
 & SSU-Rand & .910 & .773 & .687 & .888 & .685 & .011 & .499 & .003 & \underline{.387} & .521 \\
 & SSU-Mag & .888 & .721 & .615 & .793 & .447 & .019 & .382 & .002 & .099 & .254 \\ \noalign{\vskip\aboverulesep}\cdashline{2-12}[2pt/1.2pt]\noalign{\vskip\belowrulesep}
 & SSU-Wanda & \underline{.930} & \textbf{.861} & \textbf{.873} & \textbf{.945} & \textbf{.875} & \textbf{.131} & \textbf{.599} & \textbf{.035} & \textbf{.691} & \textbf{.884} \\

\bottomrule
\end{tabular}
\caption{Instruction-following performance on MT tasks, evaluated using the IFEval framework.
IFEval-style strict accuracy is measured against three verifiable criteria: (i) the response is monolingual in the specified language, (ii) it matches the reference sentence count, and (iii) it uses a language-appropriate full stop.
The best and second-best adaptation approaches for each model scale are indicated in \textbf{bold} and \underline{underlined}, respectively.}
\label{tab:combined_mt_ifeval}
\end{table*}

Evaluating instruction-following abilities in underrepresented languages is challenging due to data scarcity and unreliable LLM-based judges~\citep{azime-etal-2024-walia}.
Given these limitations, we establish a tractable proxy for the evaluation.

Specifically, we repurpose machine translation (MT) to serve as a proxy task for instruction-following in a target language.
A key aspect of the methodology is instructing models in the target language, not English (Table \ref{tab:prompt}).
This design assesses how well a model comprehends and executes instructions within a specific linguistic context, providing a more realistic test of target-language instruction-following.
This method thus measures both translation quality and the ability to perform a directed task from instructions in a non-English language.\looseness=-1

To quantify performance, we adapt the verifiable evaluation framework of IFEval~\citep{zhou2023instructionfollowingevaluationlargelanguage} and its multilingual extension~\citep{zeng-etal-2025-marco}. We compute a strict accuracy score, where a response is considered correct only if it satisfies three verifiable criteria: (i) the response is monolingual in the specified language, (ii) the number of sentences matches that of the gold reference, and (iii) the response ends with a language-appropriate full stop.\looseness=-1

\paragraph{Implementation.}
Response generation follows the exact setup described for the MT evaluation in the main paper. We score responses against the three verifiable criteria using the following checks:
\begin{itemize}
    \item \textbf{Response language}: We use GlotLID to compute a normalized confidence score for the specified language in each response (i.e., English for target-to-English MT, target language for English-to-target MT). A response with a score below 0.9 is considered code-mixed and fails this criterion.
    
    \item \textbf{Sentence count}: For target-to-English MT, we count sentences in both the response and gold reference using the NLTK tokenizer~\citep{bird-loper-2004-nltk}. For English-to-target MT, we count sentences using a regular expression pattern (\texttt{[.?!\includegraphics[width=1em,height=1em]{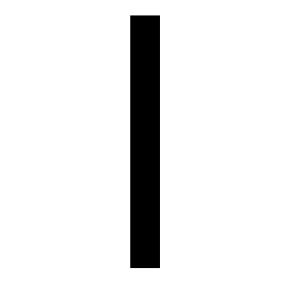}\includegraphics[width=1em,height=1em]{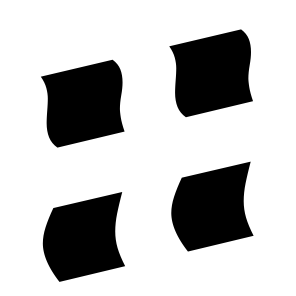}]}).

    \item \textbf{Full stop}: A target-to-English response must end with ``.''. For English-to-target translation, we check for language-specific full stops: ``\includegraphics[width=1em,height=1em]{figure/nepali_stop.pdf}'' or ``.'' for Nepali, ``\includegraphics[width=1em,height=1em]{figure/amharic_stop.pdf}'' for Amharic, and ``.'' for all other languages.
\end{itemize}

\paragraph{Results and Analysis.}
Results from the instruction-following evaluation (Table~\ref{tab:combined_mt_ifeval}) reveal the consistently strong performance of SSU-Wanda.
Across languages and both 7B and 13B model scales, SSU-Wanda achieves the highest strict accuracy scores.
This superiority is particularly pronounced in the challenging English-to-target direction, suggesting that its proactive, source-driven structured parameter selection strategy is effective for enhancing target-language instruction-following abilities related to formatting.

Nonetheless, a clear performance disparity emerges between the two translation directions.
Models consistently achieve higher accuracy on target-to-English tasks compared to English-to-target tasks.
This trend demonstrates that while models can comprehend instructions delivered in a non-English language, they more reliably execute those instructions when generating text in English.

The analysis reveals that generating non-English languages under implicit formatting constraints is a primary obstacle for current models.
This difficulty likely stems from the adaptation on unlabeled target language data.
The unlabeled target language corpus provides weak signals for formatting. Furthermore, the evaluation prompts only request translation without explicitly mentioning punctuation (Table \ref{tab:prompt}). Consequently, models learn linguistic patterns but fail to reliably apply specific formatting rules like terminal punctuation in the target language.
For instance, we observe that the best-performing SSU-Wanda models achieve only 1.8\% (7B) and 3.5\% (13B) adherence to the full stop criterion.
Therefore, developing methods to improve target-language instruction-following with only unlabeled corpora remains a crucial future research direction. Additionally, we hope this work inspires the development of extensive instruction-following benchmarks for low-resource languages.

\section{Extended Related Work} \label{appendix:related_work}

SSU addresses the core challenge of continual learning (CL) in machine learning: adapting a model to new tasks while mitigating catastrophic forgetting~\citep{goodfellow2015empiricalinvestigationcatastrophicforgetting,doi:10.1073/pnas.1611835114}.
This section situates SSU within the parameter-centric family of CL solutions.
These methods protect knowledge at the parameter level, typically without accessing data from the old task for replay.
They generally address two fundamental questions: (1) the \textbf{Identification Problem}, defining which parameters are critical to a previous task; and (2) the \textbf{Protection Problem}, determining the mechanism to enforce protection on those parameters.
Parameter-centric approaches largely fall into three categories: \textbf{soft, regularization-based} protection; \textbf{hard, architectural-based} protection; and \textbf{adaptive, hybrid} methods.

\paragraph{Soft Parameter Protection (Regularization-Based).}
These methods discourage changes to critical parameters by adding a penalty term to the loss function of the new task.
Approaches differ primarily in solving the Identification Problem. Elastic Weight Consolidation (EWC) identifies critical parameters via the Fisher Information Matrix diagonal~\citep{doi:10.1073/pnas.1611835114}, while Synaptic Intelligence (SI) computes importance online by tracking the cumulative contribution of each parameter to loss reduction~\citep{pmlr-v70-zenke17a}.
Similarly, Memory Aware Synapses (MAS) estimates importance weights based on the sensitivity of the learned function (output function) to parameter changes, eliminating the need for original labeled data~\citep{10.1007/978-3-030-01219-9_9}.
Soft-Masking of Parameter-Level Gradient Flow (SPG) protects knowledge by directly modulating gradient flow with soft masks rather than modifying the loss objective~\citep{pmlr-v202-konishi23a}.
However, such soft constraints often fail under severe distributional shifts~\citep{10205469}.
This limitation becomes particularly acute in our problem setup (i.e., adapting instruct models using unlabeled target language data), where optimization pressure from unlabeled target corpora can overpower regularization penalties.\looseness=-1

\paragraph{Hard Parameter Protection (Isolation \& Architectural).}
These methods enforce stability via structural constraints, such as freezing or allocating parameters, to ensure near-zero forgetting.
Hard Attention to the Task (HAT) learns a binary mask, forcing gradients to zero for parameters allocated by the mask from any previous task~\citep{pmlr-v80-serra18a}.
PackNet employs an ``iterative prune, fix, and retrain'' cycle, freezing the surviving ``packed'' weights and forcing new tasks to utilize only ``free'' parameters~\citep{8578908}.
Piggyback represents an extreme form, freezing an entire pre-trained backbone and learning new tasks solely by training new binary masks~\citep{10.1007/978-3-030-01225-0_5}.

\paragraph{Adaptive \& Hybrid Protection.}
This emerging class assesses the properties of an incoming task to select a protection strategy dynamically.
Context-aware Task-driven (CAT) automatically detects whether a new task resembles previous ones~\citep{NEURIPS2020_d7488039}, applying Hard Protection (binary mask) for dissimilar tasks and Soft Protection (attention) for similar tasks.
Parameter Allocation \& Regularization (PAR) identifies task relatedness and applies dynamic protection: ``easy'' tasks are handled via soft regularization, while ``difficult'' tasks trigger the hard allocation of a new, isolated expert model~\citep{10205469}.
While promising, the application of such dynamic allocation strategies to the specific constraints of LLM language adaptation remains an interesting avenue for future research.

\paragraph{Situating SSU within Continual Learning.}
SSU adapts these CL principles for the linguistic adaptation of instruct LLMs.
We characterize SSU as a source-focused method utilizing static hard parameter protection.
Specifically, it resolves the ``Identification Problem'' via source-data-driven importance scores (e.g., Wanda) and the ``Protection Problem'' via column-wise structural freezing.
While conceptually aligned with hard parameter protection, SSU overcomes specific limitations regarding \textbf{problem setting} and \textbf{scale}.
Foundational CL methods largely focus on task-incremental learning, where the model learns a sequence of discrete, labeled tasks (e.g., Task 1: MNIST, Task 2: CIFAR). Consequently, methods like HAT rely on task identifiers (Task IDs) at inference time to select the correct mask.
This requirement is incompatible with general-purpose instruct LLMs, where the input language (or task) is unknown and the model must operate as a unified entity without external task signals.
Regarding scale, foundational methods typically target architectures with fewer than 1B parameters (e.g., PackNet uses VGG-16 ($\sim$138M)~\citep{simonyan2015a}).
Methods like the iterative pruning and retraining cycles of PackNet often become computationally prohibitive when applied to billion-parameter LLMs.
In contrast, SSU utilizes a one-shot, static calculation of importance before training, making it computationally viable for modern transformer-based architectures.

\section{License} \label{appendix:license}

This study uses publicly available models and datasets with different licenses, as detailed below.
Note that all permit their use for academic research.

\paragraph{Model Licenses.}
The OLMo 2 family of models are distributed under Apache License 2.0.
\begin{itemize}
    \item 7B: \url{https://huggingface.co/allenai/OLMo-2-1124-7B-Instruct}
    \item 13B: \url{https://huggingface.co/allenai/OLMo-2-1124-13B-Instruct}
\end{itemize}

Olmo-3-7B-Instruct is also distributed under Apache License 2.0: \url{https://huggingface.co/allenai/Olmo-3-7B-Instruct}.

\paragraph{Data Licenses.}
tulu-3-sft-olmo-2-mixture is licensed under ODC-BY-1.0.
MADLAD-400 is licensed under CC-BY 4.0.
XL-Sum is licensed under CC BY-NC-SA 4.0.
Belebele and FLORES-200 are licensed under CC BY-SA 4.0.
MMLU, GSM8K, and HumanEval are distributed under the MIT License. 
Ai2 Safety Tool, AlpacaEval, IFEval, and MT-Bench are distributed under Apache License 2.0.\looseness=-1

\section{Use of Generative AI Tools} \label{appendix:genai}
The authors acknowledge the use of LLMs during the preparation of this work. Gemini 2.5 and 3.0 Pro were utilized to find related work and to improve the grammar and clarity of the draft. Additionally, GPT-5 served as a coding assistant for implementation and debugging.

\end{document}